\documentclass[10pt,conference]{IEEEtran}

\usepackage{framed,multirow}

\usepackage{amssymb}
\usepackage{placeins}
\usepackage{afterpage} 
\usepackage{latexsym}
\usepackage{listings}
\usepackage{longtable}
\usepackage{tabularx}
\usepackage{booktabs}
\usepackage{array}
\usepackage{graphicx}
\usepackage{svg}
\usepackage{amsmath}
\usepackage{mdwtab}
\usepackage{caption}
\usepackage{accents}
\usepackage{subcaption}
\usepackage{pifont}
\usepackage{dblfloatfix} 
\usepackage{enumitem} 
\setlist{noitemsep}
\usepackage[numbers,sort&compress]{natbib}
\usepackage{url}
\usepackage{xcolor}
\usepackage{hyperref}

\newcommand{\xmark}{\text{\ding{55}}}

\DeclareMathOperator*{\mean}{mean}

\begin{document}



\title{Spatiotemporal graph neural process for reconstruction, extrapolation, and classification of cardiac trajectories}%



\author{
    \IEEEauthorblockN{Jaume Banus\IEEEauthorrefmark{1}\thanks{Corresponding author: \href{mailto:jaume.banus-cobo@chuv.ch}{jaume.banus-cobo@chuv.ch}}, 
    Augustin C. Ogier\IEEEauthorrefmark{1}, 
    Roger Hullin\IEEEauthorrefmark{2}, 
    Philippe Meyer\IEEEauthorrefmark{3}, 
    Ruud B. van Heeswijk\IEEEauthorrefmark{1}, 
    Jonas Richiardi\IEEEauthorrefmark{1}\IEEEauthorrefmark{4}}
    
    \IEEEauthorblockA{\IEEEauthorrefmark{1}Department of Radiology, Lausanne University Hospital and University of Lausanne, Lausanne, Switzerland}
    \IEEEauthorblockA{\IEEEauthorrefmark{2}Cardiovascular Department, Lausanne University Hospital and University of Lausanne, Lausanne, Switzerland}
    \IEEEauthorblockA{\IEEEauthorrefmark{3}Department of Medicine, Geneva University Hospital and University of Geneva, Geneva, Switzerland}
    \IEEEauthorblockA{\IEEEauthorrefmark{4}CIBM Center for Biomedical Imaging, Lausanne, Switzerland}
}

\maketitle



%
%

\begin{abstract}
We present a probabilistic framework for modeling structured spatiotemporal dynamics from sparse observations, focusing on cardiac motion. Our approach integrates neural ordinary differential equations (NODEs), graph neural networks (GNNs), and neural processes into a unified model that captures uncertainty, temporal continuity, and anatomical structure. We represent dynamic systems as spatiotemporal multiplex graphs and model their latent trajectories using a GNN-parameterized vector field. Given the sparse context observations at node and edge levels, the model infers a distribution over latent initial states and control variables, enabling both interpolation and extrapolation of trajectories. We validate the method on three synthetic dynamical systems (coupled pendulum, Lorenz attractor, and Kuramoto oscillators) and two real-world cardiac imaging datasets—ACDC (N=150) and UK Biobank (N=526) — demonstrating accurate reconstruction, extrapolation, and disease classification capabilities. The model accurately reconstructs trajectories and extrapolates future cardiac cycles from a single observed cycle. It achieves state-of-the-art results on the ACDC classification task (up to 99\% accuracy), and detects atrial fibrillation in UK Biobank subjects with competitive performance (up to 67\% accuracy). This work introduces a flexible approach for analyzing cardiac motion and offers a foundation for graph-based learning in structured biomedical spatiotemporal time-series data.
\end{abstract}

\begin{IEEEkeywords}
Graph neural networks, Cardiac imaging, Spatiotemporal graphs, NODEs
\end{IEEEkeywords}




\section{Introduction}

Cardiovascular diseases (CVDs) are the leading global cause of mortality with a prevalence of approximately 6,000 per 100,000 people~\citep{roth_global_2020}. Assessing cardiac function is crucial for patient management, diagnosis, and risk stratification. Clinical evaluation often relies on scalar indicators derived from medical imaging, such as stroke volume (SV) or ejection fraction (EF). While these indicators provide valuable insights, they simplify the heart's complex dynamical behavior by focusing on two discrete time points - end-diastole (ED) and end-systole (ES), corresponding to the heart's fully relaxed and fully contracted states respectively.  This approach restricts the analysis of cardiac function to two instants and fails to capture the complex nature of cardiac dynamics. For instance, the heart remodels in different ways in order to meet the physiological demands, which implies that identical observed metrics (e.g. EF or chamber volumes) may arise from distinct physiological states, which highlights the need to understand the underlying dynamics that generate the observed states. 

\subsection{Cardiac Dynamics}

Cardiac dynamics during contraction and relaxation cycles are inherently spatiotemporal, governed by the interplay of structural, electrical, and mechanical processes~\citep{trayanovaCardiacElectromechanicalModels2011}. Several approaches exist to model these dynamics, ranging from detailed biophysical simulations~\citep{goos_biomechanical_2001,cedilnik_fast_2018,chapelle_energy-preserving_2012,lluch_calibration_2020} to data-driven techniques~\citep{dalton_emulation_2022}. Technically, these dynamics can be visualized as multivariate time series representing measurements across different heart regions or locations, for example motion in cardiovascular magnetic resonance imaging (CMR).

\subsection{Spatiotemporal Graphs for Cardiac Dynamics Modeling}

In this context, a promising approach is to represent the heart as a spatiotemporal graph~\citep{lu_dynamic_2021, dalton_emulation_2022, banus_deep_2023,inacio_cardiac_2023}, where nodes represent cardiac regions or mesh vertices while edges capture spatial and/or temporal relationships between them. Then a graph neural network (GNN)~\citep{scarselli_graph_2009} can be used to model cardiac anatomy and function. In~\citep{lu_dynamic_2021}, the nodes represent mesh vertices and the model is trained to predict future frames, using the latent embeddings for downstream classification tasks. Similarly, GNNs have been used to used to predict dynamic heart motion across mesh vertices, employing a multi-resolution strategy to model time-dependent position changes~\citep{dalton_emulation_2022}. In our previous work~\citep{banus_deep_2023}, the heart was represented as a multiplex graph, where nodes correspond to anatomical regions with image-derived features (e.g., wall thickness, volume), and edges encode spatial and temporal connectivity, combining spatial and temporal message passing. Depolarization wave propagation has also been modeled using GNNs applied to mesh graphs~\citep{meister_graph_2021}. Collectively, these studies demonstrate the potential of GNNs to learn both anatomical and physiological patterns in cardiac data. Leveraging this graph representation, one can use GNNs to estimate latent representations based on aggregating information of a node and its neighbors and use it for downstream tasks such as classification.

\subsection{ODEs and Neural ODEs for Cardiac Dynamics and Dynamical Systems}

Concurrently, and initially driven by physiology research, ordinary differential equations (ODEs) have been used extensively to model cardiac dynamics at different scales (see e.g.~\citep{sundnesEfficientSolutionOrdinary2001}). Recently~\citep{grigorianHybridNeuralOrdinary2024}, these classical ODE models of cardiac dynamics have started to be hybridized with Neural ODEs (NODEs)~\citep{chen_neural_2018}, which typically use multilayer perceptrons in order to estimate the evolution of a system by learning its instantaneous rate of change. When combined with neural processes~\citep{garnelo_conditional_2018, norcliffe_neural_2020} we can obtain a distribution over the NODEs functions given a set of context points, allowing to learn the system dynamics from sparse and irregular measurements. However, these approaches often ignore spatial dependencies, limiting their ability to capture coordinated spatiotemporal patterns, such as the heart behavior.

\subsection{Latent Dynamics at the Intersection of Graphs and ODEs}

Graph neural networks (GNNs) have been applied to model dynamical systems in structured domains. By reformulating graph forward passes as solving ODE initial value problems, it is possible to take advantage of the relationships between graph nodes. For instance, GNNs can approximate diffusion processes~\citep{chamberlain2021grand} or diffusion-reaction models~\citep{thorpe_grand_2022}. In these approaches time is akin to depth and a pass through a GNN layer is a step in the ODE - layer depth functions as the number of steps to evolve the system~\citep{poli_graph_2019}. The final latent representation is used for the downstream tasks, such as classification or edge prediction between different nodes. Other approaches employ GNNs in order to synchronize components of coupled homogeneous dynamical systems~\citep{gupta_learning_2022}. Here each node in graph represents a full dynamical system, e.g. an oscillator, and latent neural ODEs model the evolution of each node’s state conditioned on its neighborhood via message passing. Finally, GNNs have also been used to emulate state-space models (SSMs) in the latent space and predict time-series evolution~\citep{zambon_graph_2023,look_cheap_2022,li_state_2024,behrouz_graph_2024,chouzenoux_sparse_2024}, providing a compact representation of systems governed by differential equations. These models use discrete-time latent states evolving in a Markovian fashion, enabling probabilistic inference and uncertainty modeling. In these cases, the graph structure is used to capture relationships between variables and estimate system transitions, functioning analogously to a Kalman filter in latent space~\citep{alippi_graph_2023}. 


In general, these methods highlight the natural link between graph-based representations and dynamical systems, which offer a powerful framework for studying spatiotemporal dynamics.

\subsection{Approach Overview and Contributions}

\begin{figure*}[!hbt]
    \centering
    \includegraphics[width=1.0\textwidth]{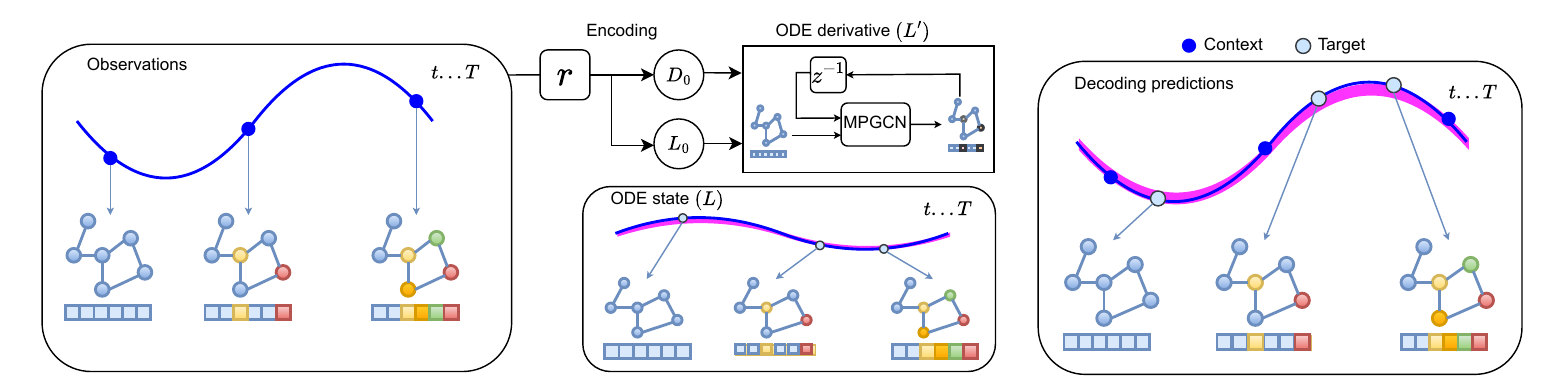}
    \caption{Schematic illustration of the Graph Neural ODE Process. Left: Observations from a time series; dots represent the context set. They are encoded and aggregated to form a representation $r$, which parametrizes the latent variables $L_{0}$ and $D_{0}$, initializing and conditioning the ODE respectively. Middle (top): Illustration of how graph neural networks estimate dynamics using the spatiotemporal structure. MPGCN: Multiplex Graph Convolutional Neural Network. Middle (bottom): A sample, $l_{0}$, from $L_{0}$. Each sample produces a plausible trajectory that controls the evolution of the latent state $L$. Right: Predictions at target times are made by decoding the latent state of the ODE $l_{i}$, and optionally together with the $D_{0}$ and the time $t_{i}$.}
    \label{fig:model_schematic}
\end{figure*}

In this work we aim at modeling the underlying heart’s spatiotemporal dynamics system governed by combining the strengths of GNNs, neural ODEs, and neural processes into a unified framework, shown schematically in Figure~\ref{fig:model_schematic}. We propose a graph-latent neural ODE Process that integrates a continuous-time latent dynamical system with a graph-based spatiotemporal representation. Specifically, nodes attributes represent system states, while edges encode spatial dependencies or causal relationships. This approach captures smooth, continuous evolution. Unlike classical NODEs, which rely on multilayer perceptrons to estimate velocity fields, we incorporate graph structures to take advantage of the heart’s spatiotemporal structure. Our architecture includes an encoder-decoder design, where a latent neural ODE models the cardiac cycle’s dynamics. The evolution of latent states is influenced by a multiplex graph, capturing spatial and temporal derivatives. This is conceptually similar to a Kalman Filter, where state transitions and observations evolve iteratively based on the previous state and the control parameters. The graph structure not only constrains the space of possible dynamics but also enables efficient learning by encoding domain knowledge using the connectivity matrix as prior. The state evolution can be decoded at each time-step and the output can be used for downstream tasks as a vector or scalar (node or graph level task). 

To evaluate our framework, we validated our model on three synthetic dynamical systems datasets to test its ability to capture system dynamics, interpolate within the observed time window, and extrapolate beyond the observed time window, then used the ACDC dataset~\citep{Bernard2018} and a subset of UK Biobank~\citep{sudlow_uk_2015} to reconstruct cardiac trajectories and classify clinical groups.

Our contributions can be summarized as follows:

\begin{itemize}
    \item Graph-Latent Neural ODE Process: We propose a novel state-space framework where latent dynamics are modeled as a neural ODE constrained by a multiplex graph structure.
    \item Unified spatiotemporal Representation: Our framework integrates spatial and temporal derivatives via graph-based methods to capture cardiac dynamics.
    \item Generalized Probabilistic Framework: We provide a flexible, probabilistic framework for spatiotemporal time series modeling, where graphs can change in topology and node representations over time.
    \item End-to-End Learning Procedure: We introduce a gradient-based method to learn latent graph states and their stochastic dynamics directly from data.
    \item Application to Cardiac Dynamics: We demonstrate the effectiveness of our approach in clinical data trajectory reconstruction tasks and classification tasks.
\end{itemize}

Empirical results show that our framework achieves state-of-the-art performance across multiple tasks, offering new insights into the latent dynamics of cardiac systems. The source code is available at \url{https://github.com/jbanusco/STGNP}.

\section{Methods}

Figure~\ref{fig:model_schematic} schematically illustrates the problem we aim to address.

We model the dynamic evolution of cardiac features over time as functions defined on a spatiotemporal graph. Let $X \in \mathbb{R}^{\mathcal{V}K}$ denote $K$ observed time-series over $\mathcal{V}$ cardiac regions or nodes, and $T = [t_{0}, ..., t_{n}]$ a set of time points. Our goal is to learn a model that, given a subset of time points (a context set), can interpolate and extrapolate the trajectories at target times leveraging the spatiotemporal graph structure.

To this end, starting from a spatiotemporal graph representation of cardiac images (Section \ref{s:methods:GraphRep}), we use a neural ODE process (Section~\ref{S:methods_NODEprocess}), where variables of the context set are encoded by different encoders (Section~\ref{s:methods:encoders}), the latent space parameterized by a spatiotemporal graph convolutional neural network (Section~\ref{S:methods_STGCN}), and a decoder is used to map the latent ODE state into variables of the target set (Section~\ref{s:methods:decoder}).

\subsection{Spatiotemporal Graph Representation}
\label{s:methods:GraphRep}

Generally, we model structured dynamical systems as \textit{spatiotemporal graphs}, where nodes $\mathcal{V}$ represent entities or regions of interest, and edges capture interactions across space and time. At each time point $t$, a graph layer $P_t$ encodes spatial relationships via intra-layer edges $\mathcal{E}_s$, while inter-layer edges $\mathcal{E}_t$ connect corresponding nodes across consecutive layers $P_t$ and $P_{t+1}$ to model temporal evolution. This multiplex formulation accommodates arbitrary node features and edge attributes, enabling the representation of complex, time-varying systems.

In the context of cardiac modeling, each plane $P_t$ corresponds to an image frame, with nodes representing anatomical regions and edges capturing spatial or temporal dependencies. This structure captures both the geometry and dynamics of the heart. Details of the graph construction process are provided in Section~\ref{s:methods:cardiacProcessing:graphCreation}.

\subsection{Neural ODE Process}
\label{S:methods_NODEprocess}

\begin{figure}[!htb]
	\centering
	\label{subfig:graphical_model}{\includegraphics[width=.5\textwidth]{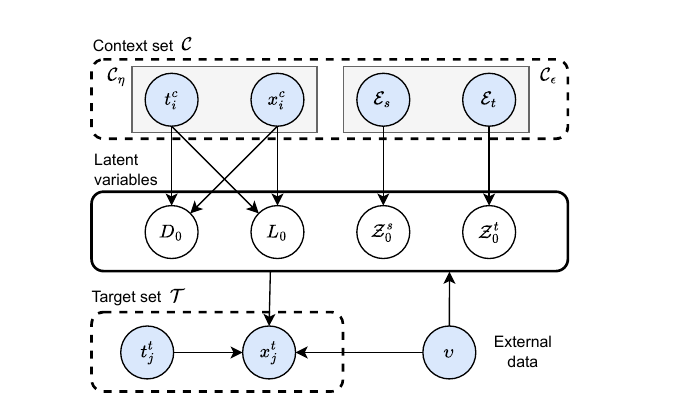}}
	\caption{Probabilistic graphical model of the latent neural ODE process. Circles represent random variables (blue: observed, white: latent). Rectangles denote graph-level quantities, solid latent and dashed observed. The system evolves from an initial latent state $l_{0}$ and control variable $d_{0}$ both inferred from a context set of node-level observations, $\mathcal{C}_{\eta}$. Spatial, $z^{s}_{0}$, and temporal, $z^{t}_{0}$, edge latents are inferred from the context edge-level attributes, $\mathcal{C}_{\epsilon}$. These variables parameterize the latent ODE, which generates trajectories $l_{t}$ that are decoded into node-level observable outputs $x_{t}$ at arbitrary times in the target set, $\mathcal{T}$. External information $\upsilon$ (e.g., subject-level covariates) is used to condition the encoding of all latent variables and the decoding.}
	\label{fig:graphical_model}
\end{figure}

Our model approximates functions over time using context observations to infer a distribution over latent neural ODE trajectories. The generative process, shown as a graphical model in Figure~\ref{fig:graphical_model}, involves inferring a distribution over the initial latent state $L_0$ and control variable $D_{0}$, which govern the trajectory of the system.

We adopt a probabilistic viewpoint inspired by neural processes~\citep{norcliffe_neural_2020}, where we do not learn a single deterministic function but rather a distribution over functions that map time to the observed time-series, $\mathcal{F}: T \rightarrow X$. We assume $\mathcal{F}$ to be the distribution over the function space, $\mathcal{D}$, induced by another distribution $\mathcal{D}'$ over some underlying dynamics, the latent space, that govern the observed time-series. 

Given a specific instance $f$ of $\mathcal{F}$, let the context set be $\mathcal{C} = \{(t_{i}, x_{i})\}^{N_{c}}_{i=1}$, where $x_{i} \in \mathbb{R}^{\mathcal{V}K}$ are node features observed at time $t_{i}$, and a target set $\mathcal{T} = \{(t_{j}, x_{j})\}^{N_{t}}_{j=1}$. The target set is a superset of the context set and we have $\mathcal{C} \in \mathcal{T}$. We assume the initial time and observation ($t_{0}, x_{0}$) are always observed and included in the context set, which in our case corresponds to the end-diastole frame. 
During training, the goal is to predict the values $x_{j}$ that $f$ takes at a set of target times. Thus, from a set of time-series sampled from $\mathcal{F}$ we aim to estimate the underlying distribution over the dynamics, $\mathcal{D}'$, as well as the induced distribution over the functions, $\mathcal{D}$. For notation clarity, we drop the explicit time dependency and use $l_{0}:=l(t_{0})$.

We model the latent dynamics using a neural ODE, where the initial state $l_{0}$ and control parameters $d_{0}$ are inferred from the context. Thus, given an initial time $(t_{0})$, and target time $(t_{i})$, the model predicts the corresponding state $(x_{i})$ by performing the following encoding, integration and decoding operations.

\begin{align}
    &l_{0} \sim q(L_{0} \mid \mathcal{C, \upsilon}), \\
    &d_{0} \sim q(D_{0} \mid \mathcal{C}, \upsilon), \\    
    &l_{i} = l_{0} + \int_{t_{0}}^{t_{i}} f_\theta(l_{t}, d_{t}, t)dt, \\
    &d_{i+1} = f_\delta(d_{i}, l_{i}), \\
    &x_{i} = g(l_{i}, \upsilon),
\end{align}

where $\upsilon$ represents potentially available global information. $q(L_{0} \mid \mathcal{C}, \upsilon)$ and $q(D_{0} \mid \mathcal{C}, \upsilon)$ are learned variational posteriors over the initial latent state and control parameters, respectively. $f_{\theta}(l, d, t)$, typically modeled as a multilayer perceptron  (MLP), parametrizes the ODE derivative $(\dot{l})$. $f_{\delta}$, a function updating the control parameters, and $g$, a function acting as decoder, can also be parametrised using a MLP. In order to take advantage of the spatiotemporal nature of our data we decided to parametrise $f_{\theta}(l, d, t)$ as a GNN able to handle spatiotemporal graphs (see Section~\ref{s:methods:GraphRep}). Thus, in our case the inputs include both node-level attributes (e.g., regional measurements) and edge-level attributes (e.g., spatial distances, anatomical connections). To account for this, we split the context set $\mathcal{C}$ into two disjoint subsets: $\mathcal{C}_{\eta}$ for node observations and $\mathcal{C}_{\epsilon}$ for edge information.

In this formulation, the function $f$ is defined implicitly via the latent initial condition $L_{0}$ and control variable $D_{0}$, which govern the trajectory in latent space. This enables the model to represent a distribution over possible trajectories conditioned on sparse context observations and generate coherent temporal predictions aligned with the graph structure. Predictions at any target time $t_{j}$ are made by integrating the ODE forward and decoding the result: $x_{j} \sim p(x_{j} \mid g(l_{j}, \upsilon))$.

\subsubsection{Encoders}
\label{s:methods:encoders}

\begin{figure}[!htb]    
	\centering
    \label{subfig:network}{\includegraphics[width=.5\textwidth,keepaspectratio]{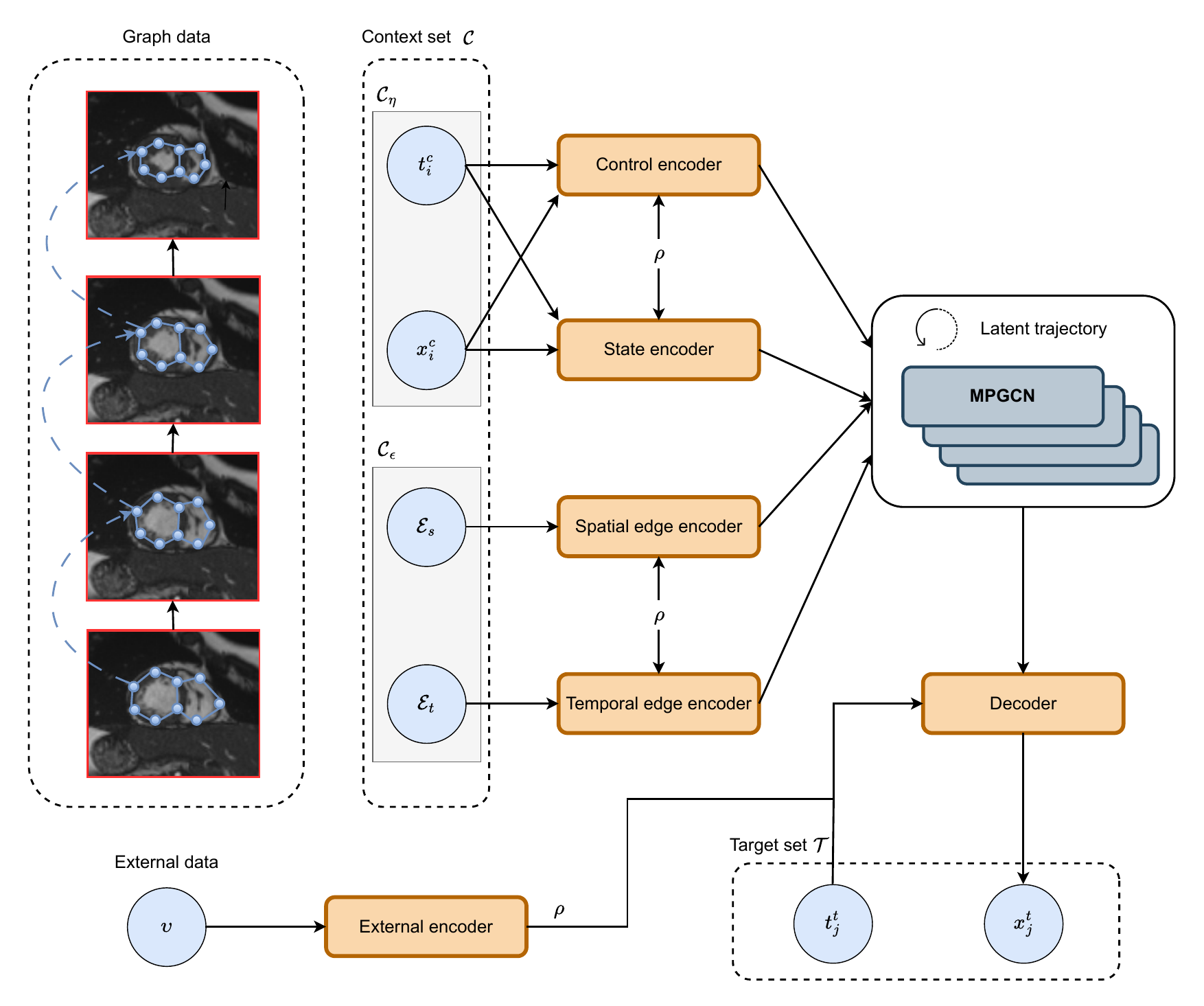}}
	\caption{Overview of the model architecture. Context node observations $\mathcal{C}_{\eta}$ and edge observations $\mathcal{C}_{\epsilon}$ are encoded into latent variables: the initial state $l_{0}$, control vector $d_{0}$, and edge latents $z^{s}_{0}$,$z^{t}_{0}$. These parameters initialize a latent neural ODE whose derivative is parameterized by a spatiotemporal graph neural network. The latent state is integrated over time and decoded to predict system observations $x_{t}$ at arbitrary target times. $\rho$ is the output of the external encoder $f_{\psi}$, which is used to condition the decoding and the encoding of the node and edge-level attributes. CMR images reproduced by kind permission of UK Biobank ©.}
	\label{fig:network}
\end{figure}

We define four distinct encoders for the context set: a control encoder, a state encoder, a spatial edge encoder, and a temporal edge encoder. Additionaly, we define an external encoder, $f_{\psi}$, for external data.
Their relationship with the variables of the context set and their position in the overall network is depicted in Figure~\ref{fig:network}. Note that $\rho=f_{\psi}(\upsilon)$, is a deterministic encoding of the external covariates. Thus, in what follows we use $\upsilon$ to refer to either the raw or encoded covariates, with a slight abuse of notation for clarity.


Given a node-level context set $\mathcal{C_{\eta}} = \{(t_i, x_i)\}$ and additional global graph-level information $\upsilon$, we infer two node-level latent variables $l_{0} \sim q_{l}\left( L_{0} \,|\, \mathcal{C_{\eta}}, \upsilon \right)$ and $d_{0} \sim q_{D}\left( D_{0}\,|\,\mathcal{C_{\eta}}, \upsilon \right)$, representing the initial state and the global control of an ODE. Moreover we also encode the initial spatial and temporal edges using the edge-level context set, $\mathcal{C}_{\epsilon
}$. Thus, the spatiotemporal graph at $t_{0}$ and the connection to the next time instance $t_{1}$ are given by $z_{0}^{s} \sim q_{Z_{0}^{s}}\left( Z_{0}^{s}\,|\,\mathcal{C}_{\epsilon}, \upsilon \right)$ and $z_{0}^{t} \sim q_{Z_{0}^{t}}\left( Z_{0}^{t}\,|\,\mathcal{C}_{\epsilon},f_{\psi}(\upsilon) \right)$. 

To parametrize the distribution of the control latent variable $D_{0}$ the control encoder produces a representation $r_{i} = f_{D}(t_{i}, x_{i})$ for each context pair $(t_{i}, x_{i})$, which are aggregated via mean pooling to preserve order invariances into a global representation $r$. The distribution of $L_{0}$ is parametrized by the state encoder as a function of the initial state $x_{0}$, although it could also be parametrised as a function of the entire context using the same $r$. Thus,
$d_{0} \sim q_{D_{0}}(D_{0} \,|\, \mathcal{C}_{\eta}, \upsilon) = \mathcal{N}(d_{0} \,|\, \mu_{D_{0}}(r), diag(\sigma_{D_{0}}(r))$, and $l_{0} \sim
q_{L_{0}}(L_{0} \,|\, \mathcal{C}_{\eta},\upsilon) = \mathcal{N}(l_{0} \,|\, \mu_{L_{0}}(x_{0}), diag(\sigma_{L_{0}}(x_{0}))$. For the edges we also obtain a latent representation $(r_{\mathcal{E}})$ after aggregating the edges from the context points. The parametrized distributions are defined as $z_{0}^{s} \sim q_{Z_{0}^{s}}(Z^{s}_{0}| \mathcal{C}_{\epsilon}, \upsilon) = \mathcal{N}(Z_{0}^{s} \,|\, \mu_{Z_{0}^{s}}(r_{\mathcal{E}}), diag(\sigma_{Z_{0}^{s}}(r_{\mathcal{E}}))$ and $z_{0}^{t} \sim q_{Z_{0}^{t}}(Z_{0}^{t} \,|\, \mathcal{C}_{\epsilon}, \upsilon) = \mathcal{N}(Z_{0}^{t} \,|\, \mu_{Z_{0}^{t}}(r_{\mathcal{E}}), diag(\sigma_{Z_{0}^{t}}(r_{\mathcal{E}}))$.

\subsubsection{Latent ODE Spatiotemporal Graph Convolutional Network}
\label{S:methods_STGCN}

\begin{figure}[!htb]
	\centering
\includegraphics[width=.5\textwidth]{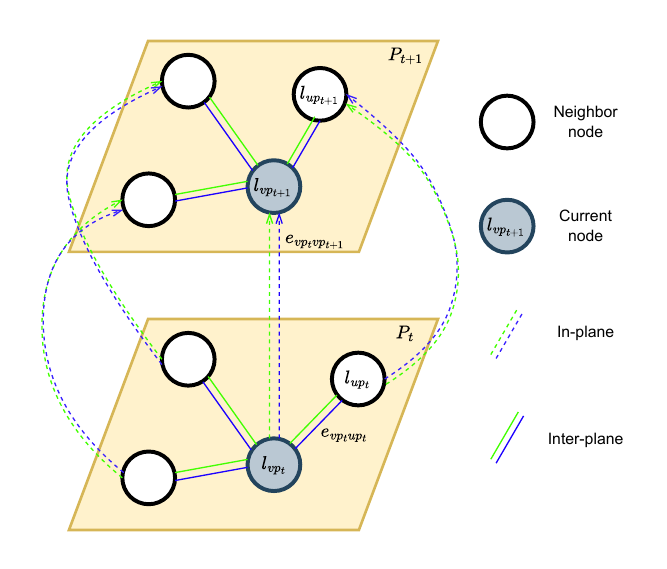}
	\caption{MPGCN layer blocks. Planes represent two consecutive time-instants, in our case cardiac frames from the 4D image. Line colors indicate different edge features. Dashed lines show inter-plane (temporal) edges. Solid lines show in-plane (spatial) edges.}
	\label{fig:GCNPlanes}
\end{figure}




In contrast to classical Neural ODEs~\citep{norcliffe2021neural} that use MLPs to parameterize state evolution, our method learns dynamics using a GNN that encodes spatial and temporal dependencies, making the estimation of the ODE derivative by $f_{\theta}$ structure-aware. Since in our graph representation we have spatial and temporal edges, we have two messages-passing streams—spatial ($m^{s}$) and temporal ($m^{t}$)—that operate simultaneously on intra-frame (spatial) and inter-frame (temporal) edges. Figure~\ref{fig:GCNPlanes} provides an illustration. The spatial message at node $v$ at time $t$ is defined as:

\begin{equation}
    m^{s}_{vt} = \mean_{u \in \mathcal{N}_{s}(v)} \left( \sum^{k_{es}}_{f} \frac{l_{ut} \cdot e^{fs}_{uv}}{c^{s}_{uv}} \right),
\end{equation}

while the temporal message $m^{t}_{v}$ is defined as

\begin{equation}
    m^{t}_{vt} = \mean_{u \in \mathcal{N}_{t}(v)} \left( \sum^{k_{et}}_{f} \frac{l_{ut} \cdot e^{ft}_{uv}}{c^{t}_{uv}} \right),
\end{equation}

where $l_{ut}$ represents the latent state of neighbor node $u$ at time $t$, $\mathcal{N}_s(v)$ and $\mathcal{N}_t(v)$ are spatial and temporal neighborhoods, and $c_{uv}^s$, $c_{uv}^t$ are normalization factors based on the in-node degree, and $f$ iterates over the edge features. The received messages are used to estimate a temporal and a spatial derivative terms, as: $\dot{l}^{s} = l_{t} - m^{s}_{t}$ and $\dot{l}^{t} = m^{t}_{t} - l_{t}$. These derivatives are fed together with the current state, $l_{t}$, and control parameters $d_{t}$ to a function $f_{\theta}$ in order to estimate the final latent derivative. Thus, the full update is given by:
\begin{equation}
\dot{l} = f_{\theta}(l_{t}, \dot{l}^{s}, \dot{l}^{t}, d(t))
\end{equation}

To enable dynamic latent control the weights of the edges, $z^{t}_{t}$, $z^{s}_{t}$, and the control parameters, $d_{t}$, evolve in an autoregressive way:

\begin{align}
    &d_{t} = f_{\delta}(d_{t-1}, l_{t-1}), \\
    &z^{s}_{t} = f_{s}(z^{s}_{t-1}, d_{t}), \\
    &z^{t}_{t} = f_{t}(z^{t}_{t-1}, d_{t}),
\end{align}

where $f_{\delta}$, $f_{s}$ and $f_{t}$ are MLPs. The GNN was implemented using deep graph library (DGL)~\citep{Wang2019} with Pytorch~\citep{Paszke2019} backend. 

\subsubsection{Decoder}
\label{s:methods:decoder}

To obtain the prediction at target time $t_{i}$ the decoder maps the latent ODE state $l_{i}$ into the observation space. Assuming observations are noisy, we model the output as a Gaussian distribution. For a given sample $l_{i}$ the decoder $g$ produces a distribution over $x_{i} \sim p(x_{i} \,|\, g(l_{i}, t_{i},  \upsilon)) = \mathcal{N}(x_{i} | \mu_{i}, \sigma_{i}^{2}I)$, where both $\mu_{i}$ and $\sigma_{i}^{2}$ are produced by the decoder $g$, an MLP. If contextual features are available, they can also be concatenated to $l_{i}$ as decoder inputs . The decoder thus maps latent dynamics into node-specific feature trajectories, allowing the system to reconstruct full spatiotemporal profiles from sparse context data.


\subsection{Learning and Inference}

Given the described structure, the generative process is given by:

\begin{align}
&p(x_{1\colon n} , l_{0}, d_{0}, z_{0}^{s}, z_{0}^{t} \,|\, t_{1\colon n} , \mathcal{C}, \upsilon) = \\ \notag
&p(l_{0} \,|\, \mathcal{C}_{\eta}, \upsilon)p(d_{0} \,|\, \mathcal{C}_{\eta}, \upsilon)p(z_{0}^{s} \,|\, \mathcal{C}_{\epsilon}, v)p(z_{0}^{t} \,|\, \mathcal{C}_{\epsilon}, \upsilon) \prod_{i = 1}^{n}p(x_{i} \,|\, g(l_{i},t_{i},\upsilon)),
\end{align}

where $x_{1 \colon n}$ denotes the sequence $(x_{1}, ..., x_{n})$, and likewise for $t_{1 \colon n}$. 

Since the true posterior is intractable due to the highly non-linear generative process we trained the model using variational inference. The lower-bound on the probability of the target value given the known context is given by:

\begin{align}
    &\mathcal{L}(l,d,z^{s},z^{t};x,
    \upsilon) = E_{q(l_{0},d_{0},z_{0}^{s},z_{0}^{t}|t,x,\upsilon)} \bigg[ \sum_{i \in \mathcal{T}} \log p(x_{i} | l_{0}, d_{0},z_{0}^{s},z_{0}^{t}, t_{i}, \upsilon) \notag \\
    &+ \log \frac{q_L(l_{0} | \mathcal{C}_{\eta}, \upsilon)}{q_L(l_{0} | \mathcal{T}_{\eta}, \upsilon)}
    + \log \frac{q_D(d_{0} | \mathcal{C}_{\eta}, \upsilon)}{q_D(d_{0} | \mathcal{T}_{\eta}, \upsilon)} \notag \\
    &+ \log \frac{q_{Z_{0}^{s}}(Z_{0}^{s} | \mathcal{C}_{\epsilon}, \upsilon)}{q_{Z_{0}^{s}}(Z_{0}^{s} | \mathcal{T}_{\epsilon}, \upsilon)}
    + \log \frac{q_{Z_{0}^{t}}(Z_{0}^{t} | \mathcal{C}_{\epsilon}, \upsilon)}{q_{Z_{0}^{t}}(Z_{0}^{t} | \mathcal{T}_{\epsilon}, \upsilon)}
    \bigg],
\end{align}

where $\mathcal{C}_{\eta}$, $\mathcal{C}_{\epsilon}$ indicate elements node and edge attributes in the context set, and $\mathcal{T}_{\eta}$, $\mathcal{T}_{\epsilon}$ indicate node and edge attributes in the target set. 


To regularize training and control overfitting, we include additional $\ell^2$ loss term based on the magnitude of the latent trajectory, as well as a $\ell^1$ on the value of the temporal and spatial edges. The final loss becomes:

\begin{equation}
    \mathcal{L}_{total} = \beta_{1}\mathcal{L}(l,d,z^{s},z^{t};x,
    \upsilon) + \beta_{2}||l||_{2}^{2} + \beta_{3}(||z^{s}||_{1}+||z^{t}||_{1}),
\end{equation}

where $\beta_{1}$, $\beta_{2}$, $\beta_{3}$ are weighting hyperparameters. Initial node features and global features were z-score normalised based on the training set samples. The edge features were not normalised. 

All experiments were trained using the Adam optimizer~\citep{Kingma14} with a learning rate of $0.01$. The learning rate was halved every $100$ epochs for a total of $300$ epochs. The ODE was solved using the Runge-Kutta (RK4) method from the \texttt{torchdiffeq} library~\citep{torchdiffeq}. Z-score normalization was applied to all inputs based on the training set. The loss hyperparameters $\beta_{1}$, $\beta_{2}$, $\beta_{3}$ were found using Optuna~\citep{optuna_2019} and as hyperparameter score we used the mean absolute error (MAE) of the validation set.


\subsection{Prediction Tasks}
\subsubsection{Reconstruction}

In the reconstruction task, the model is given a sparse set of context observations at both the node and edge levels. The objective is to predict node-level attributes at target time points \textit{within} the temporal support of the context set. Performance is evaluated using mean squared error (MSE), mean absolute error (MAE), mean squared percentage  error (MSPE), and mean absolute percentage error (MAPE) between predicted and ground truth values.

\subsubsection{Extrapolation}

In the extrapolation task, the model is again provided with a sparse set of context observations at both the node and edge levels. However, predictions are made at target time points \textit{outside} the range of the observed context set, testing the model's ability to generalize beyond the training window. We report MSE, MAE, MSPE, and MAPE as evaluation metrics. For synthetic data, we additionally include results from baseline forecasting methods: last observation carried forward (LOCF)—also known as the naive method~\citep{hyndman_forecasting_nodate} (Naive 1 in the Makridakis forecasting competitions~\citep{makridakis_m4_2020})—as well as autoregressive order 1 (AR(1)) and autoregressive integrated moving average (ARIMA). AR(1) and ARIMA models were trained using the \textit{pmdarima} package~\citep{pmdarima}, with ARIMA parameters selected via constrained auto-tuning $(p \leq 3,\, d \leq 1,\, q \leq 2)$ applied independently to each sample, node, and feature.

\subsubsection{Cardiac Classification Tasks}

To evaluate the discriminative ability of the learned representations, we explored the use of the latent parameters inferred from the context, the initial condition $l_{0}$ and the control dynamics $d_{0}$, which provide a compact summary of the underlying cardiac motion. A classifier is trained on the concatenated latent representations:

\begin{equation}
y = f_{\phi}(l_{0}, d_{0}),
\end{equation}

The latent states and control parameters are defined per region, thus the input is flattened into a vector and provided to the classifier $f_{\phi}$, which is parameterized as a Random Forest. We also compared this compact representation to a full representation of the cardiac data, comprising all node and edge features across all time frames, using Random Forest, XGBoost, or a nearest-neighbour approach. All classifier hyperparameters are tuned via cross-validation to ensure competitiveness in each scenario. The hyperparameter search space for all machine learning algorithms is defined in Appendix~\ref{s:hyperparametersOfClassifiers}.

\section{Data and Processing}

\subsection{Synthetic Data}

We use three synthetic dynamical systems of increasing complexity: coupled pendulums, Lorenz attractors, and Kuramoto oscillators. These benchmarks allow us to assess the model's ability to represent, infer, and extrapolate latent dynamics from sparse context observations.

\subsubsection{Coupled Pendulum Data}

The dynamics of two coupled pendulums are governed by:

\begin{equation}
\begin{aligned}
\ddot{\theta}_1 &= \frac{\sin(\theta_1)\big(m_1 l_1 \dot{\theta}_1^2 - g - k l_1\big) + k l_2 \sin(\theta_2)}{m_1 l_1 \cos(\theta_1)}, \\
\ddot{\theta}_2 &= \frac{\sin(\theta_2)\big(m_2 l_2 \dot{\theta}_2^2 - g - k l_2\big) + k l_1 \sin(\theta_1)}{m_2 l_2 \cos(\theta_2)},
\end{aligned}
\label{eq:coupled_pendulum_dynamics}
\end{equation}

where $\theta_{i}$ defines the angle and $\dot{\theta}_{i}$ the angular velocity of the $i^{\text{th}}$ pendulum. $m_{i}$ and $l_{i}$ are the mass and length, chosen as $1\,\mathrm{kg}$ and $1.5\,\mathrm{m}$, respectively. $k$ is the spring constant of the coupling string, set to $2$, and $g$ is gravitational acceleration ($9.81\,\mathrm{m/s^2}$). The state of the coupled system is thus defined by $[\theta_{1}, \dot{\theta}_{1}, \theta_{2}, \dot{\theta}_{2}]$. We simulated trajectories with initial angles \(\theta_i \in [-\pi/2, \pi/2]\) and angular velocities \(\dot{\theta}_i \in [-1, 1]\), sampled at $0.1\,\mathrm{s}$ intervals over $10\,\mathrm{s}$. The graph structure includes 2 nodes (one per pendulum), with a fully connected spatial adjacency matrix (no self-loops) and a temporal diagonal matrix.

\subsubsection{Lorenz Attractor Data}

The Lorenz attractor is a three-dimensional system that exhibits chaotic behavior, and is described by:

\begin{equation}
\begin{aligned}
    \dot{x} &= \sigma(y - x), \\
    \dot{y} &= x(\rho - z), \\
    \dot{z} &= xy - \beta z,
\end{aligned}
\label{eq:lorenz_attractor}
\end{equation}
with parameters \(\sigma = 10\), \(\rho = 28\), and \(\beta = 8/3\)~\citep{lorenz_deterministic_1963}.

We simulate a system with $3$ Lorenz nodes, i.e., each node is a Lorenz system. Trajectories last $2.5\,\mathrm{s}$ with a timestep of $0.05\,\mathrm{s}$. A coupling term is added to modulate interactions between nodes, based on a random matrix scaled by a coupling strength of $0.01$ and multiplied by the system's state. The graph structure for learning is defined by a fully connected spatial matrix (no self-loops) and a temporal diagonal matrix.

\subsubsection{Kuramoto System Data}

The Kuramoto model~\citep{kuramoto_chemical_2003} describes the behavior of large sets of coupled oscillators, with applications in fields such as neuroscience~\citep{escrichs_unifying_2022, rodrigues_kuramoto_2016}, power systems~\citep{guo_overviews_2021}, and vehicle coordination~\citep{tang_synchronization_2014}. The dynamics are given by:

\begin{equation}
    \dot{\theta}_i = \omega_{i} + \sum_{j=1}^{N}A_{ij}sin(\theta_{j}-\theta_{i}),
\end{equation}\label{eq:kuramoto}

where $\theta_j$ is the phase of node $j$, $\theta_i$ the phase of node $i$, $\omega_i$ its natural frequency, and $A_{ij}$ the coupling matrix. We generate systems with $10$ nodes, each connected to $5$ others using the Barabási-Albert process~\citep{barabasiEmergenceScalingRandom1999}. For learning, the graph structure uses a fully connected spatial matrix (no self-loops) and a diagonal temporal matrix.

\subsection{Cardiac MRI Datasets}

We used two cardiac magnetic resonance imaging  (MRI) datasets: a subset of the UK Biobank (UKB) and the ACDC challenge dataset. The ACDC dataset~\citep{Bernard2018} comprises 150 subjects categorized into five equally distributed groups: one healthy cohort (NOR) and four pathology classes: myocardial infarction (MINF), dilated cardiomyopathy (DCM), hypertrophic cardiomyopathy (HCM), and abnormal right ventricle (ARV). Each patient is represented by a 4D volume (x-y-z-t) of short-axis slices covering the base to the apex of the ventricles, with variable resolution ($\sim$1.5~mm$^2$ in-plane and 5–10~mm slice thickness). Images were acquired in routine clinical settings, leading to heterogeneous quality.

The UK Biobank~\citep{sudlow_uk_2015} is a large-scale population cohort study with over 500,000 participants aged 40–69. It includes genetic data, biospecimens (blood, urine, and saliva), clinical records based on ICD-10 codes, lifestyle information, and cardiac MRI acquired across multiple centers. 

To test our approach on a challenging classification task, we selected $263$ patients with an atrial fibrillation (AFib) diagnosis (ICD-10 code I48) prior to MR image acquisition, and 263 controls matched for age, sex, and body surface area (BSA), without atrial fibrillation or other cardiovascular pathologies. The list of excluded conditions is available in Appendix Table~\ref{tab:ICD_codes}. Atrial fibrillation was chosen due to its clinical relevance as the most common arrhythmia and a risk factor for numerous conditions, including heart failure and stroke. Importantly, it has recently been shown to contribute to left ventricular dysfunction~\citep{pabelEffectsAtrialFibrillation2022}, and therefore its effects should be reflected in cardiac dynamics and captured by our model.

For comparability with ACDC, we used only short-axis cine MRI, acquired with $\sim$1~mm$^2$ in-plane resolution and $\sim$8~mm slice thickness under a standardized protocol across centers using the same scanner model. Since only height and weight were available in ACDC, BSA was derived in both datasets using Mosteller’s formula~\citep{Mosteller1987}: $BSA = \left(\frac{\text{Weight [kg]} \times \text{Height [cm]}}{3600}\right)^{\frac{1}{2}}$. Height, weight, BSA, as well as left and right ventricular ejection fractions and stroke volume index, are used as global features $\upsilon$.

\subsection{Cardiac image processing}
\label{s:methods:cardiacProcessing}

\begin{figure}[!htb]
	\centering
	\label{subfig:data_processing}
    \includegraphics[width=.5\textwidth]{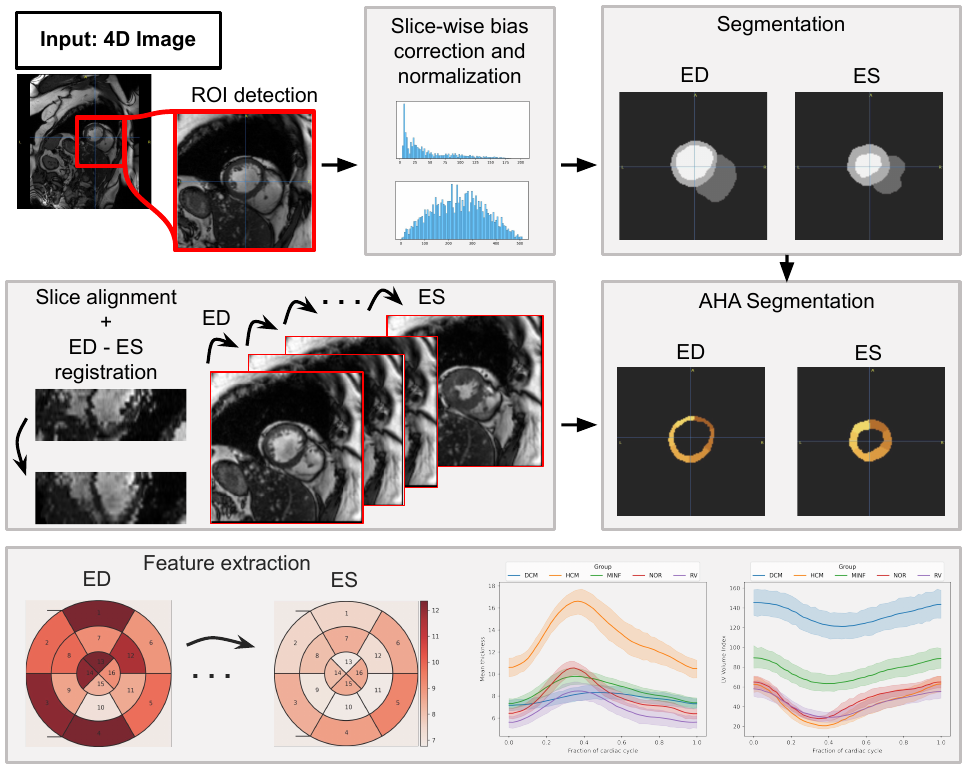}
	\caption{Cardiac image processing pipeline. The input is a 4D image composed of 2D short-axis slices. We apply slice-wise bias field correction and normalization, and fine-tune a pretrained model~\citep{Bai2018} to segment the 4D volume. Next, we align the slices of the 4D volume to correct breath-motion artifacts and obtain the diffeomorphism from ED to any cycle phase by means of frame-to-frame composition. The LV myocardium at ED is divided in 16 AHA segments, while the RV myocardium is divided in 9 regions. The composed frame-to-frame diffeomorphism is then used to propagate the region labels across the cardiac cycle. For each segment we extract features related to myocardium thickness, volume and intensity distribution. ED: end-diastole, ES: end-systole, AHA: american heart association. CMR images reproduced by kind permission of UK Biobank ©.}
	\label{fig:data_processing}
\end{figure}

As illustrated in Figure~\ref{fig:data_processing}, we processed each 4D volume using a standardized pipeline. Data were partially organized using the Medical Imaging Data Structure~\citep{seroussi_beyond_2022}, an extension of the Brain Imaging Data Structure~\citep{gorgolewski_brain_2016}. UK Biobank DICOM images were converted to NIfTI format using \texttt{pydicom}~\citep{mason2011t}.

\subsubsection{ROI and Segmentation}

We defined a region of interest (ROI) by detecting the approximate center of the LV using a circular Hough transform~\citep{khened_fully_2019,duda_use_1972}, then extracted a 128×128 patch centered on it. We applied slice-wise N4 bias field correction~\citep{tustison_n4itk_2010} and normalized intensities using $I_{\text{norm}} = (I - I_{1}) / (I_{1} - I_{99})$, where $I$ corresponds to voxel intensity, and $I_{1}$ and $I_{99}$ are the 1st and 99th percentiles, respectively.

To segment the images in the ACDC dataset, we fine-tuned a pretrained model~\citep{Bai2018} using the ED and ES segmentation maps available in the ACDC training set. For the UK Biobank images, we used the original pretrained model without fine-tuning, since it was trained on UK Biobank data. We also applied a series of 2D+t morphological corrections to address temporal inconsistencies and segmentation errors. These included removing small disconnected components ($<$20 voxels), closing 2D holes in each frame (radius 3), and extending hole filling over time (2D+t) using a radius of 3 voxels and $\pm$1 frame. We further dilated the LV (radius 3) and ensured it was surrounded by myocardium, verified that dilated LV and RV masks did not overlap (i.e., myocardium separation existed), and closed holes inside blood pools. From the corrected segmentation maps, we computed the volumes of the left ventricle (LV), right ventricle (RV), and LV myocardium mass, as well as ejection fraction (EF) for each region.

\subsubsection{Thickness and Myocardium division}

We defined local coordinate axes per heart: the primary axis runs from the LV center of mass to the RV center of mass ; the secondary axis extends from the base to the apex; and the third is their cross-product. Using these axes, we divided the LV myocardium at ED into 16 segments according to the American Heart Association (AHA) model~\citep{Cerqueira2002}. The apex segment was excluded due to possible field-of-view limitations. ED was chosen as reference because it represents the end of the filling phase—when the ventricles are maximally dilated—and is the standard reference frame for AHA division.

To estimate RV wall thickness, we approximated the RV myocardium by dilating its mask uniformly by $2\,\text{mm}$~\citep{plaisier_image_2012}. We then partitioned the RV into 9 regions (3 each at basal, mid, and apical levels) using k-means clustering to define three equal-length segments per level.

Finally, we estimated myocardial thickness for both ventricles at each time frame using the Ezzi method~\citep{yezzi_eulerian_2003}, implemented in the \texttt{pyezzi} package~\citep{cedilnik_pyezzi_nodate}.

\subsubsection{Slice alignment and registration}

To correct for breath-hold misalignment, we re-centered each slice by applying 2D rigid transforms between consecutive slices, using a mid-ventricular slice as reference. We then disassembled the 4D volume and registered consecutive frames slice-wise using the SyN algorithm~\citep{Avants2008} from the ANTs toolkit~\citep{tustison_antsx_2021}. This 2D approach aligns with slice-wise acquisition and enables recovery of diffeomorphisms from ED to any phase of the cycle by composing the inter-frame deformations. We further refined the registration using the method from~\citep{ogier_individual_2017}, averaging forward and backward deformations to obtain a smoothed transformation field. The final diffeomorphic maps were used to propagate anatomical segment labels (AHA and RV regions) from ED to the full cardiac cycle.

\subsubsection{Graph creation}
\label{s:methods:cardiacProcessing:graphCreation}

Each AHA and RV myocardium segment was considered a node in our graph. For each region, we extracted features related to volume, thickness, and intensity. Specifically, we computed the mean, median, interquartile range, and lower and upper quartiles. Volumes were normalized by body surface area (BSA).

To define edge features, we computed the Wasserstein distance between voxel intensity distributions of regions within a frame and between frames. We also calculated distances between centers of mass using the heart-specific local coordinate axes. Table~\ref{tab:available_data} lists all features, their anatomical location, and whether they are node or edge attributes.

We generated graphs in two versions: (i) anatomically adjacent segments only (Anat), and (ii) fully connected graphs (Full). Graphs were stored in \texttt{.bim} format for efficient downstream loading. The entire pipeline is available as docker images and Nextflow~\citep{ditommasoNextflowEnablesReproducible2017} scripts for reproducible batch processing.

\begin{table}[!htb]
\setlength\tabcolsep{3pt}
\caption{Edge and node features. For intensity and thickness, mean, median, and quartiles were extracted. CM: center of mass, Myo: myocardium, LV: left ventricle, RV: right ventricle, AHA: american heart association.}
\label{tab:available_data}
\centering
\begin{tabular}{lccc}
\hlx{c{1-4}} 
\textbf{Feature} & \textbf{RV Myo} & \textbf{LV Myo (AHA)} & \textbf{Type} \\
\hlx{hvh}
Volume             & \checkmark & \checkmark & Node \\
Voxel intensity    & \checkmark & \checkmark & Node \\
Position (CM)      & \checkmark & \checkmark & Node \\
Thickness          & \checkmark & \checkmark & Node \\
Wasserstein dist.  & \checkmark & \checkmark & Edge \\ 
CM distance        & \checkmark & \checkmark & Edge \\
\hlx{h}
\end{tabular}
\end{table}

\section{Experimental results}

\subsection{Trajectory prediction on synthetic data}

Each synthetic dataset includes $500$ randomly initialized trajectories. We split $80\%$ for training, $20\%$ for validation, and $20\%$ for testing. One-third of the training time points are selected as context points, chosen with equal spacing and shared across all samples. For extrapolation, predictions are made beyond the training window using the learned dynamics for the same number of time steps. During training, the graph's spatial connectivity is fully connected (without self-loops), and its temporal connectivity is defined by a diagonal matrix.

To select hyperparameters for each setting, we ran $100$ Optuna~\citep{optuna_2019} trials, using the MAE on the validation set as the optimization objective. The selected hyperparameter values for each experiment are listed in Appendix Table~\ref{tab:hyperparams_multiplex_values}. For all settings, we used a fixed batch size of $30$ and an initial learning rate of $0.01$. A learning rate scheduler was used to reduce the learning rate by a factor of $2$ when the validation metric stopped improving, with a patience of $10$ and a minimum learning rate of $10^{-5}$.

\subsubsection{Coupled Pendulum Results}

\begin{figure}[!htb]
    \centering
    \begin{subfigure}[b]{0.5\textwidth}
        \includegraphics[width=\textwidth]{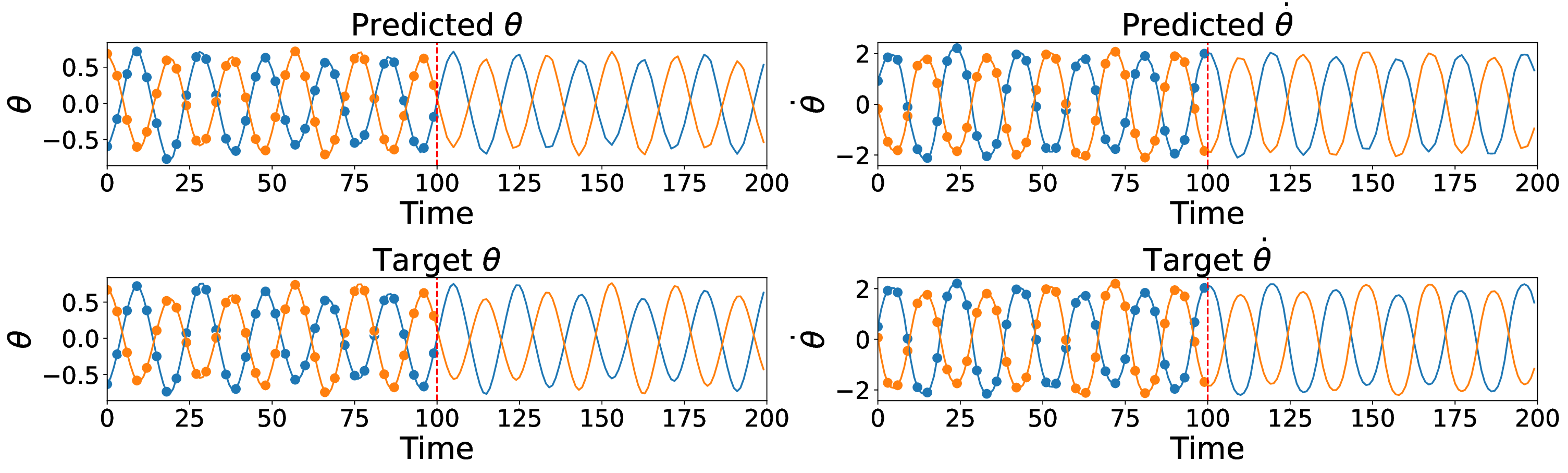}
        \caption{Training set.}
        \label{subfig:pendulum_train}
    \end{subfigure}
    \hfill
    \begin{subfigure}[b]{0.5\textwidth}
        \includegraphics[width=\textwidth]{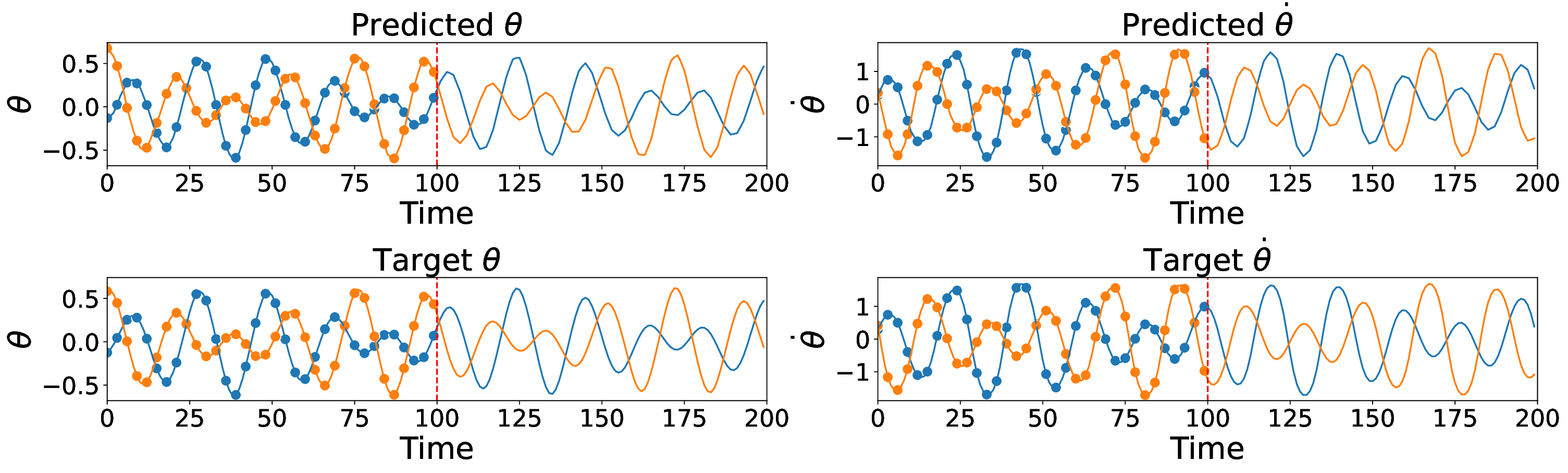}
        \caption{Testing set.}
        \label{subfig:pendulum_test}
    \end{subfigure}
    \caption{Example pendulum trajectories. a) Training set and b) Testing set. The vertical dashed line marks the end of the training window. Points beyond this line are extrapolated using the learned dynamics. Context points, shown as dots, span one-third of the training frames.}
\label{fig:pendulum}
\end{figure}

Figure~\ref{fig:pendulum} shows the predicted trajectories for a sample from the training set and another from the test set, while Table~\ref{tab:feature_errors_pendulum} reports the error metrics on the test set. Prediction error was lower within the training window (e.g., $\theta$ MAPE 6.56\%) and increased when extrapolating beyond it ($\theta$ MAPE 8.28\%). All forecasting baselines failed to capture the system's dynamics (e.g., $\theta$ ARIMA MAPE 402.04\%).

\begin{table}[!htb]
\centering
\setlength\tabcolsep{4pt}
\caption{Errors per feature in the coupled pendulum test set (N=100). MSE: mean squared error, MAE: mean absolute error, MSPE: mean squared percentage error, MAPE: mean absolute percentage error, IQR: inter-quartile range.}
\label{tab:feature_errors_pendulum}
\resizebox{\columnwidth}{!}{%
$\begin{tabular}{|c|c|c|c|c|c|c|}
\hline
\textbf{Method} & \textbf{Task} & \textbf{Feature} & \textbf{MSE $\pm$ IQR} & \textbf{MAE $\pm$ IQR} & \textbf{MSPE [$\%$]} & \textbf{MAPE [$\%$]} \\
\hline
\multirow{4}{*}{\centering\arraybackslash Ours} & \multirow{2}{*}{\centering\arraybackslash Reconstruction}   & $\theta$       & 0.0002 $\pm$ 0.0006 & 0.012 $\pm$ 0.019 & 0.43\% & 6.56\% \\
 &   & $\dot{\theta}$ & 0.0015 $\pm$ 0.0051 & 0.039 $\pm$ 0.057 & 0.34\% & 5.81\% \\
 \cline{2-7}
 & \multirow{2}{*}{\centering\arraybackslash Extrapolation}  & $\theta$       & 0.0003 $\pm$ 0.0009 & 0.018 $\pm$ 0.023 & 0.69\% & 8.28\% \\
 &   & $\dot{\theta}$ & 0.0032 $\pm$ 0.013 & 0.056 $\pm$ 0.091 & 0.89\% & 9.44\% \\
\hline
\multirow{2}{*}{\centering\arraybackslash LOCF} & \multirow{2}{*}{\centering\arraybackslash Extrapolation}  & $\theta$       & 0.20 $\pm$ 0.40 & 0.35 $\pm$ 0.40 & 409.92\% & 1146.81\% \\
 &  & $\dot{\theta}$ & 1.7552 $\pm$ 1.19 & 0.99 $\pm$ 1.19 & 691.32\% & 764.92\% \\
 \hline
\multirow{2}{*}{\centering\arraybackslash AR(1)} & \multirow{2}{*}{\centering\arraybackslash Extrapolation} & $\theta$ & 0.14 $\pm$ 0.32 & 0.30 $\pm$ 0.32 & 182.13\% & 741.07\% \\
 &  & $\dot{\theta}$ & 0.96 $\pm$ 0.86 & 0.79 $\pm$ 0.86 & 142.16\% & 306.58\% \\
 \hline
\multirow{2}{*}{\centering\arraybackslash ARIMA} & \multirow{2}{*}{\centering\arraybackslash Extrapolation} & $\theta$ & 0.11 $\pm$ 0.32 & 0.25 $\pm$ 0.32 & 76.91\% & 402.04\% \\
 &  & $\dot{\theta}$ & 0.92 $\pm$ 0.90 & 0.77 $\pm$ 0.90 & 187.79\% & 309.63\% \\
\hline
\end{tabular}
$
}
\end{table}

\subsubsection{Lorenz Attractor Results}

\begin{figure}[!htb]
    \centering
    \begin{subfigure}[b]{0.5\textwidth}
        \includegraphics[width=\textwidth]{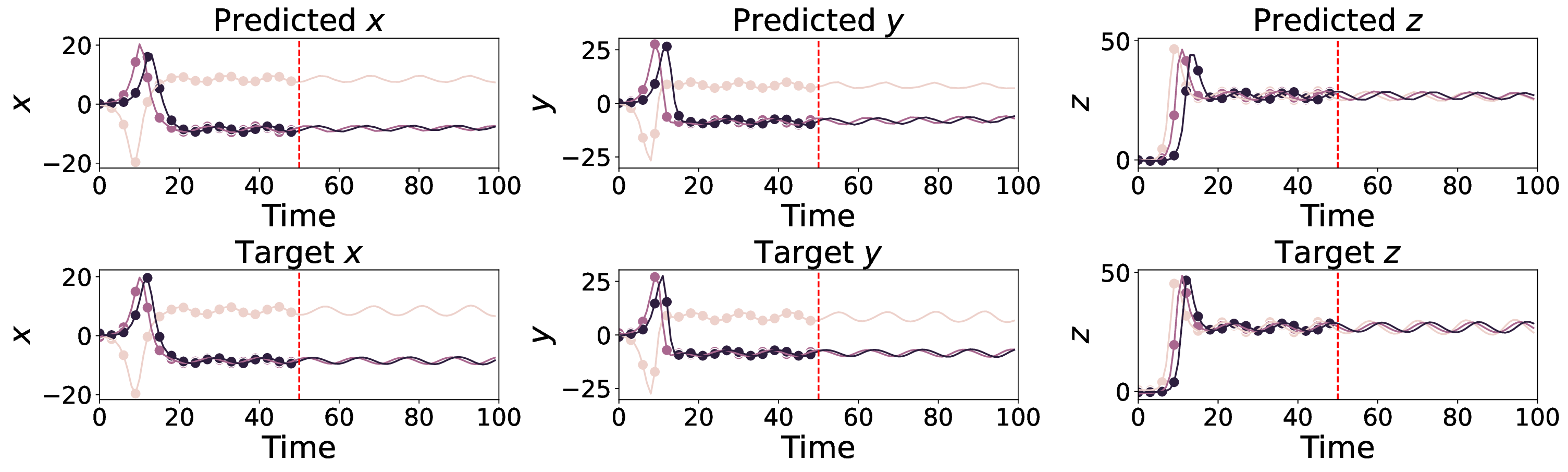}
        \caption{Training set.}
        \label{subfig:lorenz_train}
    \end{subfigure}
    \hfill
    \begin{subfigure}[b]{0.5\textwidth}
        \includegraphics[width=\textwidth]{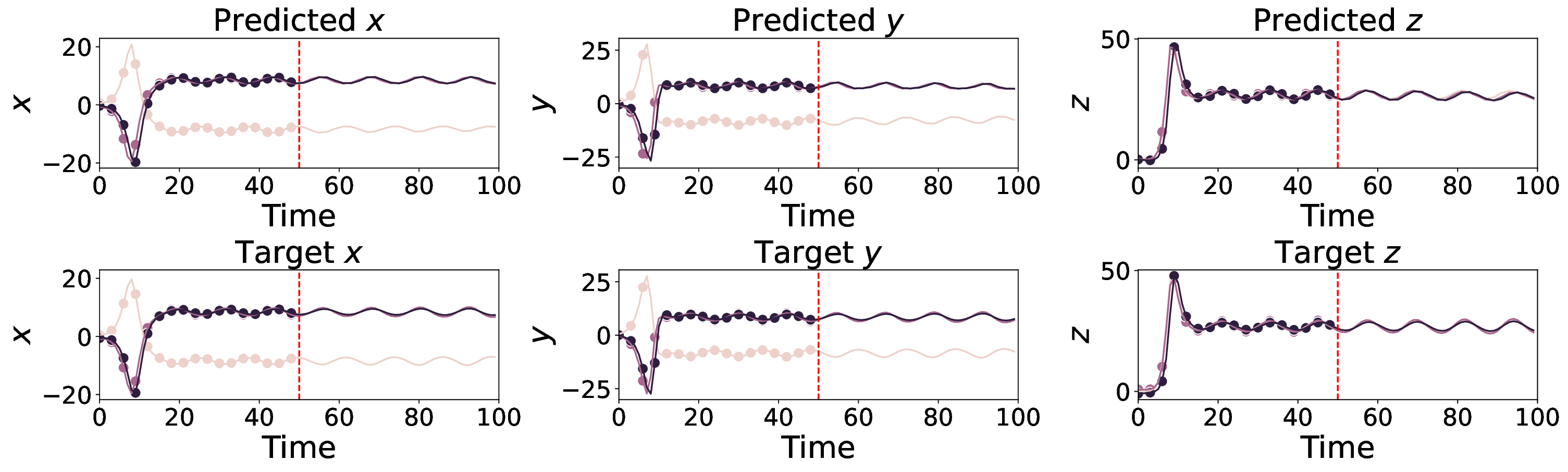}
        \caption{Testing set.}
        \label{subfig:lorenz_test}
    \end{subfigure}
    \caption{Example Lorenz attractor trajectories. a) Training set and b) Testing set. The vertical dashed line marks the end of the training window. Points beyond this line are extrapolated using the learned dynamics. Context points, shown as dots, cover one-third of the training frames.}
    \label{fig:lorenz}
\end{figure}

Figure~\ref{fig:lorenz} shows representative examples from the training and testing sets. Table~\ref{tab:feature_errors_lorenz} reports the corresponding quantitative results. In this case, we observed similar performance in extrapolation (e.g., MAPE for $x$: 3.51\%) and within the reconstruction window (e.g., MAPE for $x$: 3.11\%). Among baseline methods, LOCF performed best (MAPE for $x$: 11.37\%).

\begin{table}[!htb]
\centering
\setlength\tabcolsep{4pt}
\caption{Errors per feature in the Lorenz attractor test set (N=100). MSE: mean squared error, MAE: mean absolute error, MSPE: mean squared percentage error, MAPE: mean absolute percentage error, IQR: inter-quartile range.}
\label{tab:feature_errors_lorenz}
\resizebox{\columnwidth}{!}{%
$\begin{tabular}{|c|c|c|c|c|c|c|}
\hline
\textbf{Method} & \textbf{Task} & \textbf{Feature} & \textbf{MSE $\pm$ IQR} & \textbf{MAE $\pm$ IQR} & \textbf{MSPE [$\%$]} & \textbf{MAPE [$\%$]} \\
\hline
\multirow{6}{*}{\centering\arraybackslash Ours} & \multirow{3}{*}{\centering\arraybackslash Reconstruction}   & $x$ & 0.085 $\pm$ 0.44 & 0.29 $\pm$ 0.54 & 0.12\% & 3.51\% \\
 &   & $y$ & 0.16 $\pm$ 0.65 & 0.40 $\pm$ 0.61 & 0.21\% & 4.61\% \\
 &   & $z$ & 0.14 $\pm$ 0.57 & 0.38 $\pm$ 0.58 & 0.02\% & 1.54\% \\
 \cline{2-7}
 & \multirow{3}{*}{\centering\arraybackslash Extrapolation}  & $x$ & 0.070 $\pm$ 0.18 & 0.26 $\pm$ 0.32 & 0.10\% & 3.11\% \\
 &   & $y$ & 0.11 $\pm$ 0.33 & 0.33 $\pm$ 0.43 & 0.15\% & 3.89\% \\
 &   & $z$ & 0.13 $\pm$ 0.40 & 0.36 $\pm$ 0.48 & 0.02\% & 1.33\% \\
\hline
\multirow{3}{*}{\centering\arraybackslash LOCF} & \multirow{3}{*}{\centering\arraybackslash Extrapolation}  & $x$ & 1.58 $\pm$ 1.26 & 0.99 $\pm$ 1.26 & 17.57\% & 11.37\% \\
 &   & $y$ & 3.42 $\pm$ 1.97 & 1.45 $\pm$ 1.97 & 35.85\% & 15.97\% \\
 &   & $z$ & 4.06 $\pm$ 1.70 & 1.66 $\pm$ 1.70 & 15.19\% & 6.22\% \\
\hline
\multirow{3}{*}{\centering\arraybackslash AR(1)} & \multirow{3}{*}{\centering\arraybackslash Extrapolation} & $x$ & 11.14 $\pm$ 1.81 & 3.06 $\pm$ 1.81 & 126.45\% & 35.73\% \\
 &  & $y$ & 14.55 $\pm$ 2.40 & 3.51 $\pm$ 2.40 & 159.68\% & 40.52\% \\
 &  & $z$ & 29.73 $\pm$ 3.62 & 4.88 $\pm$ 3.62 & 106.85\% & 17.83\% \\
\hline
\multirow{3}{*}{\centering\arraybackslash ARIMA} & \multirow{3}{*}{\centering\arraybackslash Extrapolation} & $x$ & 1.76 $\pm$ 1.36 & 1.02 $\pm$ 1.36 & 19.34\% & 11.49\% \\
 &  & $y$ & 19.27 $\pm$ 2.21 & 2.53 $\pm$ 2.21 & 237.55\% & 29.70\% \\
 &  & $z$ & 9.45 $\pm$ 3.03 & 2.53 $\pm$ 3.03 & 33.76\% & 9.18\% \\
\hline
\end{tabular}
$
}
\end{table}

\subsubsection{Kuramoto System Results}

\begin{figure}[!htb]
\centering
\begin{subfigure}[b]{0.5\textwidth}
    \includegraphics[width=\textwidth]{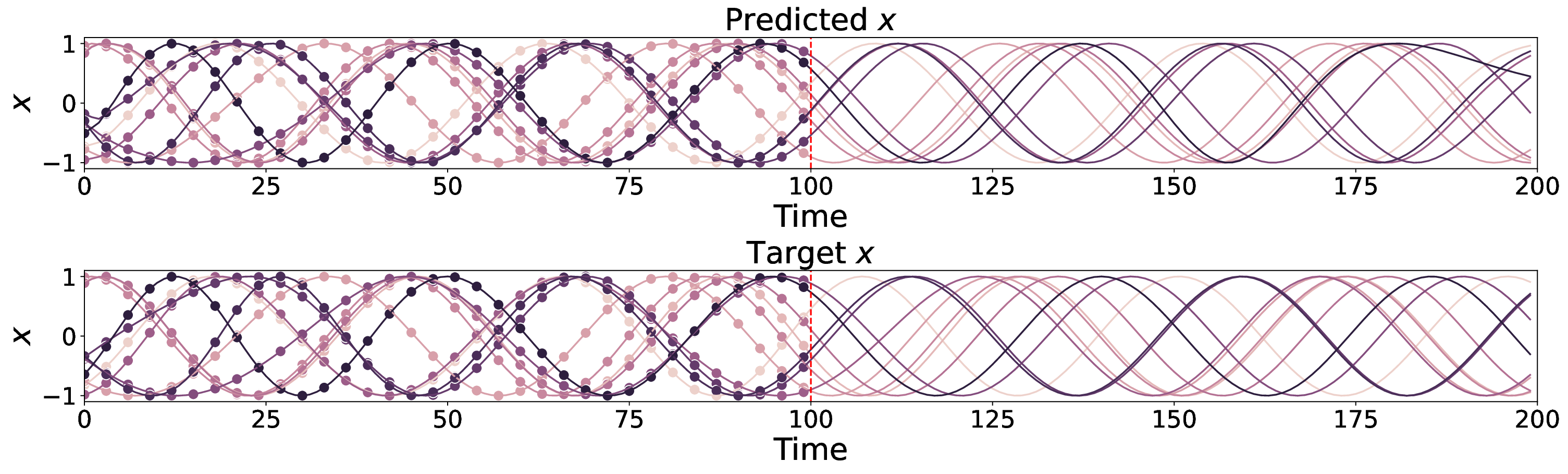}
    \caption{Training set.}
    \label{subfig:kuramoto_train}
\end{subfigure}
\hfill
\begin{subfigure}[b]{0.5\textwidth}
    \includegraphics[width=\textwidth]{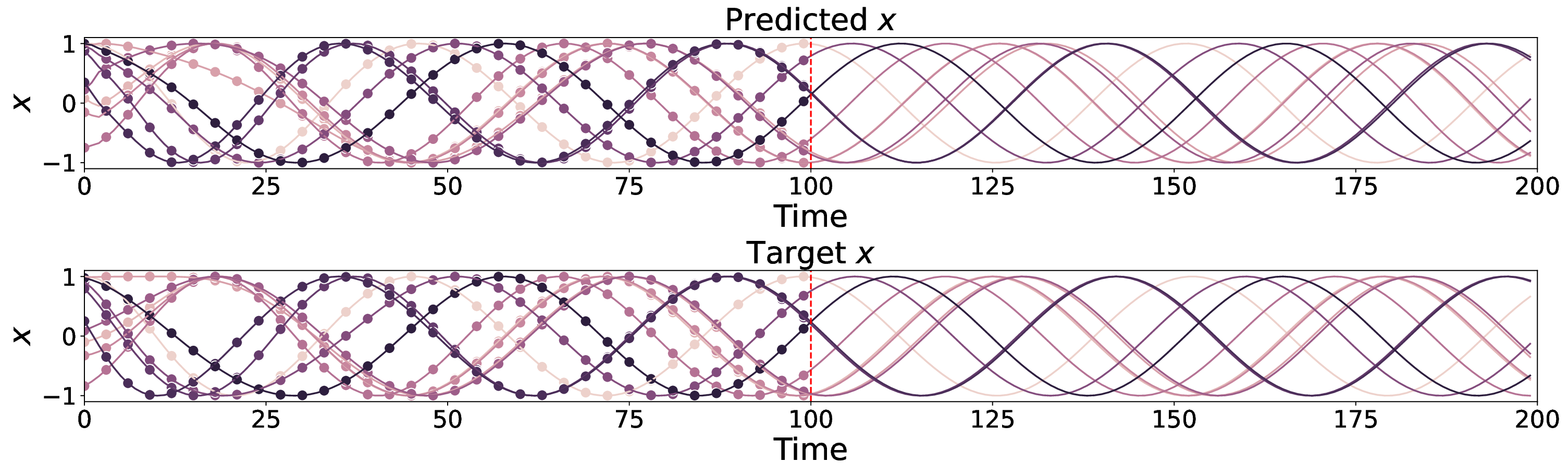}
    \caption{Testing set.}
    \label{subfig:kuramoto_test}
\end{subfigure}
\caption{Example Kuramoto trajectories. a) Training set and b) Testing set. The vertical dashed line marks the end of the training window. Points beyond this line are extrapolated using the learned dynamics. Context points, shown as dots, cover one-third of the training frames. For visualization purposes, the phase ($x$) of each node has been transformed using the sine function.}
\label{fig:kuramoto}
\end{figure}

Figure~\ref{fig:kuramoto} shows examples of trajectories for the Kuramoto system, while Table~\ref{tab:feature_errors_kuramoto} presents the corresponding quantitative error metrics. As with the coupled pendulum system, extrapolation error beyond the training window (MAPE 3.33\%) was higher than reconstruction error within the training window (MAPE 0.99\%). In this case, ARIMA achieved better MAPE (1.59\%) but worse performance in terms of MSPE (3.38\% vs. 0.11\%).

\begin{table}[!htb]
\centering
\setlength\tabcolsep{4pt}
\caption{Errors per feature in the Kuramoto oscillator test set (N=100). MSE: mean squared error, MAE: mean absolute error, MSPE: mean squared percentage error, MAPE: mean absolute percentage error, IQR: inter-quartile range.}
\label{tab:feature_errors_kuramoto}
\resizebox{\columnwidth}{!}{%
$\begin{tabular}{|c|c|c|c|c|c|c|}
\hline
\textbf{Method} & \textbf{Task} & \textbf{Feature} & \textbf{MSE $\pm$ IQR} & \textbf{MAE $\pm$ IQR} & \textbf{MSPE [$\%$]} & \textbf{MAPE [$\%$]} \\
\hline
\multirow{2}{*}{\centering\arraybackslash Ours} & Reconstruction   & $x$ & 0.013 $\pm$ 0.043 & 0.11 $\pm$ 0.16 & 0.01\% & 0.99\% \\
\cline{2-7}
 & Extrapolation  & $x$ & 1.02 $\pm$ 5.12 & 1.01 $\pm$ 1.92 & 0.11\% & 3.33\% \\
\hline
LOCF & Extrapolation  & $x$ & 126.01 $\pm$ 9.31 & 9.50 $\pm$ 9.31 & 338.56\% & 28.15\% \\
\hline
AR(1) & Extrapolation & $x$ & 131.39 $\pm$ 9.50 & 9.70 $\pm$ 9.50 & 353.26\% & 28.77\% \\
\hline
ARIMA & Extrapolation & $x$ & 1.20 $\pm$ 0.62 & 0.53 $\pm$ 0.62 & 3.38\% & 1.59\% \\
\hline
\end{tabular}
$
}
\end{table}

\subsection{Trajectory Prediction on Cardiac MRI Data}

Similar to the synthetic data experiments, we split each dataset into $80\%$ for training, $20\%$ for validation, and $20\%$ for testing. One-third of the training time points are selected as context points, spaced evenly and shared across all samples. For extrapolation, predictions are made beyond the training window using the learned dynamics for an equivalent number of time steps.

For each cardiac dataset, we explored two spatial connectivity settings: (i) a fully connected graph without self-loops (Full), and (ii) a graph where only anatomically adjacent regions are connected (Anat). Temporal connectivity was always represented by a diagonal adjacency matrix.

To select the hyperparameters for each setting, we ran $100$ Optuna~\citep{optuna_2019} trials, using the MAE on the validation set as the optimization objective. The selected hyperparameters for each experiment are listed in Appendix Table~\ref{tab:hyperparams_multiplex_values}. All experiments used a fixed batch size of $30$ and an initial learning rate of $0.01$. A learning rate scheduler was used to reduce the learning rate by a factor of 2 when the validation metric plateaued, with a patience of $10$ and a minimum learning rate of $10^{-5}$.

\subsubsection{ACDC: Trajectory Prediction}

\begin{figure}[!htb]
    \centering
    \begin{minipage}{0.49\textwidth}
        \centering
        \includegraphics[width=\linewidth]{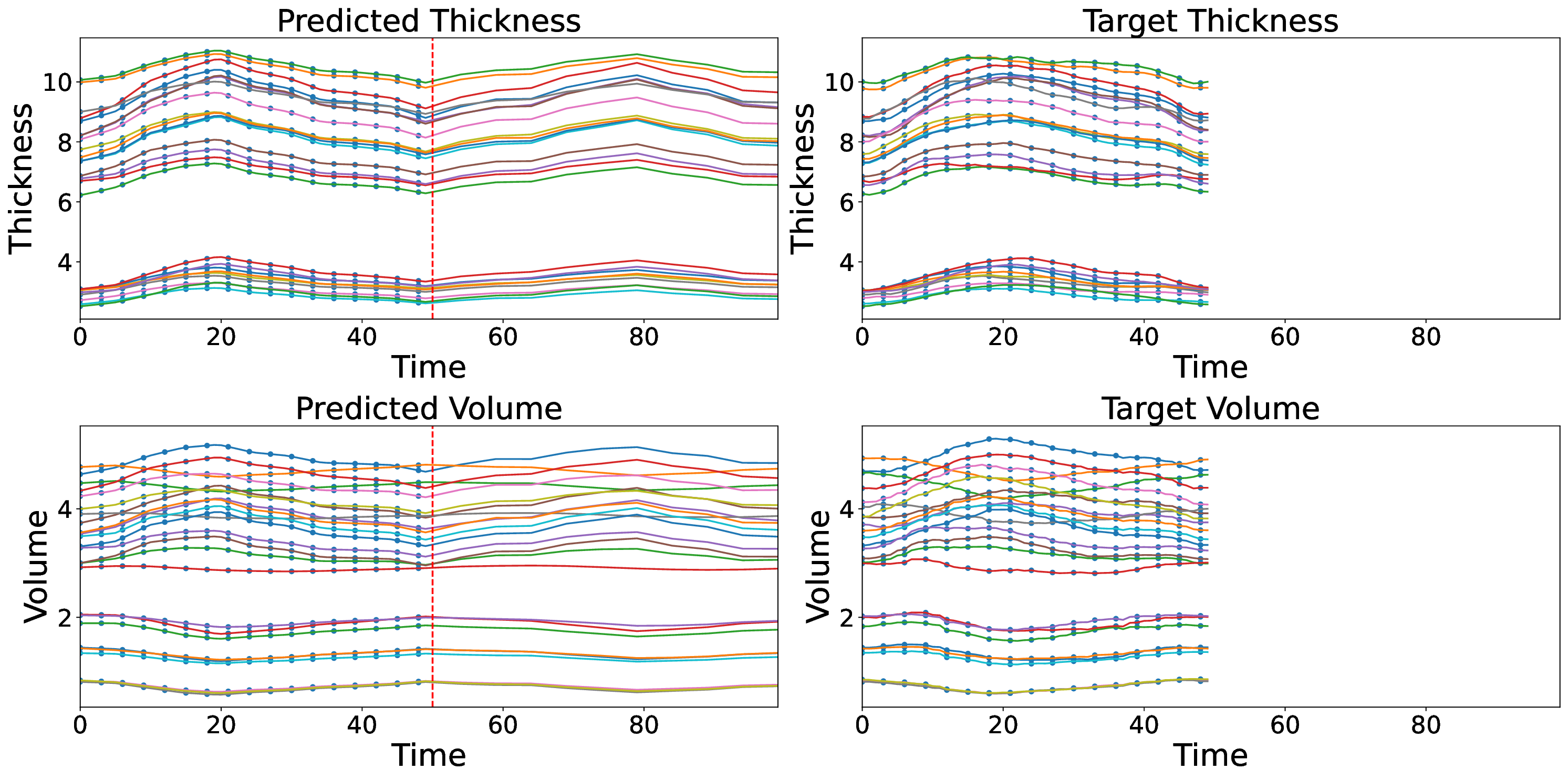}
        \subcaption{Full}
        \label{subfig:acdc_trajectories_avg_all}
    \end{minipage}
    \vfill
    \begin{minipage}{0.49\textwidth}
        \centering
        \includegraphics[width=\linewidth]{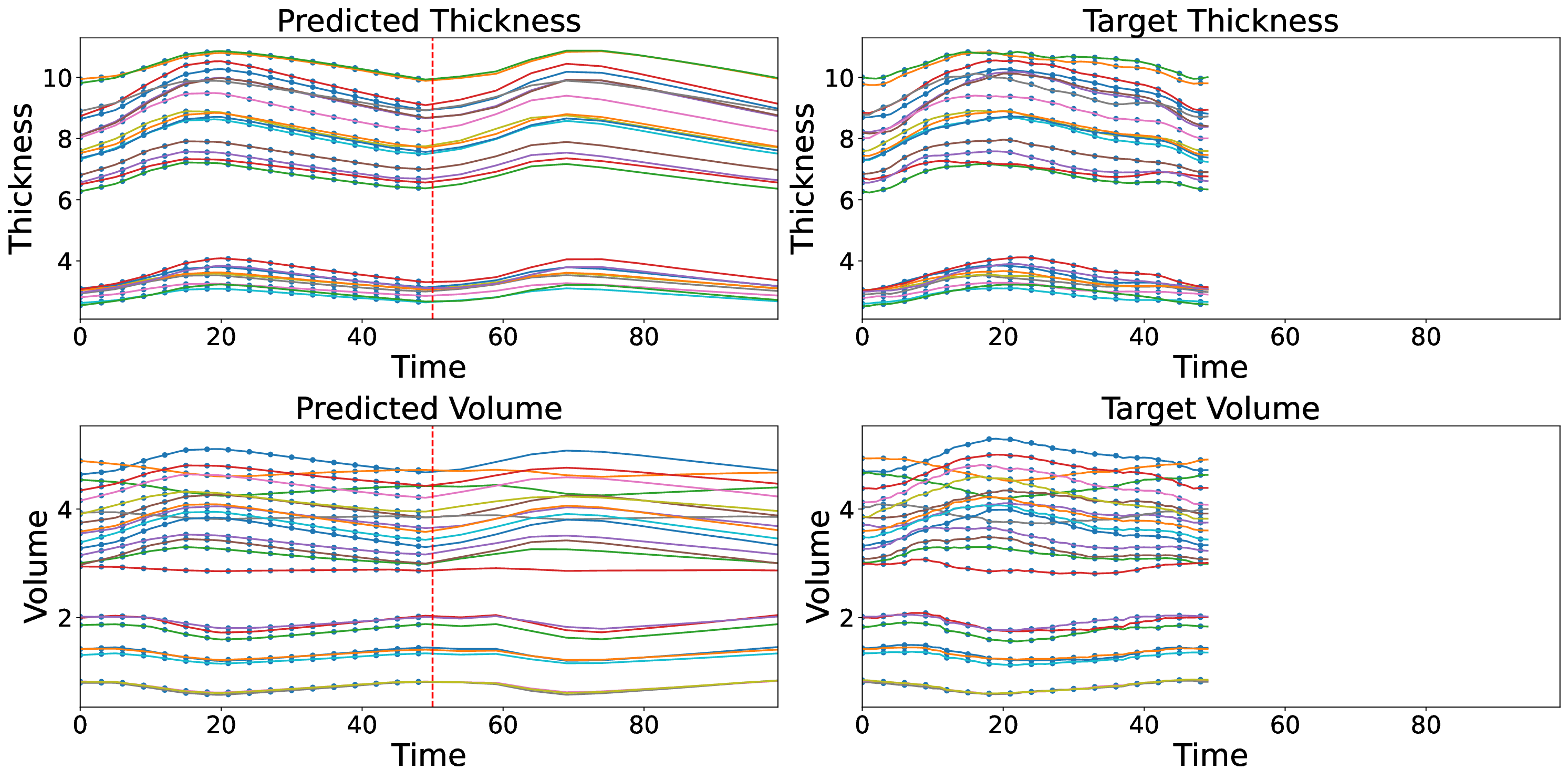}
        \subcaption{Anat}
        \label{subfig:acdc_trajectories_avg_aha}
    \end{minipage}
    \caption{
    Average predicted feature trajectories on the ACDC dataset. a) Fully connected (Full) graph. b) Anatomically adjacent connectivity (Anat). Each line represents a different cardiac region. The vertical dashed line marks the end of the training window.}    \label{fig:acdc_trajectories_average}
\end{figure}

After training, we evaluated the model’s ability to reconstruct and extrapolate regional feature trajectories from sparse context observations. Figure~\ref{fig:acdc_trajectories_average} shows average predicted trajectories across the ACDC dataset using the two spatial connectivity schemes: Figure~\ref{subfig:acdc_trajectories_avg_aha} for anatomical connectivity (Anat) and Figure~\ref{subfig:acdc_trajectories_avg_all} for fully connected graphs (Full). Despite training on a single cardiac cycle, the model successfully predicts a plausible continuation of the cycle based on the learned underlying dynamics. 

\begin{figure}[!htbp]
    \centering    \includegraphics[width=0.49\textwidth]{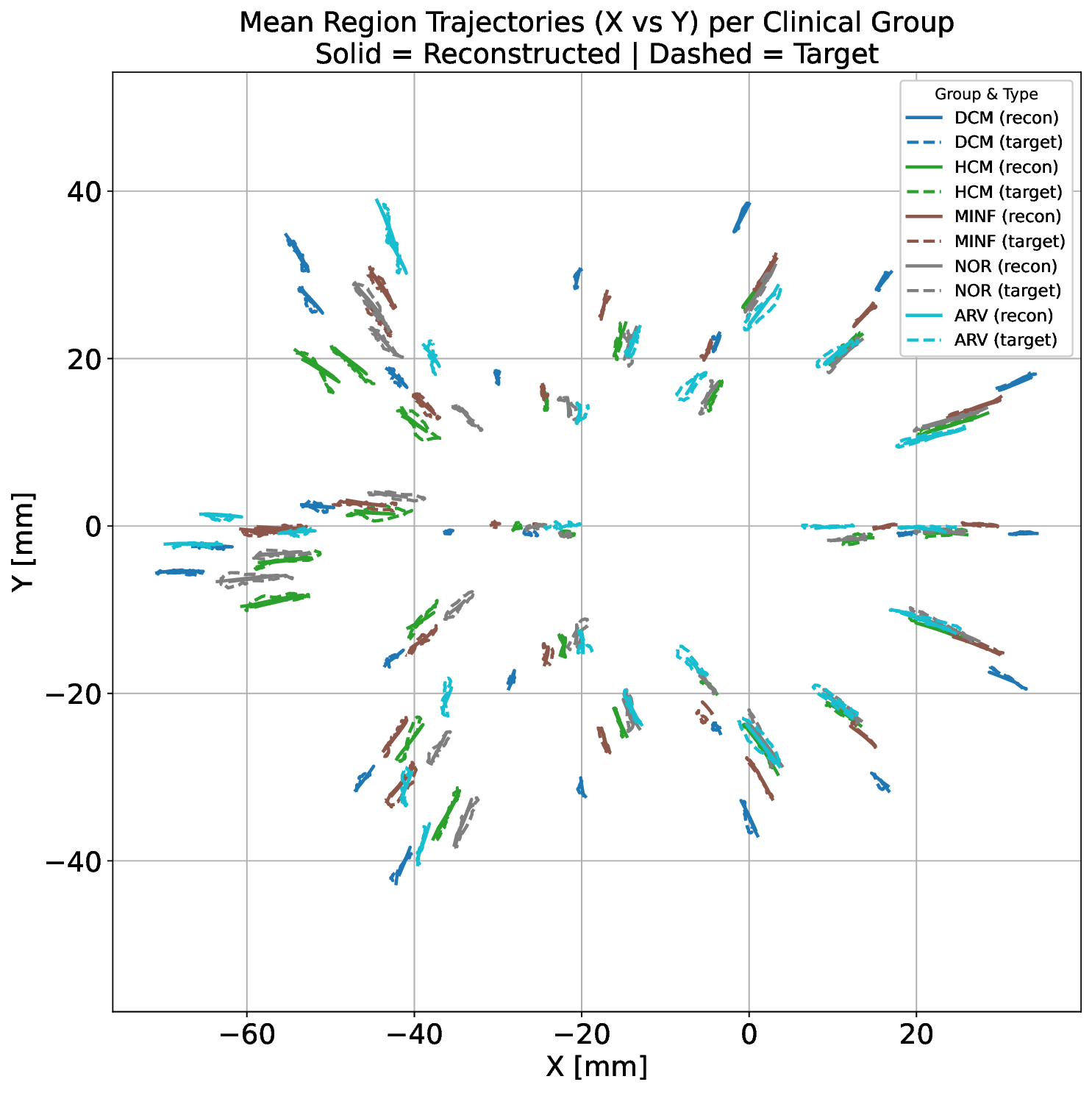}
    \caption{
    Average predicted trajectories of the center of mass for each region on the ACDC dataset using anatomical (Anat) connectivity. Line color indicates clinical group; line type distinguishes between target and reconstructed trajectories.}
    \label{fig:acdc_aha_positions_average}
\end{figure}

Table~\ref{tab:feature_errors_ACDC} presents reconstruction error metrics for thickness and volume, with slightly better results for anatomical connectivity. Figure~\ref{fig:acdc_aha_positions_average} shows the average trajectory of the center of mass for each cardiac region, projected onto the X-Y plane and stratified by clinical diagnosis. These trajectories reveal clear differences across cardiac pathologies. This spatial information complements the observed variations in thickness and volume per region, shown in Appendix Figures~\ref{fig:group_trajectories_aha_acdc} and~\ref{fig:group_trajectories_all_acdc}.

\begin{table}[!htb]
\centering
\setlength\tabcolsep{4pt} 
\caption{Reconstruction error metrics on the ACDC test set (N=30). MAE: mean absolute error, MSE: mean squared error, IQR: inter-quartile range.}
\label{tab:feature_errors_ACDC}
\resizebox{\columnwidth}{!}{%
$\begin{tabular}{|c|c|c|c|c|c|}
\hline
\textbf{Connectivity} & \textbf{Feature} & \textbf{MSE $\pm$ IQR} & \textbf{MAE $\pm$ IQR} & \textbf{MSPE [$\%$]} & \textbf{MAPE [$\%$]} \\
\hline
\multirow{2}{*}{\centering\arraybackslash Full} & Thickness [mm] & 0.093 $\pm$ 0.28 & 0.30 $\pm$ 0.42 & 0.26$\%$ & 5.09$\%$ \\
 & Volume [ml] & 0.028 $\pm$ 0.076 & 0.17 $\pm$ 0.22 & 0.40$\%$ & 6.32$\%$ \\
 \hline
\multirow{2}{*}{\centering\arraybackslash Anat} & Thickness [mm] & 0.082 $\pm$ 0.26 & 0.29 $\pm$ 0.40 & 0.24$\%$ & 4.90$\%$ \\
 & Volume [ml] & 0.028 $\pm$ 0.072 & 0.17 $\pm$ 0.21 & 0.37$\%$ & 6.11$\%$ \\
\hline
\end{tabular}
$
} 
\end{table}

\subsubsection{UK Biobank: Trajectory Prediction}

\begin{figure}[!htb]
    \centering
    \begin{minipage}{0.49\textwidth}
        \centering
        \includegraphics[width=\linewidth]{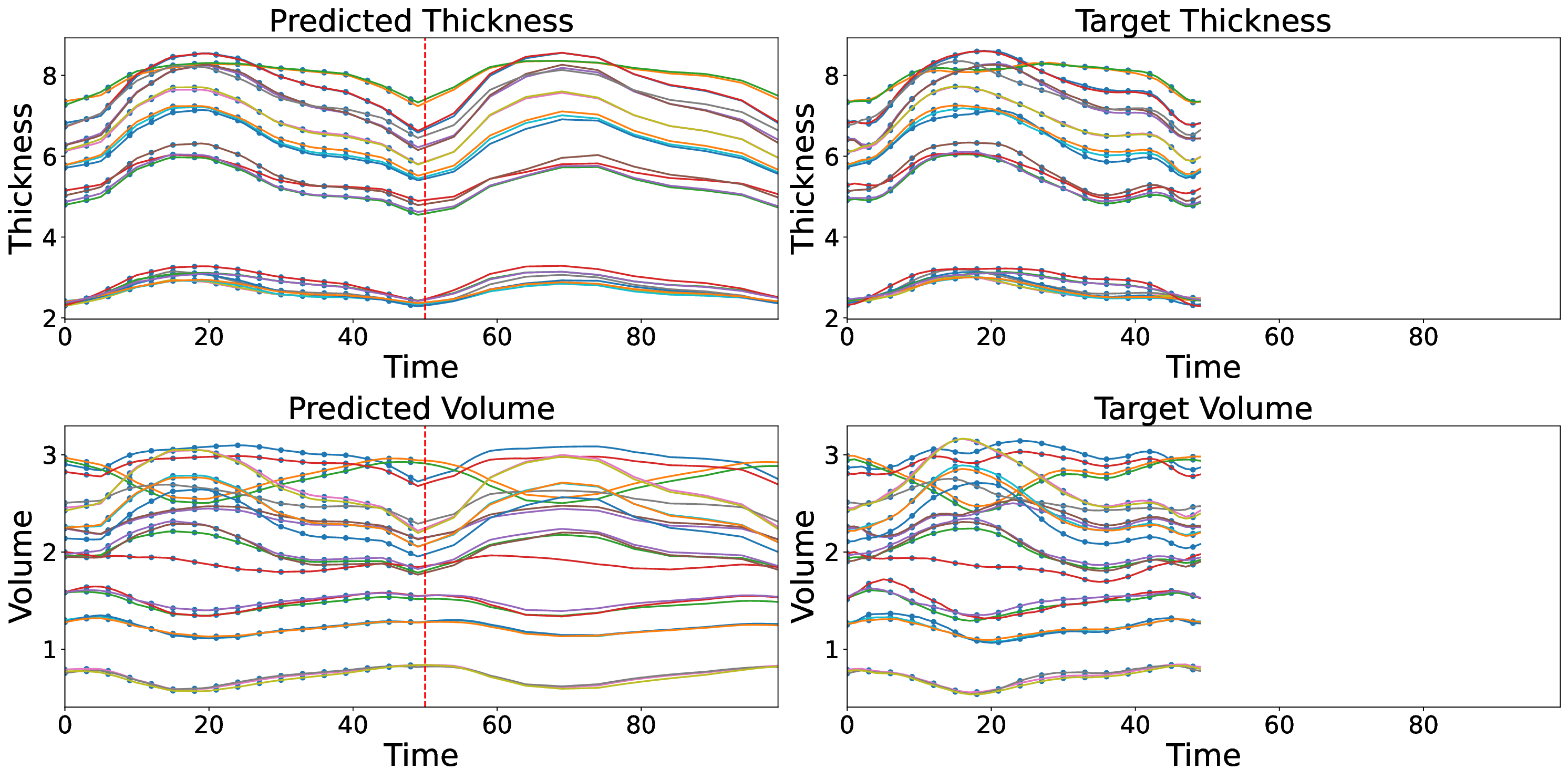}
        \subcaption{Full}
        \label{subfig:ukb_trajectories_avg_all}
    \end{minipage}
    \vfill
    \begin{minipage}{0.49\textwidth}
        \centering
        \includegraphics[width=\linewidth]{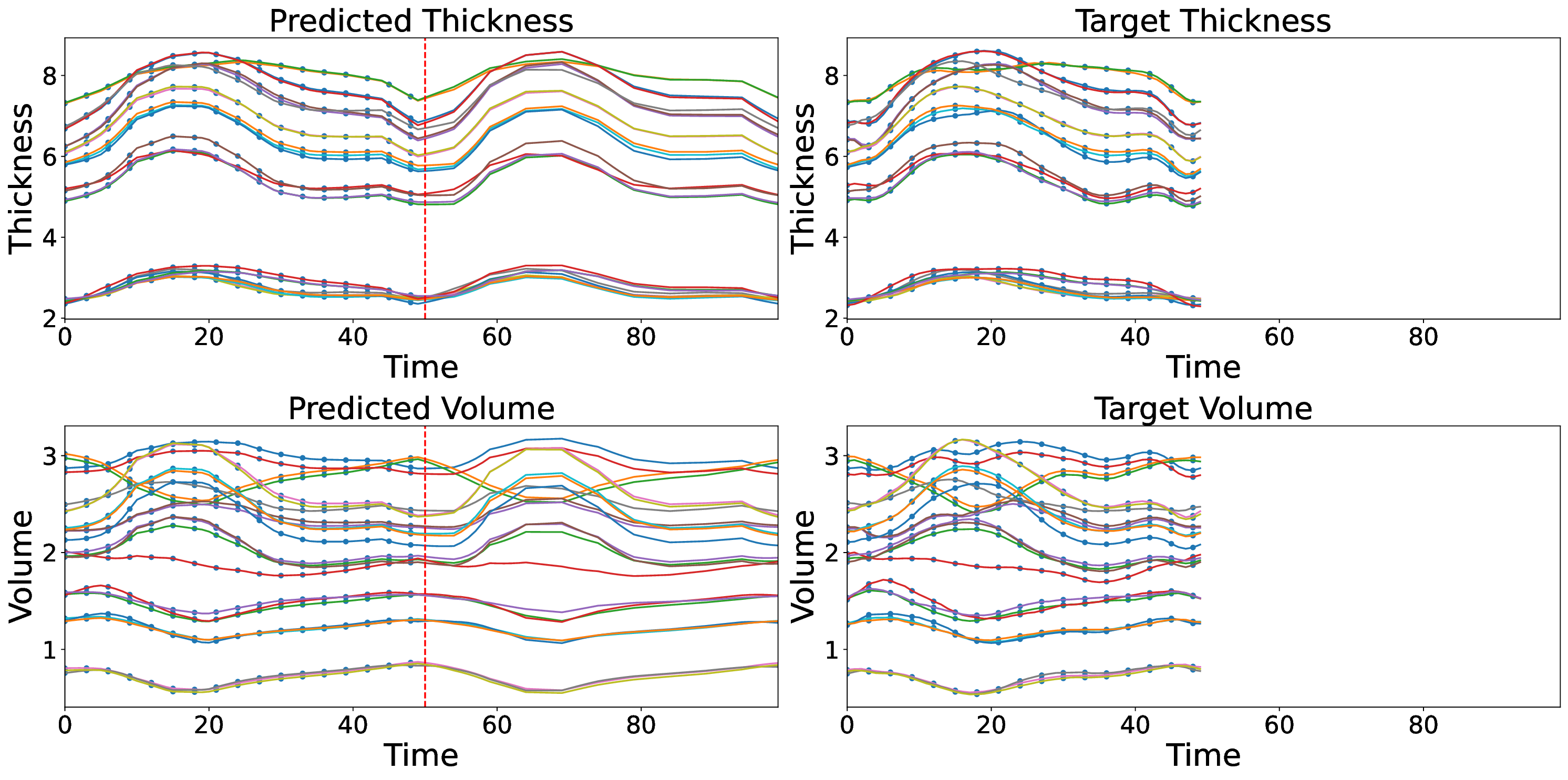}
        \subcaption{Anat}
        \label{subfig:ukb_trajectories_avg_aha}
    \end{minipage}
    \caption{Average predicted feature trajectories for the UKB dataset. a) Fully connected (Full) graph. b) Anatomically adjacent connectivity. Each line represents a different cardiac region. The vertical dashed line marks the end of the training window.}
    \label{fig:ukb_trajectories_average}
\end{figure}

Figure~\ref{fig:ukb_trajectories_average} shows the average predicted feature trajectories for the UKB dataset, with Figure~\ref{subfig:ukb_trajectories_avg_aha} using anatomical (Anat) graph connectivity and Figure~\ref{subfig:ukb_trajectories_avg_all} using a fully connected graph. As in the ACDC results, the model successfully learns the underlying system and can predict the continuation of the cardiac cycle, despite being trained on data from only a single cycle. Reconstruction errors for thickness and volume are reported in Table~\ref{tab:feature_errors_ukb}, showing very similar performance between anatomical and fully connected configurations.

\begin{figure}[!htbp]
    \centering    \includegraphics[width=0.49\textwidth]{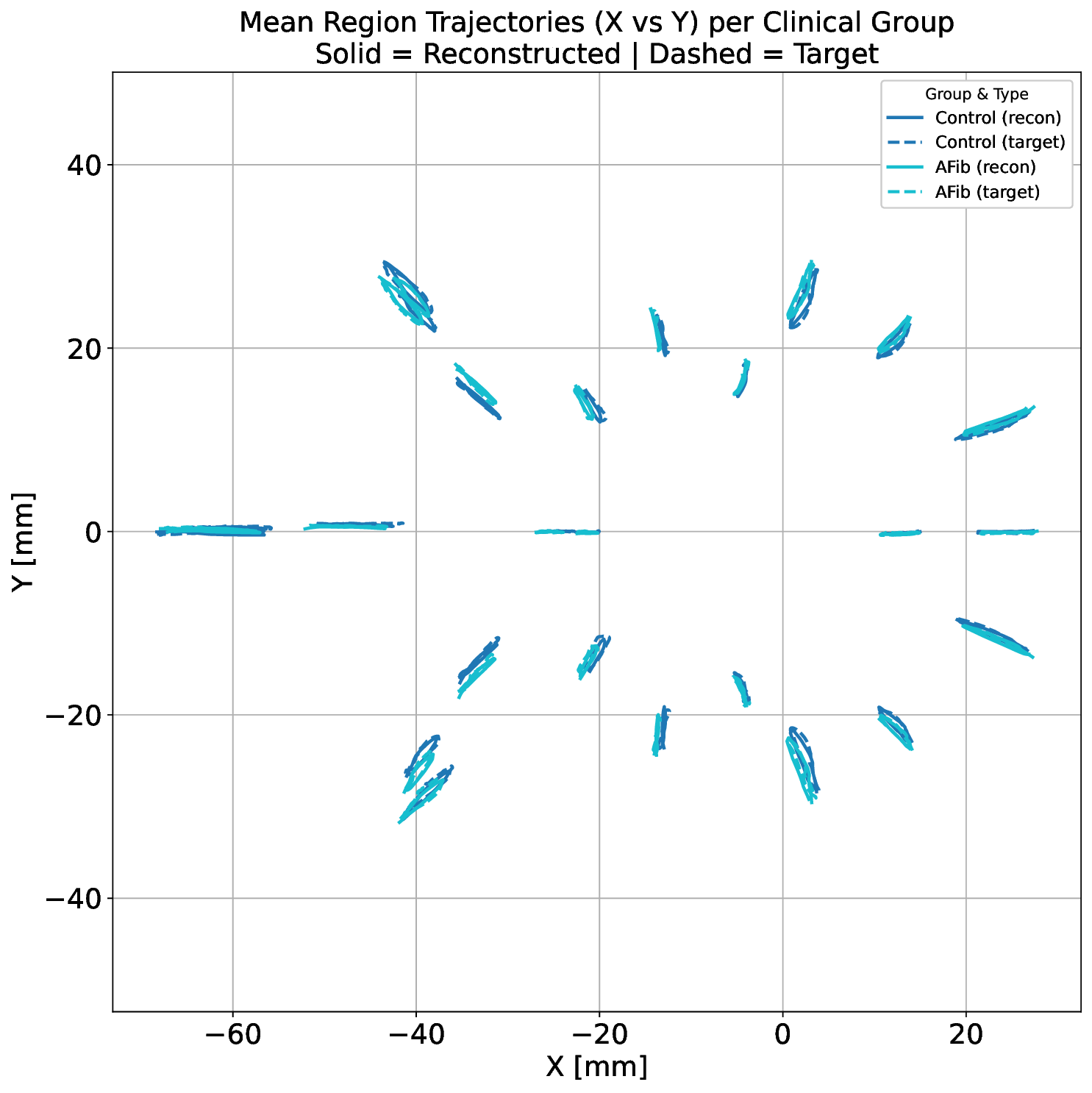}
    \caption{Average predicted center of mass trajectories for the UKB dataset using anatomical (Anat) connectivity. Each line represents a cardiac region. Line color indicates the clinical group; line type denotes target vs. reconstructed trajectories.}
    \label{fig:ukb_aha_positions_average}
\end{figure}

Figure~\ref{fig:ukb_aha_positions_average} illustrates the average trajectory of each region's center of mass in the X-Y plane, stratified by clinical diagnosis (control, AFib). These trajectories capture subtle but consistent differences in cardiac motion patterns between AFib and control groups. Additional results are provided in Appendix Figures~\ref{fig:group_trajectories_aha_ukb} and~\ref{fig:group_trajectories_all_ukb}.

\begin{table}[!htb]
\centering
\setlength\tabcolsep{4pt} 
\caption{Interpolation error metrics on the UK Biobank test set (N=83). MAE: mean absolute error, MSE: mean squared error, IQR: inter-quartile range.}
\label{tab:feature_errors_ukb}
\resizebox{\columnwidth}{!}{%
$\begin{tabular}{|c|c|c|c|c|c|}
\hline
\textbf{Task} & \textbf{Feature} & \textbf{MSE $\pm$ IQR} & \textbf{MAE $\pm$ IQR} & \textbf{MSPE [$\%$]} & \textbf{MAPE [$\%$]} \\
\hline
\multirow{2}{*}{\centering\arraybackslash Full} & Thickness & 0.065 $\pm$ 0.16 & 0.25 $\pm$ 0.31 & 0.22$\%$ & 4.64$\%$ \\
 & Volume & 0.011 $\pm$ 0.033 & 0.11 $\pm$ 0.14 & 0.36$\%$ & 5.99$\%$ \\
\hline
\multirow{2}{*}{\centering\arraybackslash Anat} & Thickness [mm] & 0.063 $\pm$ 0.15 & 0.25 $\pm$ 0.30 & 0.22$\%$ & 4.66$\%$ \\
 & Volume [ml] & 0.011 $\pm$ 0.032 & 0.10 $\pm$ 0.14 & 0.35$\%$ & 5.93$\%$ \\
\hline
\end{tabular}
$
} 
\end{table}


\subsection{Classification on Cardiac Data}

We next assessed whether the learned latent representations can be used to classify cardiac conditions. Hyperparameters for the classifiers were tuned via 3-fold cross-validation using HalvingGridSearch~\citep{pedregosa_scikit-learn_2011}. The list of hyperparameters is provided in Table~\ref{tab:hyperparameters}. We used the same data splits as in the trajectory reconstruction task, i.e., identical training and test sets.

\subsubsection{ACDC: Multi-class Classification}

We evaluated performance on the 5-class classification task defined in the ACDC challenge, which includes healthy subjects and four cardiac pathologies. We trained classifiers using both the latent representations ($l_0$, $d_0$) produced by our model and the full set of graph-derived features. Random Forest, XGBoost, and $k$-Nearest Neighbors ($k$NN) classifiers were tested. Results in Table~\ref{tab:results_classification_acdc} show that the low-dimensional latent representations learned by our model offer performance equal to or better than that of full graph features, despite being two orders of magnitude smaller in dimensionality.

\begin{table}[!htb]
    \centering
    \setlength\tabcolsep{3pt} 
    \caption{Test set classification accuracy (mean $\pm$ std across CV folds) on the ACDC 5-class task.}
    \label{tab:results_classification_acdc}    
    \resizebox{\columnwidth}{!}{%
    \begin{tabular}{|c|c|c|r@{\hspace{1mm}}l|}
    \hline
    \textbf{Connectivity} & \textbf{Features (dimension)} & \textbf{Method} & \multicolumn{2}{c|}{\textbf{Accuracy}} \\
    \hline
    \multirow{4}{*}{\centering\arraybackslash Full} & $L_{0} + D_{0}$ (550) & Multiplex + RF & 0.95 & $\pm$ 0.01 \\
    \cline{2-5}
     & \multirow{3}{*}{All features (82,532)} & Random Forest & 0.89 & $\pm$ 0.04 \\
     & & XGBoost & 0.94 & $\pm$ 0.04 \\
     & & $k$NN & 0.63 & $\pm$ 0.03 \\
    \cline{2-5}
    \hline
    \multirow{4}{*}{\centering\arraybackslash Anat} & $L_{0} + D_{0}$ (525) & Multiplex + RF & 0.99 & $\pm$ 0.02 \\ 
    \cline{2-5}
     & \multirow{3}{*}{All features (36,332)} & Random Forest & 0.93 & $\pm$ 0.00 \\
     &  & XGBoost & 0.94 & $\pm$ 0.02 \\
     &  & $k$NN & 0.60 & $\pm$ 0.08 \\
    \cline{2-5}
    \hline
    \end{tabular}%
    }
\end{table}

\subsubsection{UK Biobank: Binary Classification}

We evaluated classification performance on the UK Biobank cohort for distinguishing atrial fibrillation (AFib) from controls. As in the ACDC experiment, we compared the latent representations from our model to the full graph-derived features using Random Forest, XGBoost, and $k$NN classifiers. Table~\ref{tab:results_classification_ukb} shows that while the latent representations achieved competitive performance, they were slightly outperformed by the full feature set. Nevertheless, the learned latent space remains compact, two orders of magnitude smaller, highlighting its efficiency.

\begin{table}[!htb]
\centering
\setlength\tabcolsep{3pt} 
\caption{Classification accuracy (mean $\pm$ std across CV folds) on the UK Biobank AFib vs. control task.}
\label{tab:results_classification_ukb}
\resizebox{\columnwidth}{!}{%
\begin{tabular}{|c|c|c|r@{\hspace{1mm}}l|}
\hline
\textbf{Connectivity} & \textbf{Features (dimension)} & \textbf{Method} & \multicolumn{2}{c|}{\textbf{Accuracy}} \\
\hline
\multirow{4}{*}{\centering\arraybackslash Full} & $L_{0} + D_{0} (425)$ & Multiplex + RF & 0.62 & $\pm$ 0.01 \\
\cline{2-5}
 & \multirow{3}{*}{All features (82,532)} & Random Forest & 0.68 & $\pm$ 0.01 \\
 & & XGBoost & 0.70 & $\pm$ 0.02 \\
 & & $k$NN & 0.52 & $\pm$ 0.03 \\
\cline{2-5}
\hline
\multirow{4}{*}{\centering\arraybackslash Anat} & $L_{0} + D_{0} (450) $ & Multiplex + RF & 0.67 & $\pm$ 0.01 \\
\cline{2-5}
 & \multirow{3}{*}{All features (36,332)} & Random Forest & 0.70 & $\pm$ 0.01 \\
 &  & XGBoost & 0.64 & $\pm$ 0.02 \\
 &  & $k$NN & 0.45 & $\pm$ 0.02 \\
\cline{2-5}
\hline
\end{tabular}%
}
\end{table}

\subsection{Latent Space Exploration}

\begin{figure}[!htb]
    \centering
    \begin{minipage}{0.49\textwidth}
        \centering
        \includegraphics[width=\linewidth]{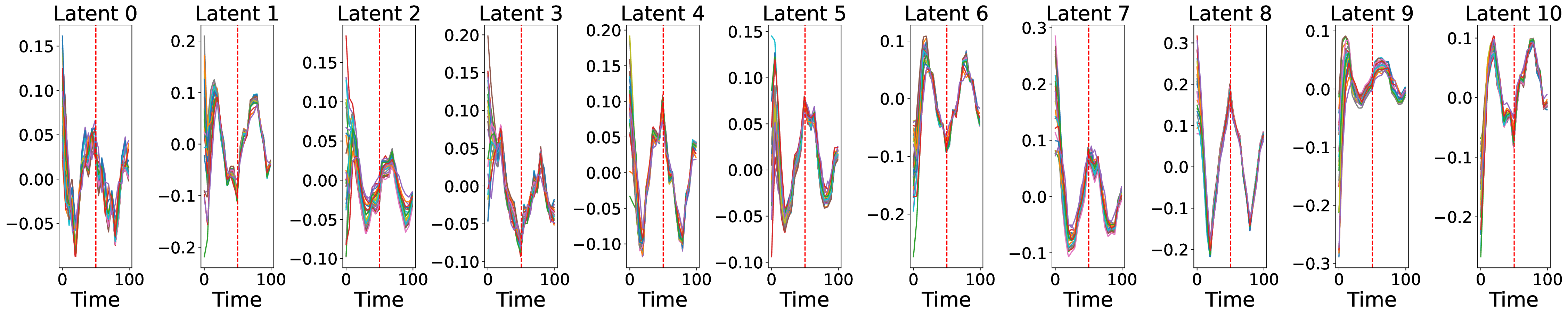}
        \subcaption{Full}
        \label{subfig:acdc_latent_avg_all}
    \end{minipage}
    \vfill
    \begin{minipage}{0.49\textwidth}
        \centering
        \includegraphics[width=\linewidth]{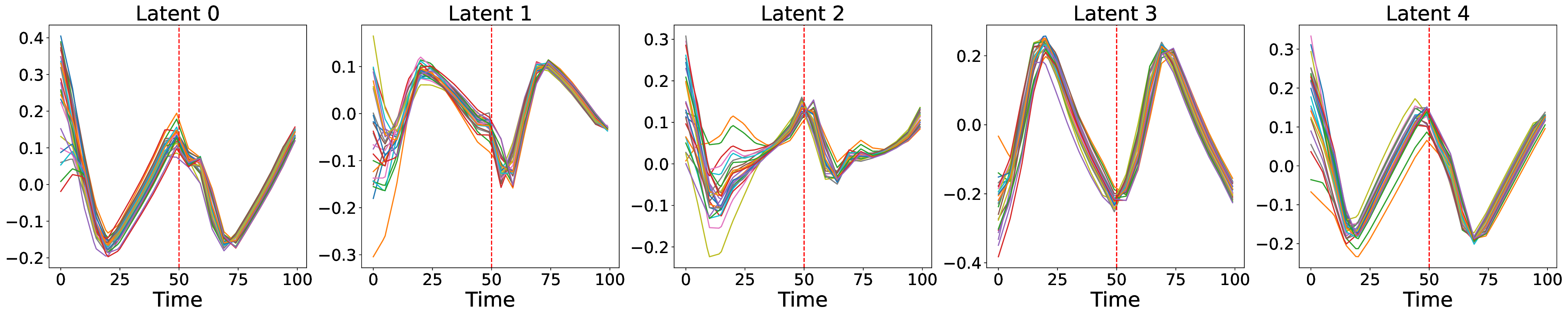}
        \subcaption{Anat}
        \label{subfig:acdc_latent_avg_aha}
    \end{minipage}
    \caption{Average latent feature trajectories for ACDC using a) fully connected (Full) and b) anatomical (Anat) connectivity. Each line represents a different cardiac region. The vertical dashed line marks the end of the training window.}
    \label{fig:latent_avg_acdc}
\end{figure}

We explored the latent trajectories ($L$) of different cardiac regions across time, both within the training window and during extrapolation. Figure~\ref{fig:latent_avg_acdc} shows the average latent trajectories for the ACDC dataset, using anatomical connectivity in Figure~\ref{subfig:acdc_latent_avg_aha} and full connectivity in Figure~\ref{subfig:acdc_latent_avg_all}. In both cases, we observe that the latent trajectories exhibit a cyclic behavior, continuing plausibly into the extrapolation window. Notably, the fully connected model uses a higher-dimensional latent space due to hyperparameter selection. Some inter-regional differences are also apparent. 

\begin{figure}[!htb]
    \centering
    \begin{minipage}{0.49\textwidth}
        \centering
        \includegraphics[width=\linewidth]{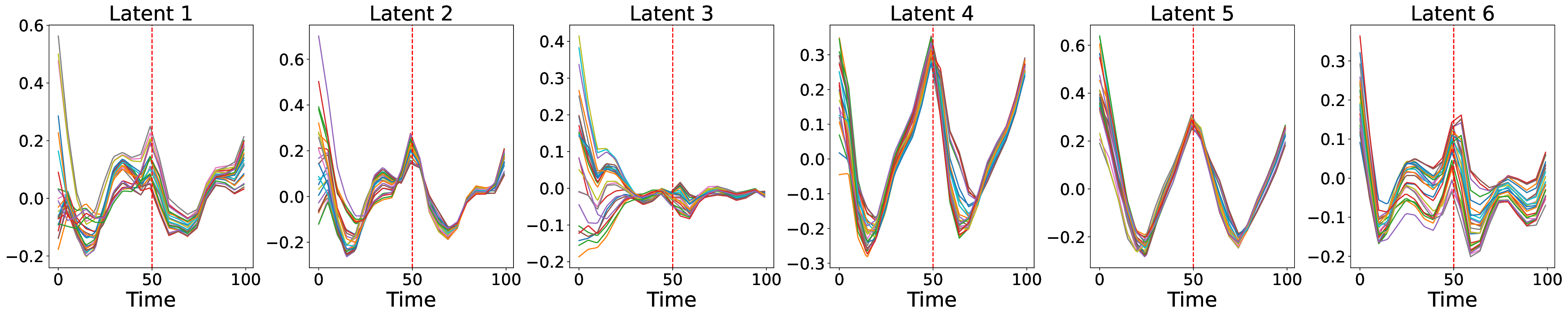}
        \subcaption{Full}
        \label{subfig:ukb_latent_avg_all}
    \end{minipage}
    \vfill
    \begin{minipage}{0.49\textwidth}
        \centering
        \includegraphics[width=\linewidth]{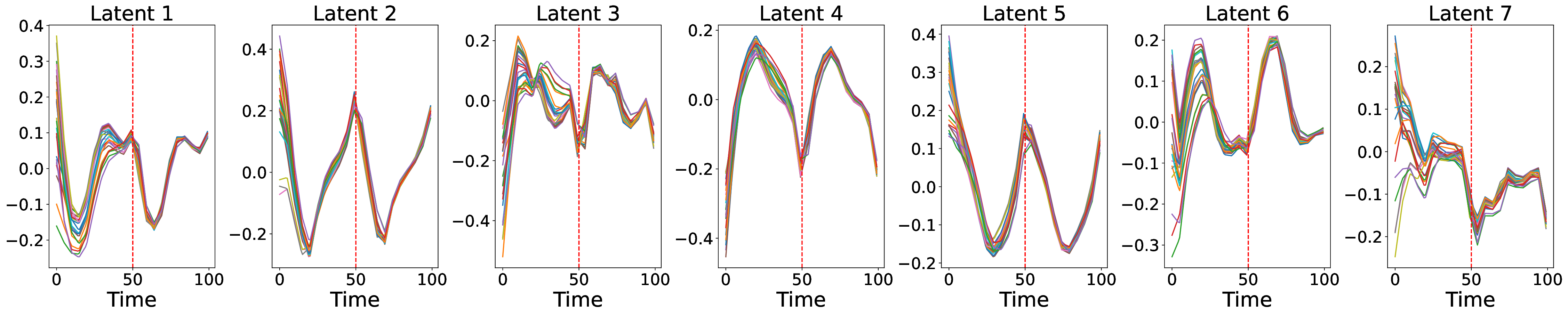}
        \subcaption{Anat}
        \label{subfig:ukb_latent_avg_aha}
    \end{minipage}
        \caption{Average latent feature trajectories for UKB using a) fully connected (Full) and b) anatomical (Anat) connectivity. Each line shows a different cardiac region. The vertical dashed line marks the end of the training window.}
    \label{fig:latent_avg_ukb}
\end{figure}

Figure~\ref{fig:latent_avg_ukb} presents similar results for the UK Biobank dataset. Again, a cyclic pattern emerges, with extrapolated trajectories closely mimicking the training window. Additionally, subtle patterns appear to be captured, suggesting the latent space encodes clinically meaningful variations.


Given the stronger classification performance obtained using anatomical connectivity, we used this configuration to analyze associations between the latent space and clinical biomarkers. We first flattened the latent representation ($l_{0}$, $d_{0}$) and applied principal component analysis (PCA), retaining components that explained 95\% of the variance—$41$ components for ACDC and $61$ for UKB.

\begin{figure}[!htb]
    \centering
    \includegraphics[width=\linewidth]{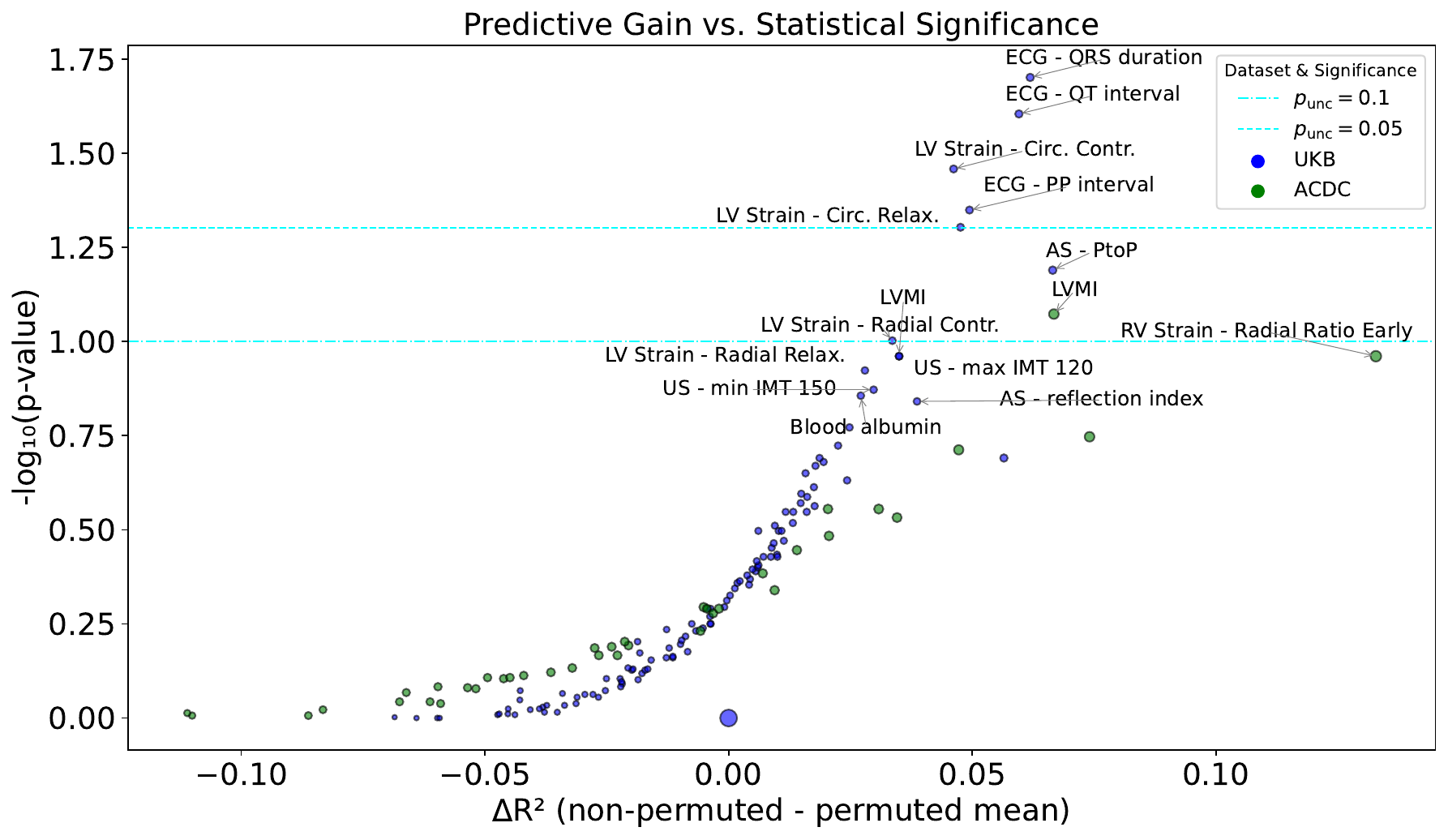}
    \caption{Ridge regression results predicting clinical biomarkers from latent PCA components. The x-axis shows the $R^2$ difference between non-permuted and mean permuted scores; the y-axis shows $-\log_{10}(p_{unc})$ from a permutation test ($n=200$). Biomarkers with higher $R^{2}$ difference and lower p-values lie toward the top-right quadrant. US: ultrasound, ECG: electrocardiogram, AS: arterial stiffness.}
    \label{fig:pca_ridge_regression}
\end{figure}

We trained a ridge regression model using the PCA-transformed latent space to predict a set of functional biomarkers, including myocardial strain and ejection fraction. For the UKB dataset, we also included ECG-derived features, pulse wave analysis (PWA), arterial stiffness, carotid intima-media thickness (IMT) assessed by ultrasound, demographics, blood (e.g., albumin, hemoglobin) and urine biomarker among others. Note that blood and urine biomarkers were collected during a prior visit, often years before the imaging acquisition.
We assessed significance using permutation testing ($n=200$). No models survived FDR multiple comparison correction for multiple biomarkers at nominal $\alpha=0.05$, but uncorrected results consistently prioritized physiologically plausible biomarkers. The uncorrected permutation $p$-values are reported in Figure~\ref{fig:pca_ridge_regression}. The UKB dataset revealed significant predictive structure for several ECG-derived metrics (e.g., QRS duration, PP interval, QT interval) and LV circumferential strain markers, relaxation (from ES to end of cycle) and contraction (ED to ES). Several additional features, including arterial stiffness and ultrasound-based carotid thickness (IMT), showed near-significance. In both datasets, left ventricular mass index (LVMI) appeared among the top predicted features.

As a complementary analysis, we computed univariate Spearman’s rank correlations between each principal component and the same biomarkers.  No correlations survived FDR correction; we report those that were significant at the uncorrected $p < 0.05$ level. The results are available in the Appendix Figure~\ref{fig:acdc_latent_aha_clustermap} and Figure~\ref{fig:ukb_latent_aha_clustermap}, which present the results for ACDC and UKB respectively. Across both datasets, we found weak but consistent associations between latent components and radial and circumferential strain (LV and RV), thickness, as well as stroke volume (SV) and ejection fraction metrics. In the UKB cohort, additional associations were found with ECG-derived metrics, arterial stiffness, ultrasound and PWA.

These results highlight the clinical relevance of the latent representation despite the absence of direct supervision.

\section{Discussion}

Our framework is designed to address the challenge of modeling multivariate time-series with latent structure, while incorporating inductive biases through graph representations. By capturing cardiac dynamics as a latent ODE process, we provide a new perspective for understanding and diagnosing heart conditions using graph-based neural models. This is especially relevant in real-world scenarios where data are sampled at discrete intervals, while the underlying physiological processes evolve continuously in time. Inspired by approaches in reaction-diffusion modeling~\citep{choi_gread_2023} and neural dynamics~\citep{gupta_learning_2022}, our method disentangles spatial and temporal information via directed message passing across multiplex planes. Spatial edges encode anatomical structure, while temporal edges capture transitions between time frames and represent the sequence’s temporal ordering. 

Classical time-series models such as ARIMA~\citep{hamilton_time_2020}, and modern architectures like LSTMs and GRUs, perform well in unstructured sequential prediction tasks. However, they often struggle to model relational structure in spatiotemporal data. Recent work has sought to address this limitation. ST-GCN~\citep{YanXL18}, for example, combines spatial graph convolutions with temporal convolutions, but differs from our approach in key ways—it does not incorporate edge features or directional message passing. Prior work~\citep{Lu2021} modeled LV contours as point clouds and used ST-GCNs to predict future contour positions, but did not explore latent dynamics or leverage the model for diagnostic classification. Furthermore, in their setting, all nodes are always fully connected and edge attributes are ignored.

Our latent space analysis revealed consistent cyclic structures across datasets and connectivity settings, suggesting that the model captures temporal regularities in cardiac dynamics. Using PCA to reduce dimensionality, we identified suggestive associations between latent components and cardiac biomarkers, such as myocardial strain, stroke volume, and ejection fraction. In UKB, the model also captured ECG-derived metrics and vascular markers (e.g., arterial stiffness and ultrasound-based IMT). Notably, we found suggestive associations with QRS duration, QT interval, and PP interval—metrics frequently linked to electrical conduction abnormalities and arrhythmias~\citep{magnani_p_2009, goldenberg_qt_2006, kowlgi_what_2025}. The PP interval, in particular, may reflect atrial rhythm disturbances, making it especially relevant to our atrial fibrillation cohort~\citep{kowlgi_what_2025}. These findings are physiologically plausible and consistent with the clinical profile of the population. Although none of the correlations or regressions survived multiple-comparison correction, the uncorrected results highlight meaningful biological signals. This suggests that our framework encodes clinically relevant structure even without explicit supervision.


Our work has several limitations. First, we evaluated extrapolation only over short time spans, corresponding to a single training window. Due to the nature of ACDC and UKB data, which only include one cardiac cycle per subject, extrapolation on real data could only be qualitatively assessed. Additionally, the use of short-axis images, chosen for compatibility across datasets, may limit performance. Routine clinical acquisitions typically also include long-axis imaging, which provides complementary information about cardiac motion, and could thus offer improved predictive performance.  Another important consideration is the non-real-time nature of cine MRI. These sequences are constructed by averaging data acquired over multiple heartbeats—typically five to ten—to produce a single representative cardiac cycle. As a result, each frame is not a direct snapshot in time but an averaged reconstruction. In patients with arrhythmia such as atrial fibrillation, this averaging process may distort the true dynamics due to beat-to-beat variability. Thus, we hypothesize that our model would perform better when applied to real-time cardiac imaging modalities, which are becoming increasingly common and could enable more faithful assessment of latent cardiac dynamics.

Future work will explore the robustness of our approach and develop interpretability tools to identify which cardiac regions contribute most to each diagnosis. It may also be valuable to explore latent phase plots or impose a structured manifold on the latent space to enhance representation learning. Leveraging predictive uncertainty could further improve performance in clinical decision making. Another promising direction is the end-to-end learning of graph structure~\citep{Shang2021}, potentially guided by anatomical priors such as anatomical connectivity patterns used as regularizer. Advances in higher-order GNNs~\citep{Bodnar2021, Thiede2021}, which move beyond pairwise message passing, could improve the model’s expressivity. While these models can still be framed within a message-passing paradigm~\citep{velickovic2022message}, they offer a path toward supporting mesh-like or manifold-based cardiac representations, without requiring one-to-one node correspondence across subjects.

\section{Conclusions}

To the best of our knowledge, this is the first method to represent cardiac dynamics as a spatiotemporal multiplex graph, integrating both spatial and temporal inductive biases with latent continuous-time dynamics. Our model demonstrates strong performance in trajectory reconstruction, extrapolation, and classification tasks across synthetic and real-world datasets.

We show that the learned latent representations capture physiologically meaningful dynamics and enable accurate classification of cardiac pathologies. On the ACDC challenge dataset, our model achieves performance comparable to state-of-the-art methods. On the UK Biobank cohort, it yields competitive results for atrial fibrillation detection, despite using representations two orders of magnitude smaller than baseline models with extensively tuned hyperparameters.

By framing the cardiac cycle as a graph-based latent ODE process, we unify dynamic modeling, relational structure, and uncertainty quantification within a single framework. Graph-based neural ODEs offer a compelling alternative by jointly learning both the system's dynamics and latent interactions. Neural Processes complement this by modeling distributions over dynamics, enabling uncertainty-aware inference and improved adaptability.

Overall, our architecture is flexible and generalizable, with potential applications beyond cardiology—including other biological systems, physical simulations, and structured control domains.






\section*{Funding}

This research was supported by the Swiss National Science Foundation (grant $\textrm{CRSII5}\_202276/1$).

\section*{CRediT authorship contribution statement}

\textbf{Jaume Banus:} Writing – original draft, Visualization, Validation, Software, Methodology, Data curation, Conceptualization. 
\textbf{Augustin C. Ogier:} Writing – review \& editing.
\textbf{Roger Hullin:} Writing – review \& editing, Funding acquisition. 
\textbf{Philippe Meyer:} Writing – review \& editing, Funding acquisition. 
\textbf{Ruud B. van Heeswijk:} Writing – review \& editing, Funding acquisition. 
\textbf{Jonas Richiardi:} Writing – original draft, Supervision, Conceptualization, Funding acquisition.

\section*{Declaration of competing interest}

The authors declare no competing financial or personal interests that could have influenced the work reported in this paper.

\section*{Data availability}

This research has been conducted using the UK Biobank Resource under Application Number $80108$. Access can be requested at \url{https://www.ukbiobank.ac.uk/}. We also used data from the ACDC challenge, available at \url{https://www.creatis.insa-lyon.fr/Challenge/acdc/}.

\bibliographystyle{IEEEtran}
\bibliography{refs}

\newpage
\appendix
\renewcommand{\thesection}{\Alph{section}} 

\counterwithin{figure}{section}
\counterwithin{table}{section}

\renewcommand{\thefigure}{\thesection.\arabic{figure}}
\renewcommand{\thetable}{\thesection.\arabic{table}}

\section{Cohort Selection: UK Biobank}
\label{s:appendix:cohort_ukb}

\begin{table}[!htbp]
\setlength\tabcolsep{3pt} 
\caption{ICD-10 codes for AFib selection and exclusion criteria in UK Biobank.}
\label{tab:ICD_codes}
\centering
\resizebox{\columnwidth}{!}{%
\begin{tabular}{|l|l|}
\hline
\textbf{Category}                      & \textbf{ICD-10 Codes}                                                                                         \\ \hline
\textbf{Targeted Condition}            &                                                                                                                \\ \hline
Atrial Fibrillation (AFib)             & I48.0, I48.1, I48.2, I48.3, I48.4, I48.9                                                                        \\ \hline
\textbf{Conditions Considered for Exclusion} &                                                                                                                \\ \hline
Ischemic Heart Disease                 & I20, I21, I22, I23, I24, I25                                                                                   \\ \hline
Cerebrovascular Disease                & I60, I61, I62, I63, I64, I65, I66, I67, I68, I69                                                               \\ \hline
Hypertensive Disease                   & I10, I11, I12, I13, I14, I15, I16                                                                              \\ \hline
Arterial Diseases                      & I70, I71, I72, I73, I74, I75, I76, I77, I78, I79                                                               \\ \hline
Venous Diseases                        & I80, I81, I82, I83, I84, I85, I86, I87, I88, I89                                                               \\ \hline
Pulmonary Heart Disease                & I26, I27, I28                                                                                                  \\ \hline
Dilated Cardiomyopathy (DCM)           & I42.0                                                                                                          \\ \hline
Hypertrophic Cardiomyopathy (HCM)      & I42.1, I42.2                                                                                                   \\ \hline
Heart Failure                          & I50.0, I50.1, I50.2, I50.3, I50.4, I50.5, I50.6, I50.9                                                         \\ \hline
Stroke                                 & I63.0, I63.1, I63.2, I63.3, I63.4, I63.5, I63.6, I63.7, I63.8, I63.9                                            \\ \hline
\end{tabular}%
}
\end{table}
\FloatBarrier

\section{Hyperparameter Search Space for our Model}

\begin{table}[!htbp]
\setlength\tabcolsep{3pt} 
\caption{Hyperparameter ranges for our model.}
\label{tab:hyperparameters_multiplex}
\centering
\resizebox{0.6\columnwidth}{!}{%
\begin{tabular}{|l|c|}
\hline
\textbf{Hyperparameter} & \textbf{Range} \\
\hline
weight decay & $10^{-4} - 10^{-1}$ \\
$\beta_{1}$ & $0.5 - 1$ \\
$\beta_{2}$ & $0 - 0.5$ \\
$\beta_{3}$ & $0 - 0.5$ \\
hidden dim ($L_{0} + D_{0}$) & $12 - 25$ \\
latent dim ($L_{0}$) & $2 - 11$ \\
spatial planes & $2 - 6$ \\
temporal planes & $2 - 6$ \\
depth node encoders (state, control) & $1 - 2$ \\
depth edge encoders (spatial, temporal) & $1 - 2$ \\
decode just latent & [True, False] \\
\hline
\end{tabular}%
}
\end{table}

\begin{table}[!htbp]
\setlength\tabcolsep{6pt} 
\caption{Selected optimal hyperparameters for each dataset and setting.}
\label{tab:hyperparams_multiplex_values}
\centering
\resizebox{1\columnwidth}{!}{%
\begin{tabular}{|l|c|c|c|c|c|c|c|}
\hline
\textbf{Hyperparameter} & \textbf{Pendulum} & \textbf{Lorenz} & \textbf{Kuramoto} & \multicolumn{2}{c|}{\textbf{ACDC}} & \multicolumn{2}{c|}{\textbf{UKB}} \\
\cline{5-8}
                        &          &        &          & Full   & Anat                 & Full   & Anat                \\
\hline
$L_0$ Dimension         & 5        & 7       & 6       & 11     & 5                   & 6      & 7                  \\
$D_0 + L_{0}$ Dimension & 17       & 20      & 17      & 22     & 21                  & 17     & 18                 \\
Spatial Planes          & 2        & 2       & 5       & 4      & 5                   & 5      & 6                  \\
Temporal Planes         & 5        & 5       & 5       & 3      & 2                   & 3      & 5                  \\
Depth node encoders     & 1        & 1       & 1       & 2      & 2                   & 2      & 2                  \\
Depth edge encoders     & 1        & 2       & 1       & 2      & 2                   & 1      & 2                  \\
$\beta_1$               & 0.93     & 0.88    & 0.90    & 0.57   & 0.85                & 0.70   & 0.69               \\
$\beta_2$               & 0.25     & 0.14    & 0.29    & 0.36   & 0.31                & 0.33   & 0.32               \\
$\beta_3$               & 0.22     & 0.10    & 0.25    & 0.35   & 0.34                & 0.27   & 0.075              \\
Weight Decay            & 0.002    & 0.00003 & 0.0009  & 0.005  & 0.0004              & 0.005  & 0.00027            \\
Decode Just Latent      & \xmark   & \xmark  & \xmark  & \xmark & \xmark              & \xmark & \xmark              \\
\hline
\end{tabular}%
}
\end{table}
\FloatBarrier

\section{Hyperparameter Search Space for Classifiers}
\label{s:hyperparametersOfClassifiers}

\begin{table}[!htbp]
\setlength\tabcolsep{3pt} 
\caption{Hyperparameter ranges for Random Forest, XGBoost, and $k$NN classifiers used in UK Biobank and ACDC experiments. $k$NN: K-Nearest neighbors, RF: Random forest, BS: balanced subsample, num.: number of, reg.: regularisation.}
\label{tab:hyperparameters}
\centering
\resizebox{0.6\columnwidth}{!}{%
\begin{tabular}{|l|l|c|}
\hline
\textbf{Method} & \textbf{Hyperparameter} & \textbf{Range} 
\\
\hline
\multirow{5}{*}{$k$NN} & num. neighbors & $3-6$ \\
 & p & $[1, 2]$ \\
 & leaf size & $[20, 30, 40]$ \\
 & weights & [uniform, distance] \\
\hline
\multirow{6}{*}{XGBoost} & reg. alpha & $0.1-0.9$ \\
 & reg. lambda & $0.1-0.9$ \\
 & subsample & $[0.5, 0.75]$ \\
 & max depth & $[10, 20]$ \\
 & num. estimators & $[601, 1001]$ \\
\hline
\multirow{3}{*}{RF} & criterion & [gini, entropy, log loss] \\
 & num. estimators & $[601, 1001]$ \\
 & class weight & [balanced, BS] \\
 & max depth & [10, 20] \\
\hline
\end{tabular}%
}
\end{table}
\FloatBarrier

\section{Latent space mass univariate analysis}

\begin{figure}[!htbp]
    \centering
    \includegraphics[width=0.7\linewidth]{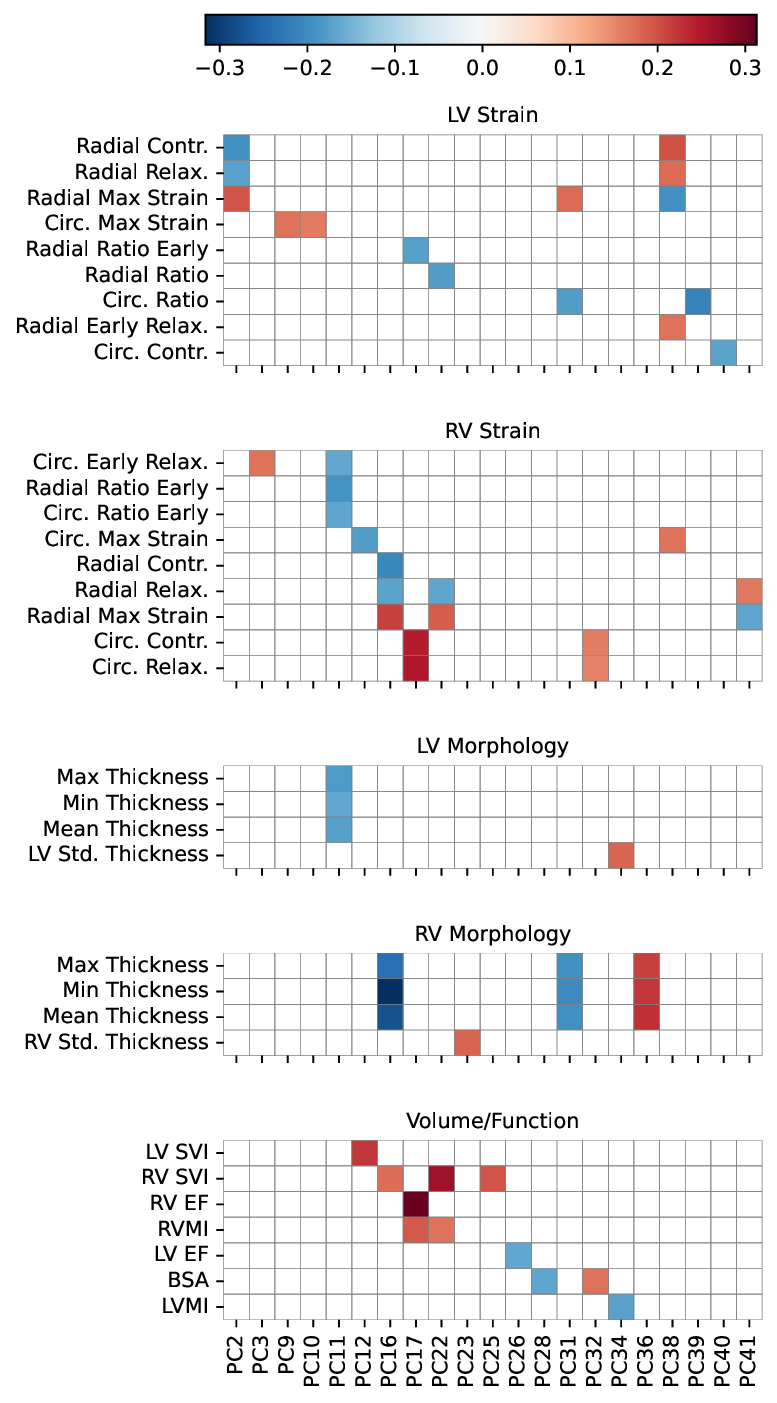}
    \caption{Significant Spearman's rank correlations in the ACDC dataset between principal components of the latent representation and functional biomarkers ($p_\text{uncorrected} < 0.05$). Circ.: circumferential, Relax.: relaxation, Contr.: contraction, Std.: standard deviation.}
    \label{fig:acdc_latent_aha_clustermap}
\end{figure}

\begin{figure}[!htbp]
    \centering
    \includegraphics[width=1.\linewidth]{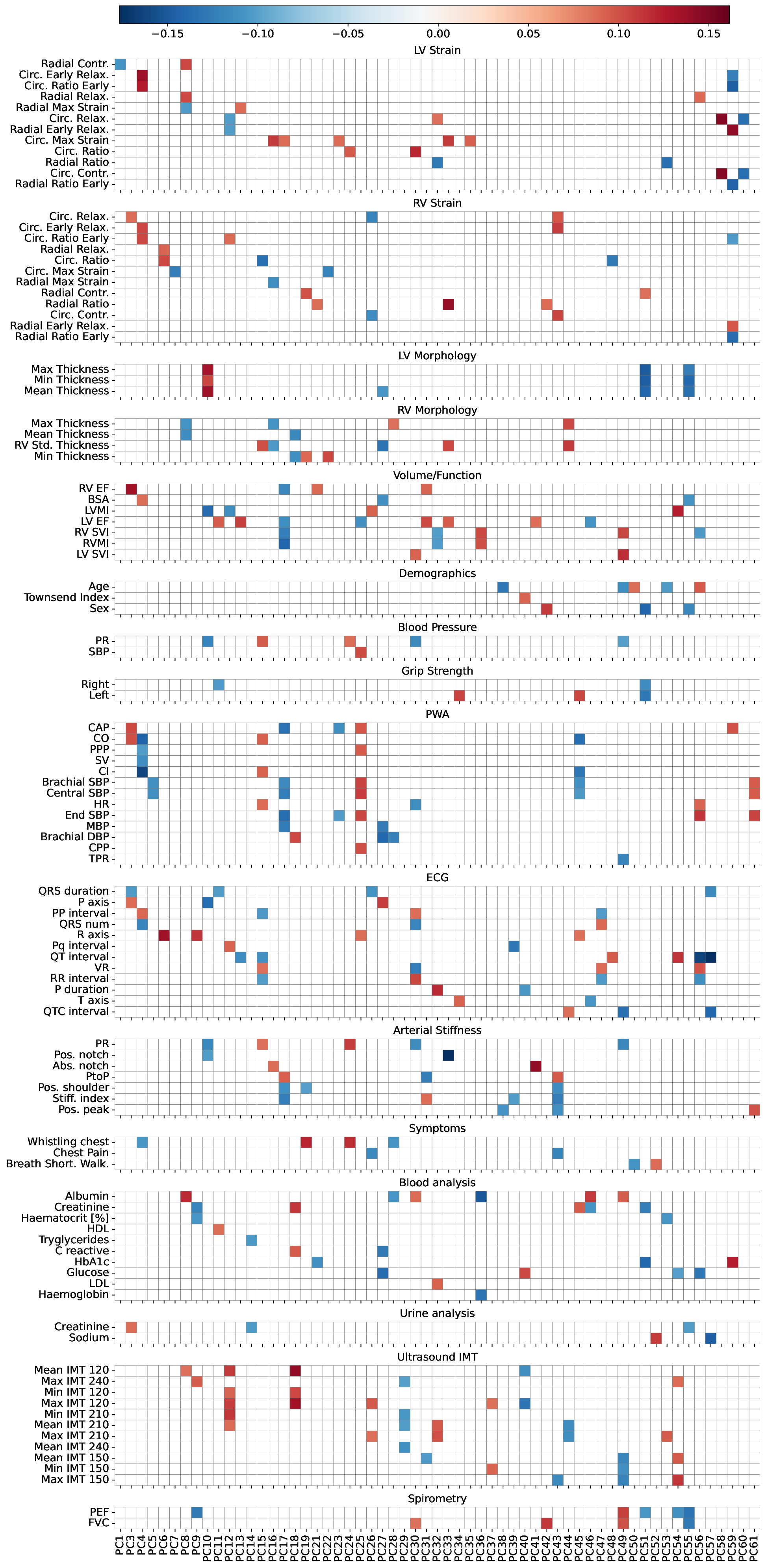}
    \caption{Significant Spearman's rank correlations in the UKB dataset between principal components of the latent representation and functional biomarkers ($p_\text{uncorrected} < 0.05$). Circ.: circumferential, Relax.: relaxation, Contr.: contraction, Std.: standard deviation, PR: pulse rate, SBP: systolic blood pressure, CAP: central augmented pressure, CO: cardiac output, PPP: peripheral pulse pressure, CI: cardiac index, MBP: mean blood pressure, DBP: diastolic blood pressure, CPP: central pulse pressure, TPR: total peripheral resistance, VR: ventricular rate, Pos.: position, Abs.: absence, PtoP: peak-to-peak, Stiff.: sitffness, Short.: shortness, Walk.: walking, IMT: intima-media thickness, PEF: peak expiratory flow, FVC: forced vital capacity. Blood and urine analysis correspond to were obtained in previous visit (not the imaging one).}
    \label{fig:ukb_latent_aha_clustermap}
\end{figure}
\FloatBarrier

\section{Cardiac Division and Graph Construction}

\begin{figure}[!htbp]
	\centering
	\begin{subfigure}[t]{0.98\columnwidth}
		\includegraphics[width=\linewidth]{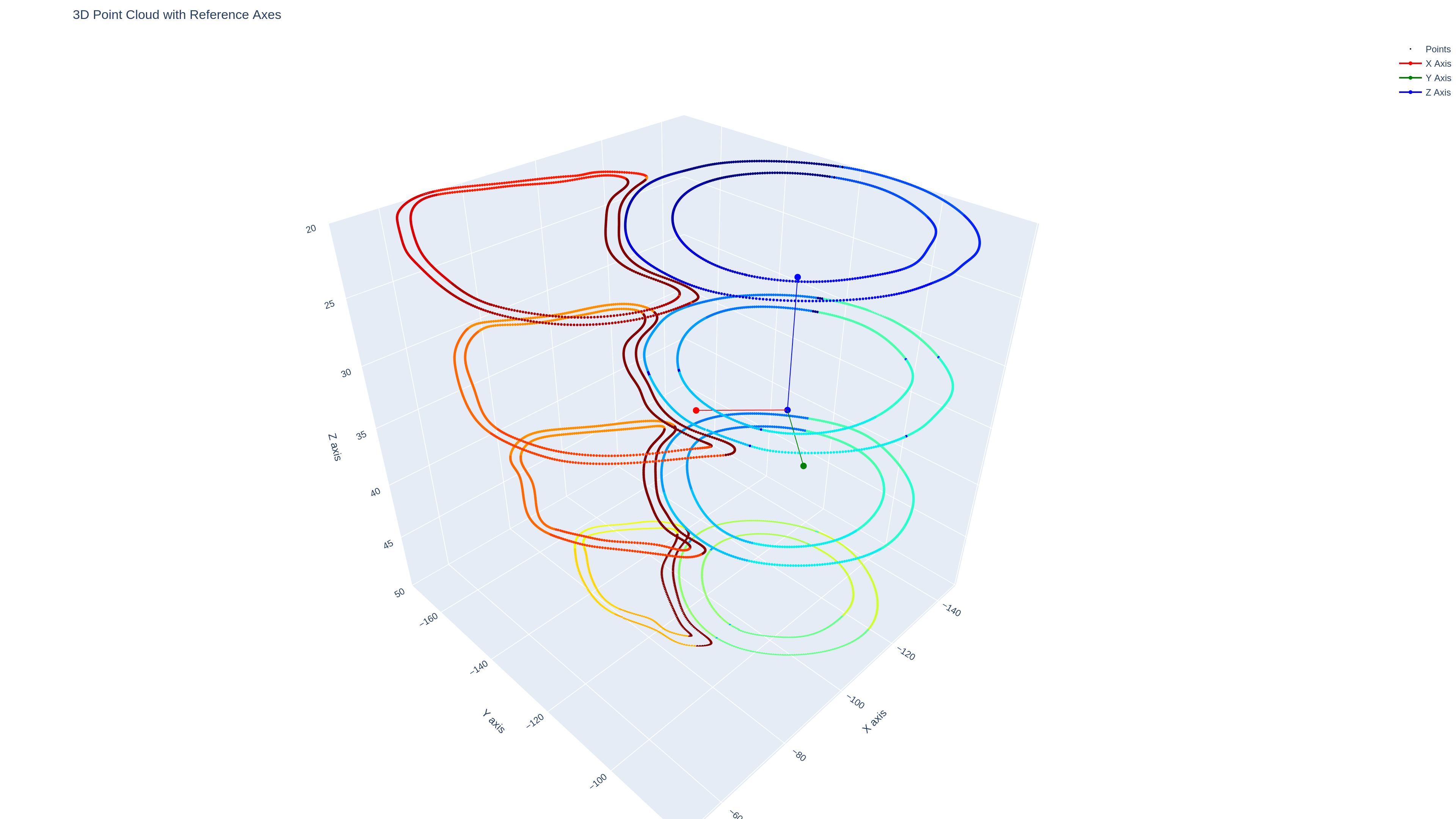}
		\caption{Example cardiac boundaries for a subject.}
		\label{subfig:cardiac_shape}
	\end{subfigure}
	\hfill
	\begin{subfigure}[t]{0.60\columnwidth}
		\includegraphics[width=\linewidth]{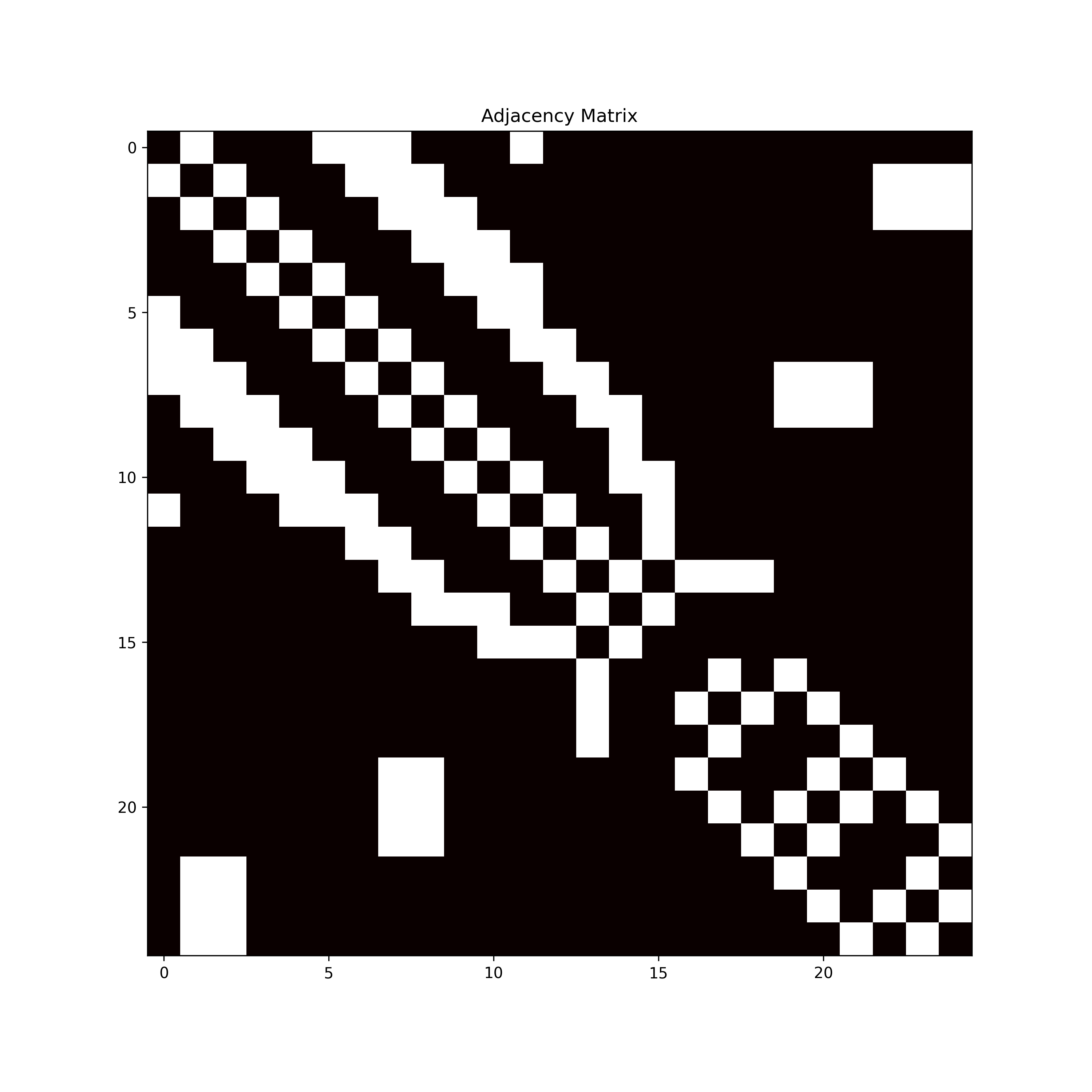}
		\caption{Anat connectivity adjacency matrix.}
		\label{subfig:aha_division}
	\end{subfigure}
	\hfill
	\begin{subfigure}[t]{0.98\columnwidth}
		\includegraphics[width=\linewidth]{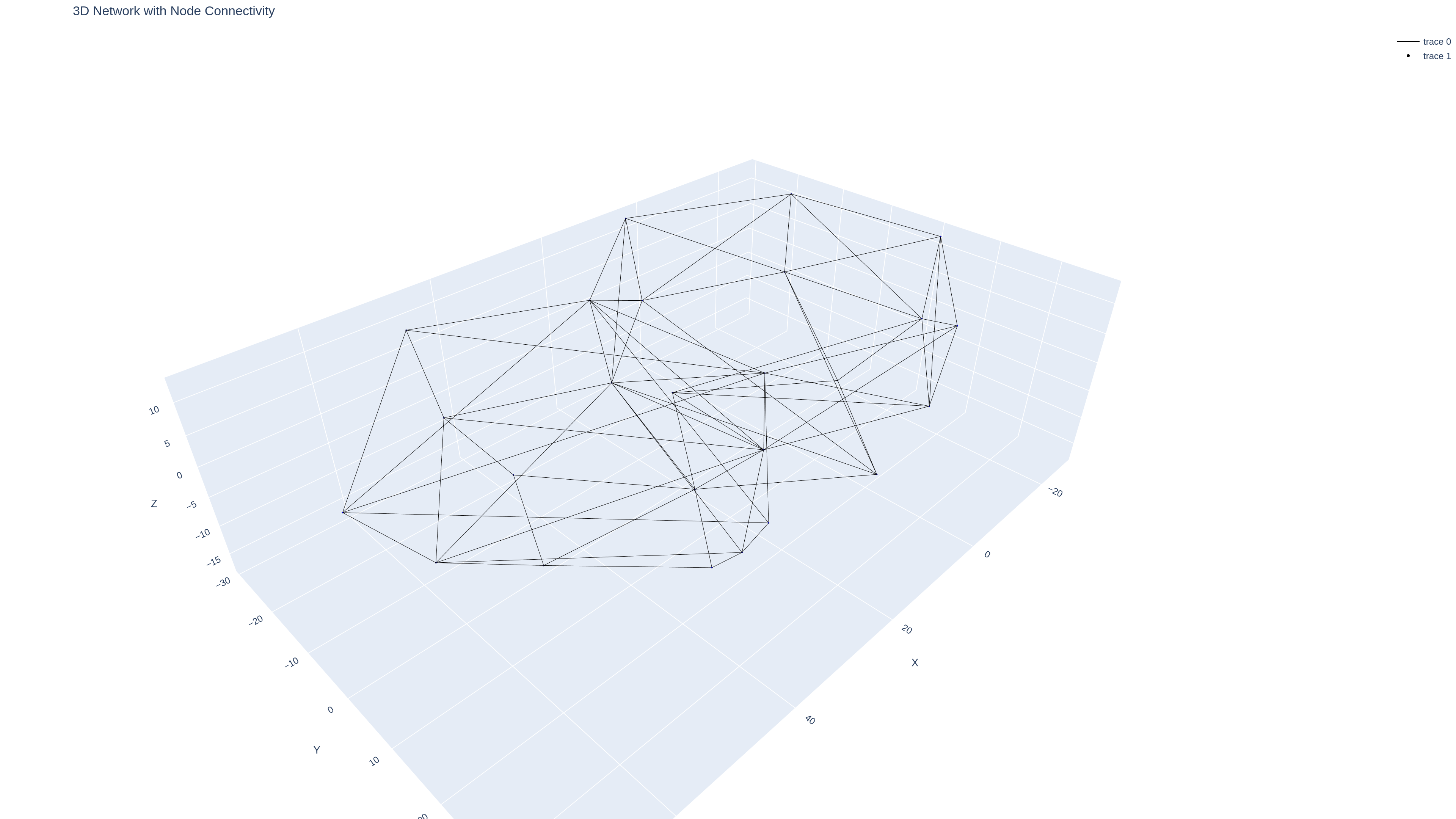}
		\caption{Graph representation.}
		\label{subfig:aha_graph}
	\end{subfigure}
	
	\caption{a) Representation of cardiac boundaries. The colors represent different cardiac regions. The axes indicate a local coordinate system defined by each heart: 
	the short axis (center of mass of the left ventricle to the right ventricle), the long axis (from apex to base), 
	and their cross-product. b) Adjacency matrix for the Anat connectivity, connecting anatomically adjacent regions. c) Graph representation generated following the Anat adjacency.}
	\label{fig:cardiac_representation}
\end{figure}

This section visualizes the anatomical segmentation and adjacency scheme used for graph construction. In Figure~\ref{fig:cardiac_representation} we present a) region-level boundaries; b) the adjacency matrix used for anatomical connectivity; and c) the corresponding graph structure visualized.

\FloatBarrier

\section{Group-Level Trajectories}


\begin{figure*}[tbp]
    \centering
    \begin{minipage}{0.33\textwidth}
        \centering
        \includegraphics[width=\linewidth]{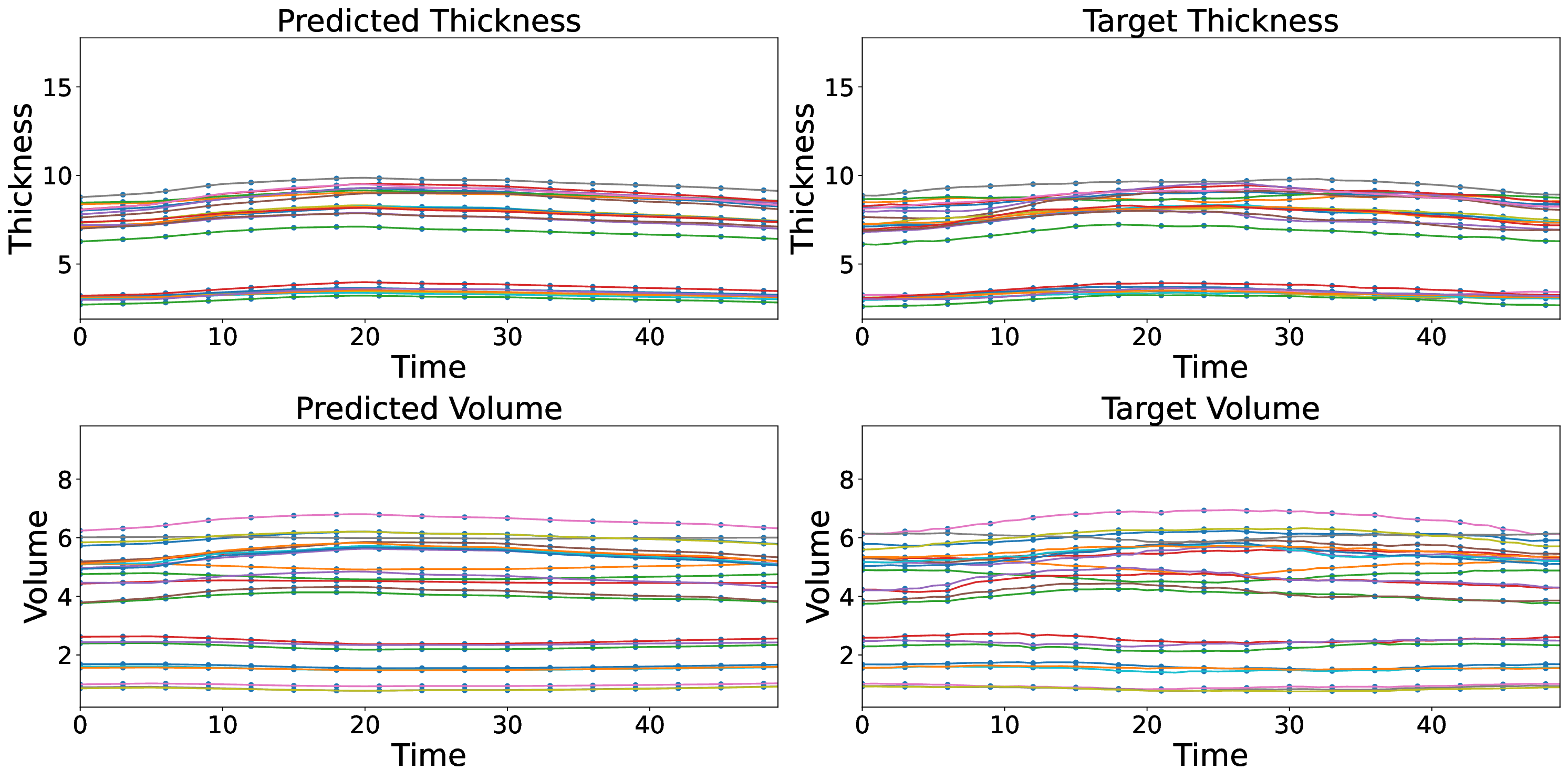}
        \subcaption{DCM Trajectory}
        \label{subfig:trajectory_DCM_all}
    \end{minipage}
    \hfill
    \begin{minipage}{0.33\textwidth}
        \centering
        \includegraphics[width=\linewidth]{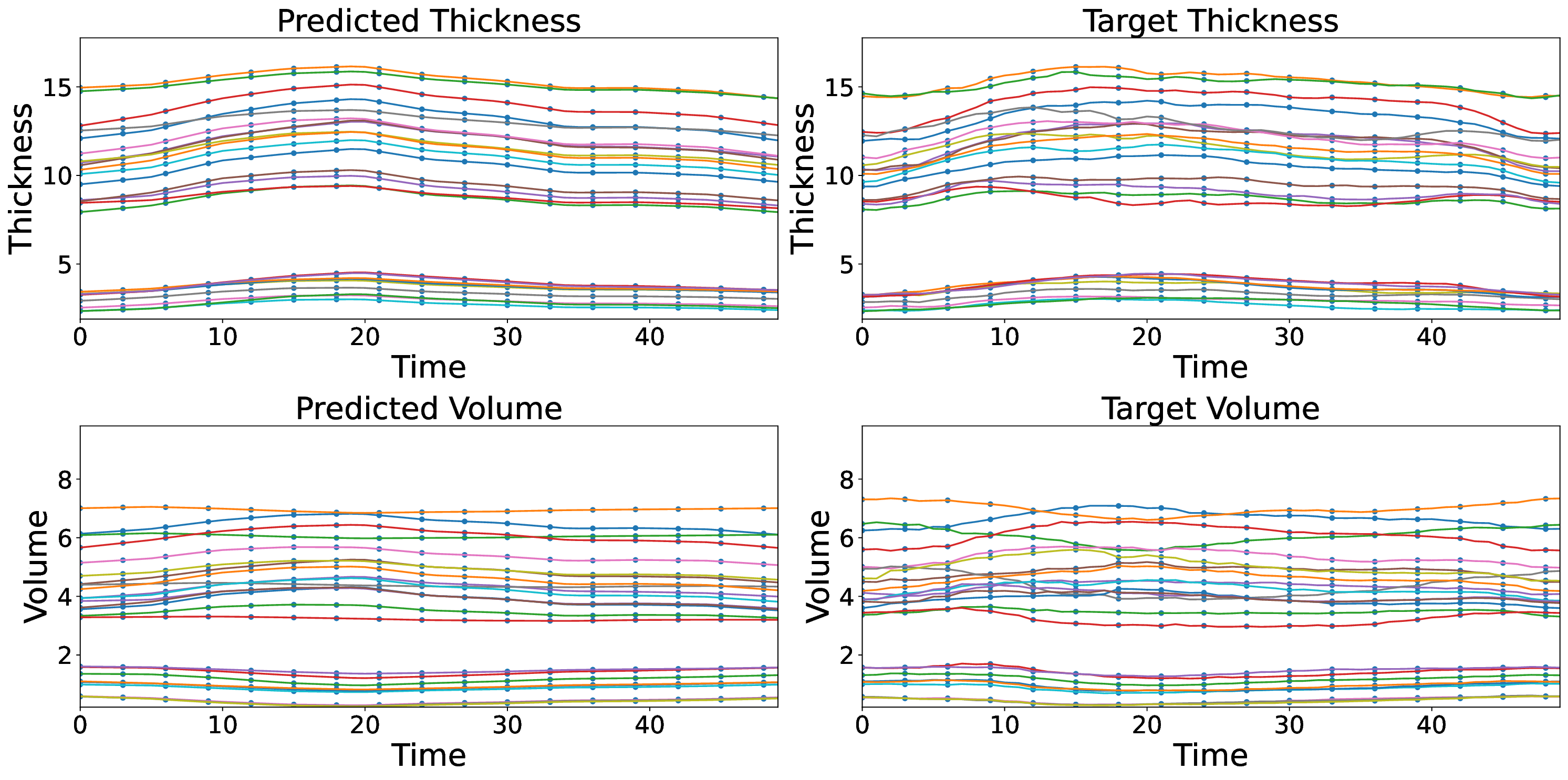}
        \subcaption{HCM Trajectory}
        \label{subfig:trajectory_HCM_all}
    \end{minipage}
    \hfill
    \begin{minipage}{0.33\textwidth}
        \centering
        \includegraphics[width=\linewidth]{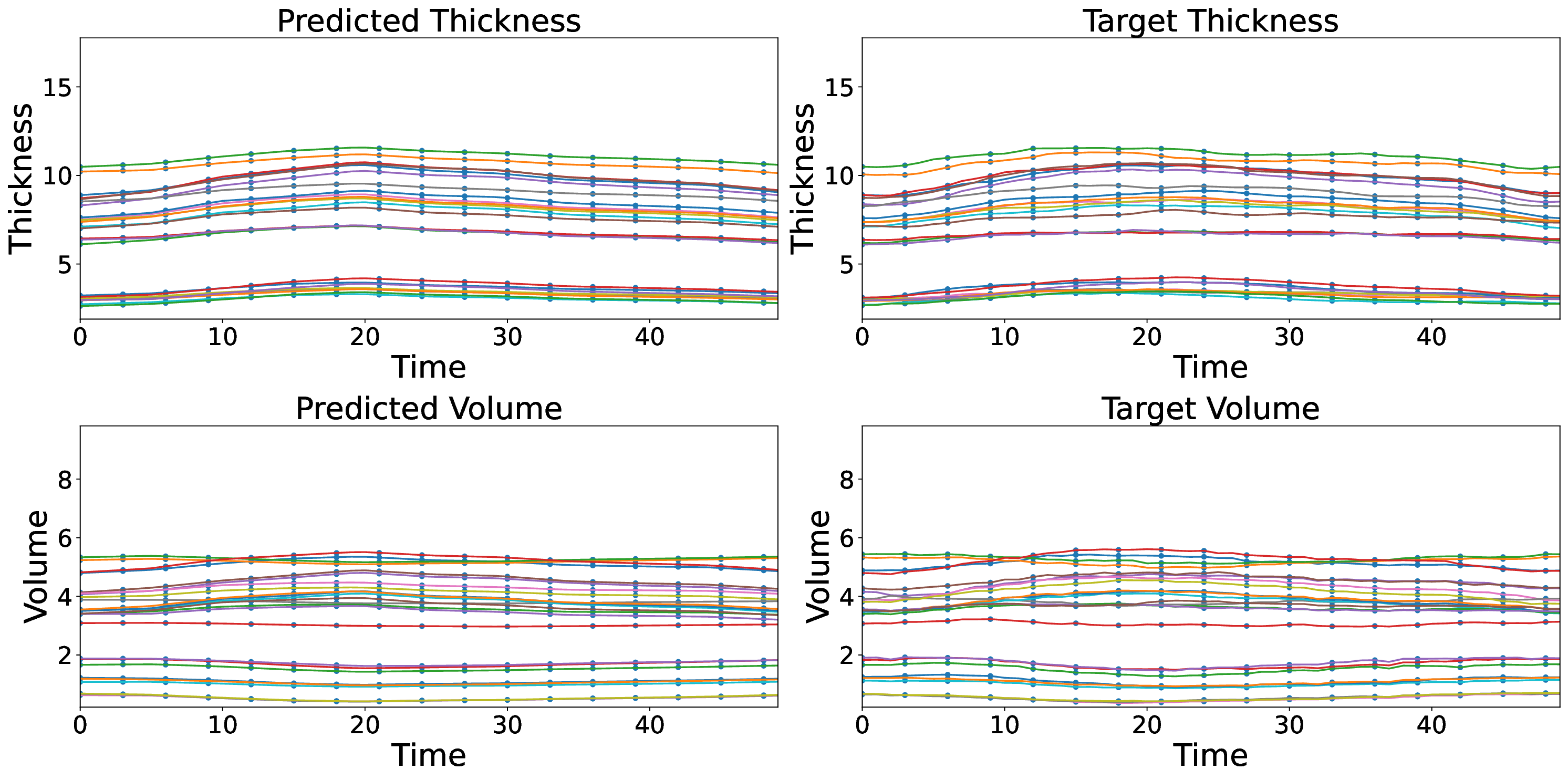}
        \subcaption{MINF Trajectory}
        \label{subfig:trajectory_MINF_all}
    \end{minipage}
    
    \vspace{1em} 
    
    \begin{minipage}{0.33\textwidth}
        \centering
        \includegraphics[width=\linewidth]{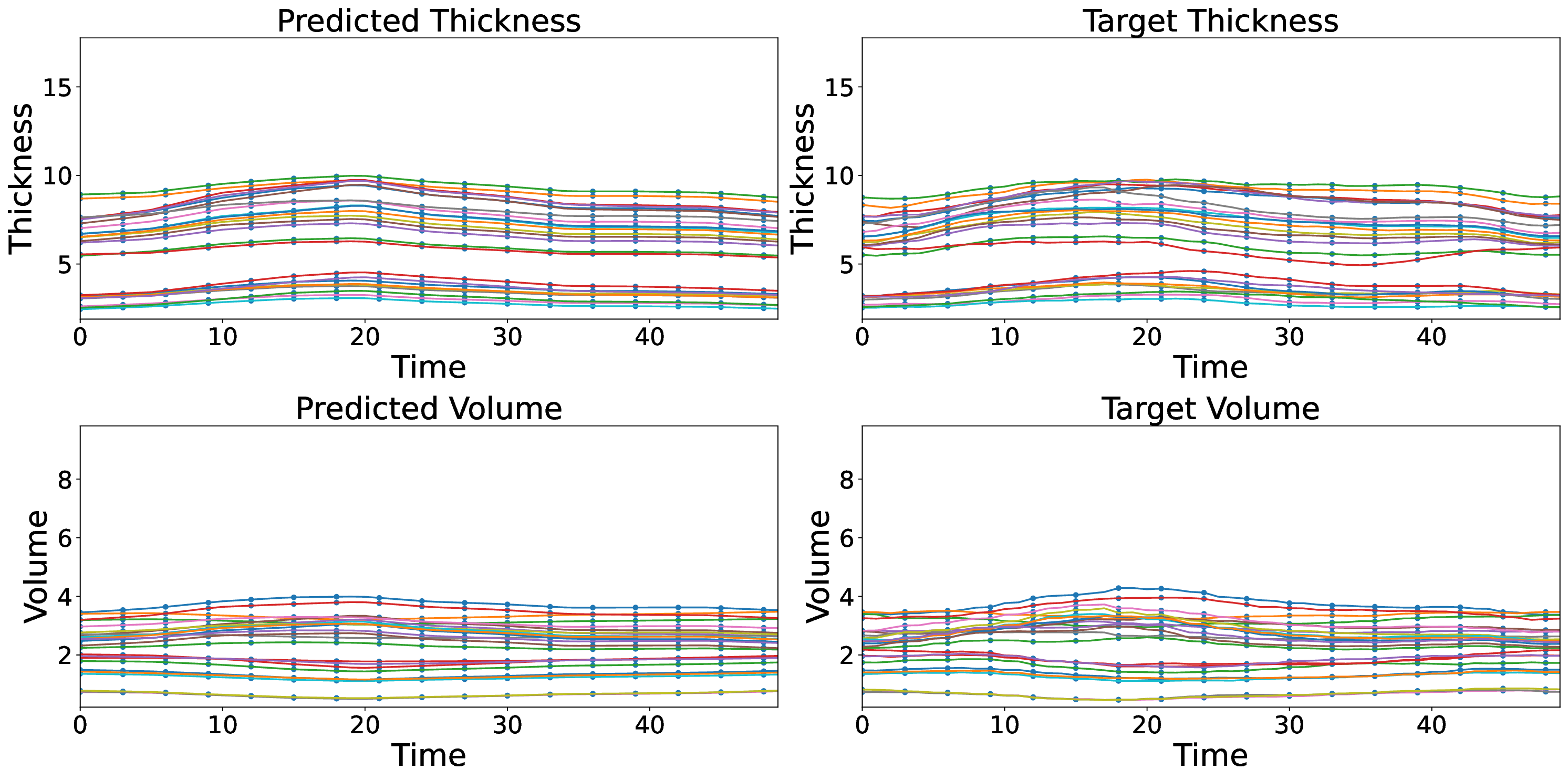}
        \subcaption{NOR Trajectory}
        \label{subfig:trajectory_NOR_all}
    \end{minipage}
    \begin{minipage}{0.33\textwidth}
        \centering
        \includegraphics[width=\linewidth]{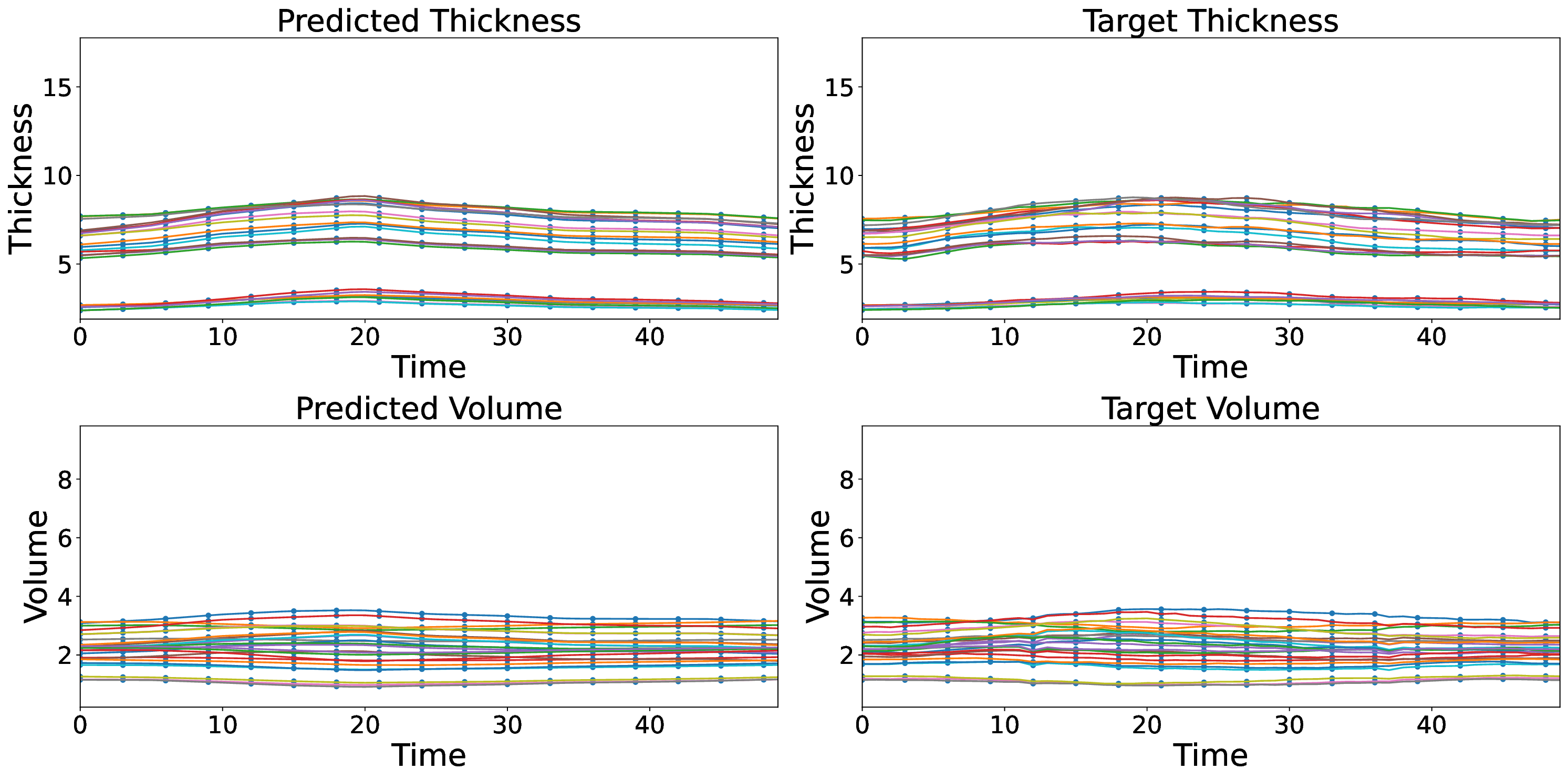}
        \subcaption{ARV Trajectory}
        \label{subfig:trajectory_RV_all}
    \end{minipage}
    
    \caption{Group trajectories on ACDC dataset using fully connected graph (Full). DCM: Dilated cardiomyopathy, HCM: Hypertrophic cardiomyopathy, MINF: Myocardial infarction, NOR: Normal, ARV: Abnormal right-ventricle.}
    \label{fig:group_trajectories_all_acdc}
\end{figure*}

\begin{figure*}[tbp]
    \centering
    \begin{minipage}{0.33\textwidth}
        \centering
        \includegraphics[width=\linewidth]{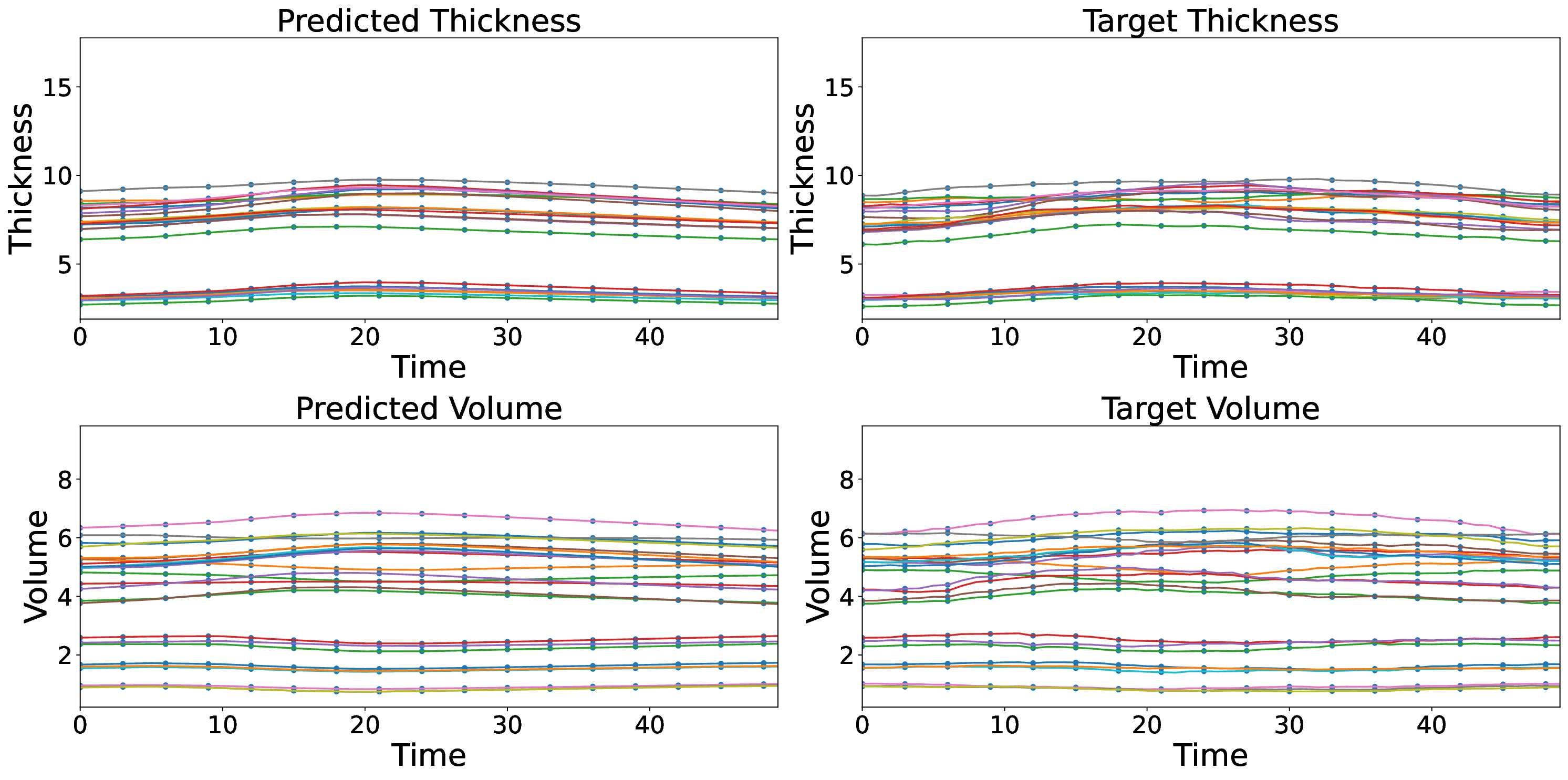}
        \subcaption{DCM Trajectory}
        \label{subfig:trajectory_DCM_aha}
    \end{minipage}
    \hfill
    \begin{minipage}{0.33\textwidth}
        \centering
        \includegraphics[width=\linewidth]{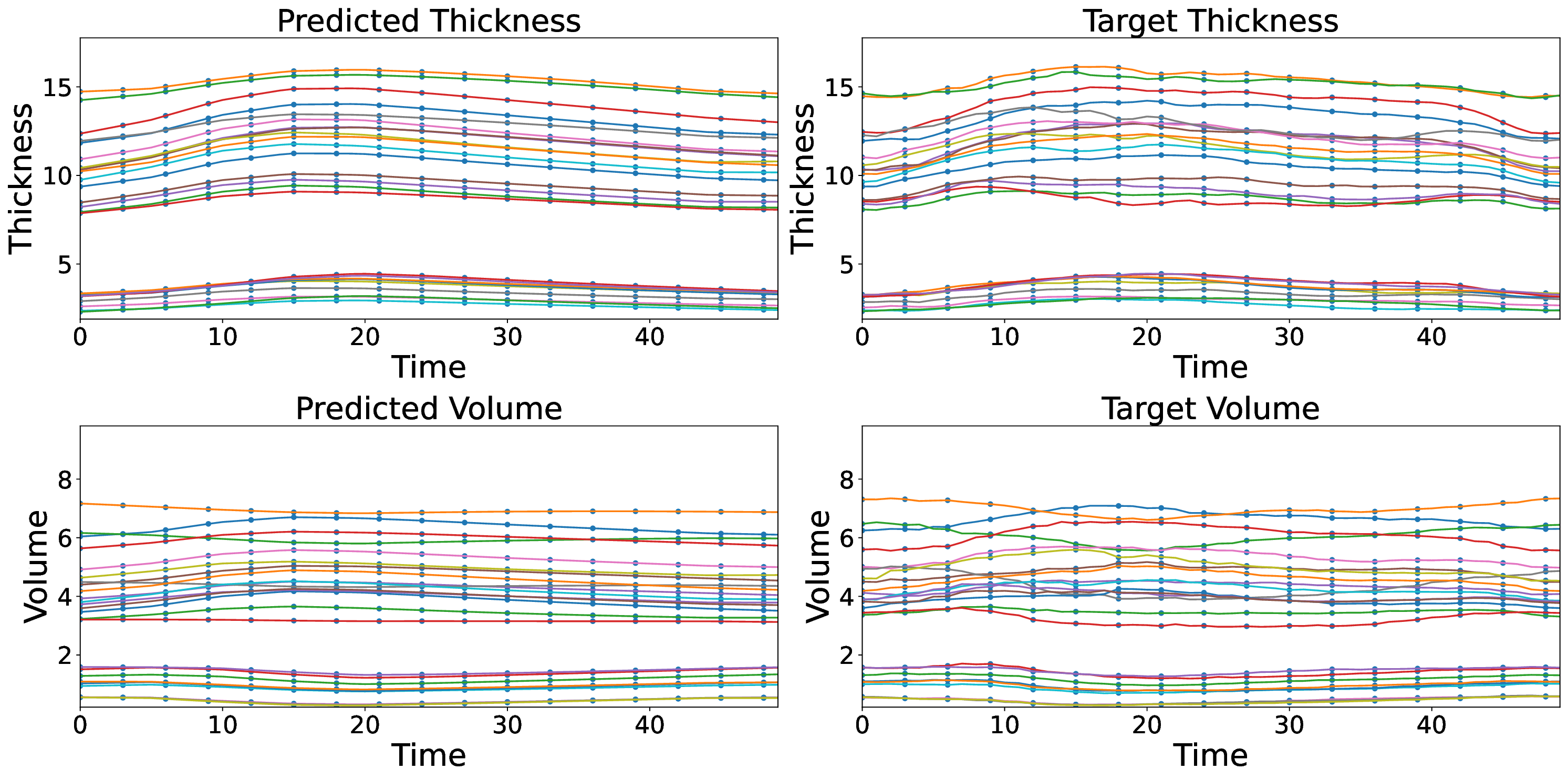}
        \subcaption{HCM Trajectory}
        \label{subfig:trajectory_HCM_aha}
    \end{minipage}
    \hfill
    \begin{minipage}{0.33\textwidth}
        \centering
        \includegraphics[width=\linewidth]{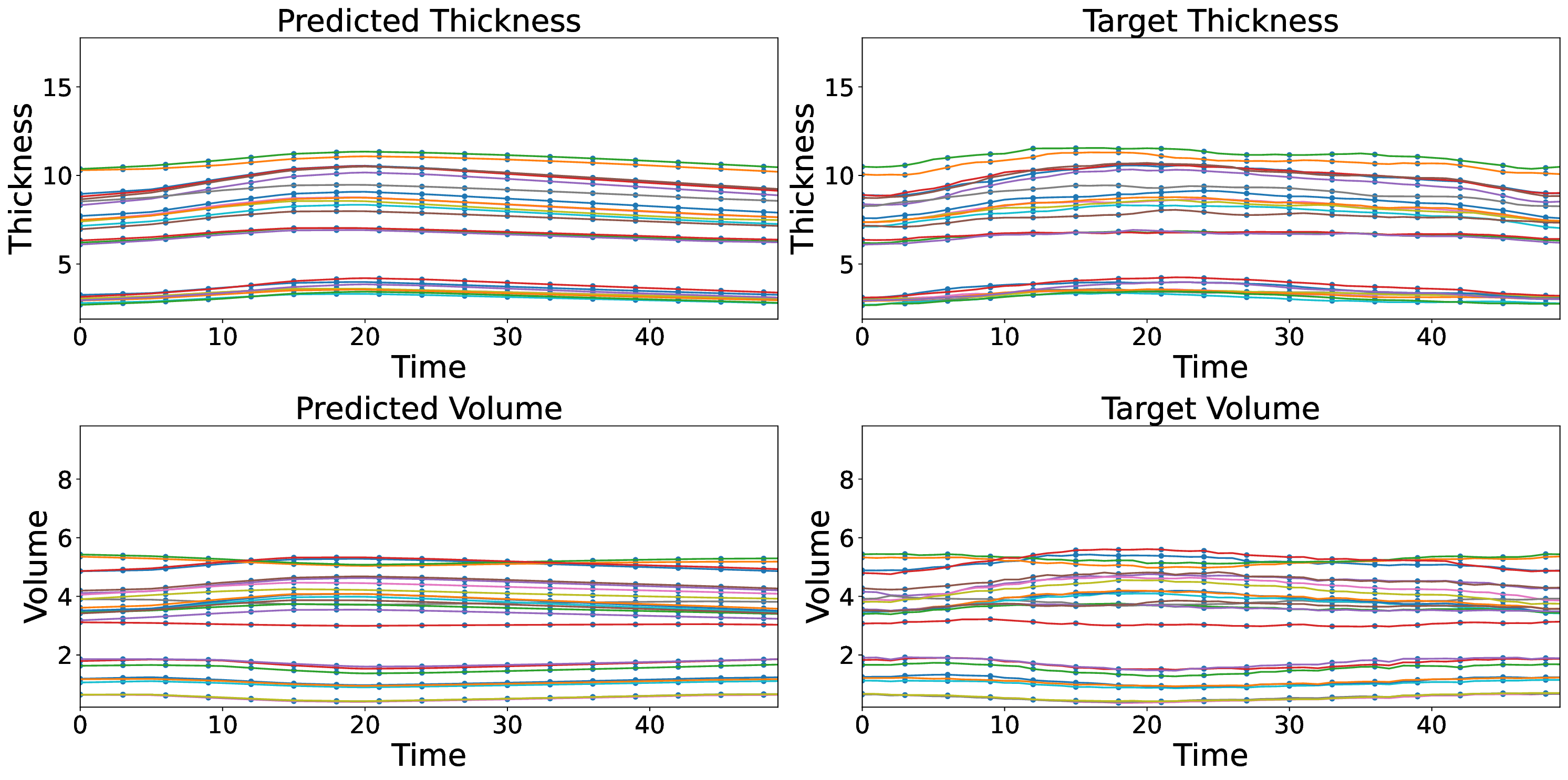}
        \subcaption{MINF Trajectory}
        \label{subfig:trajectory_MINF_aha}
    \end{minipage}
    
    \vspace{1em} 
    
    \begin{minipage}{0.33\textwidth}
        \centering
        \includegraphics[width=\linewidth]{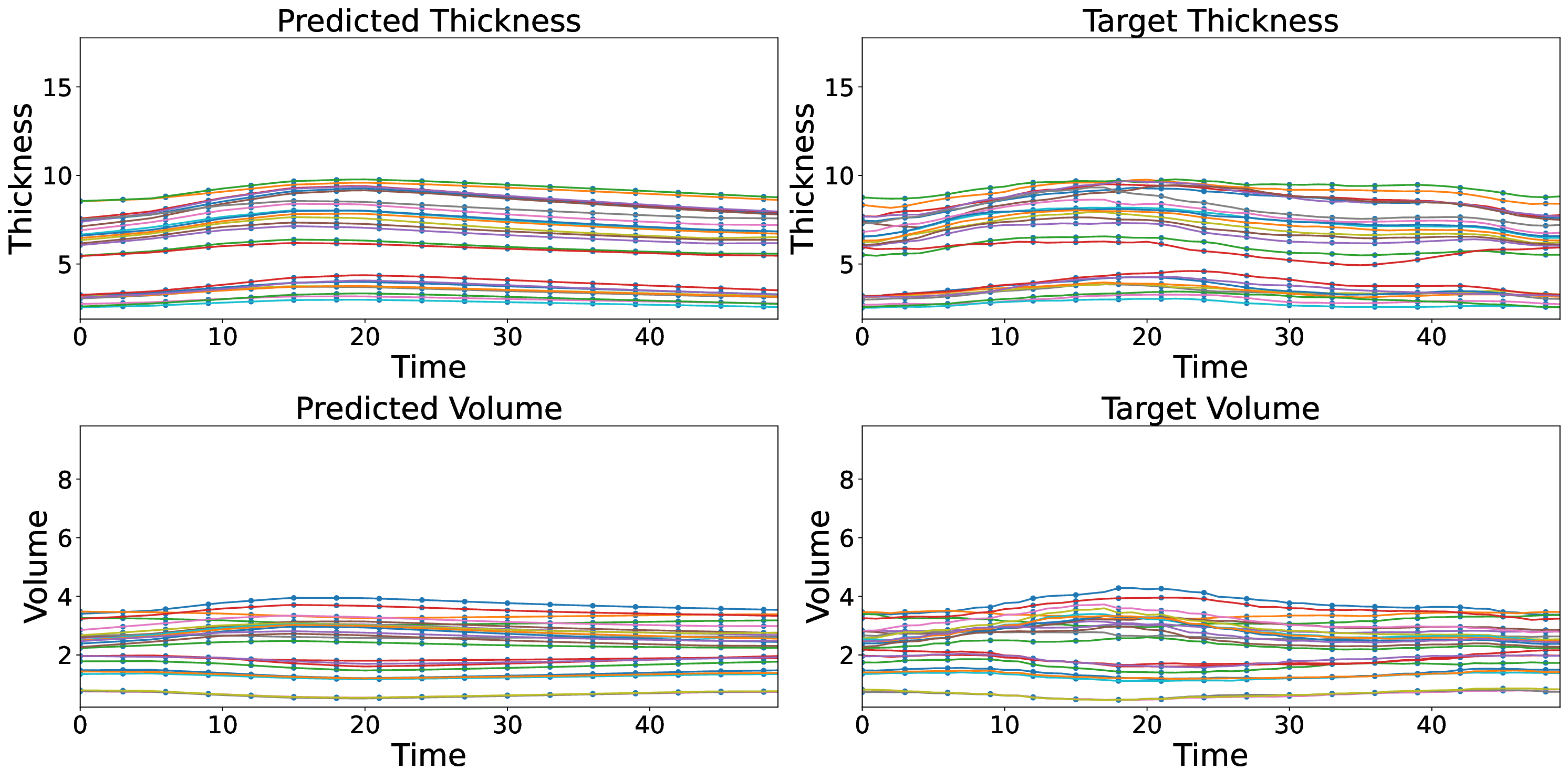}
        \subcaption{NOR Trajectory}
        \label{subfig:trajectory_NOR_aha}
    \end{minipage}
    \begin{minipage}{0.33\textwidth}
        \centering
        \includegraphics[width=\linewidth]{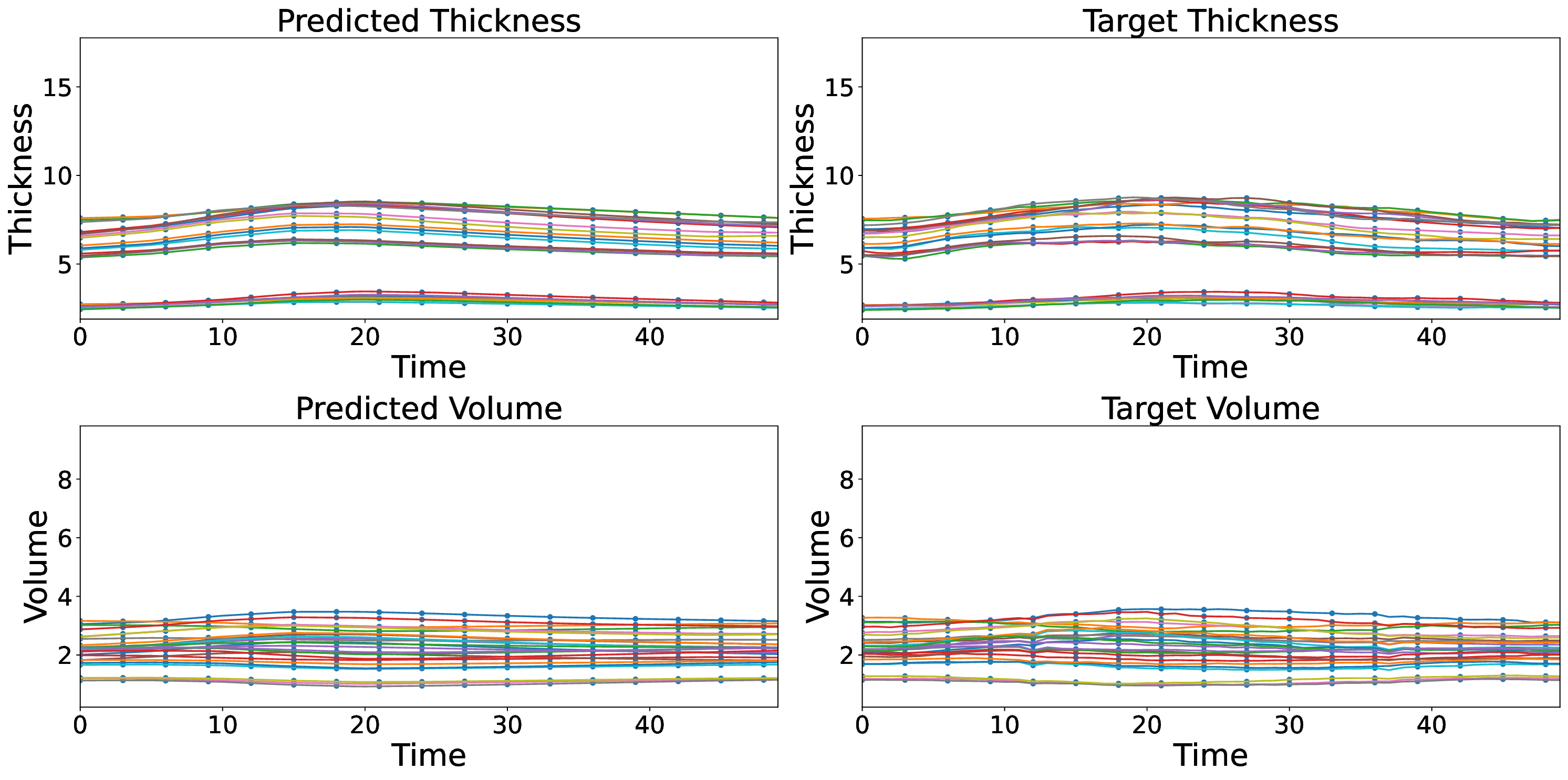}
        \subcaption{ARV Trajectory}
        \label{subfig:trajectory_RV_aha}
    \end{minipage}
    
    \caption{Group trajectories on ACDC dataset using anatomical connectivity (Anat). DCM: Dilated cardiomyopathy, HCM: Hypertrophic cardiomyopathy, MINF: Myocardial infarction, NOR: Normal, ARV: Abnormal right-ventricle.}
    \label{fig:group_trajectories_aha_acdc}
\end{figure*}

\clearpage


\begin{figure*}[tbp]
    \centering
    \begin{minipage}{0.4\textwidth}
        \centering
        \includegraphics[width=\linewidth]{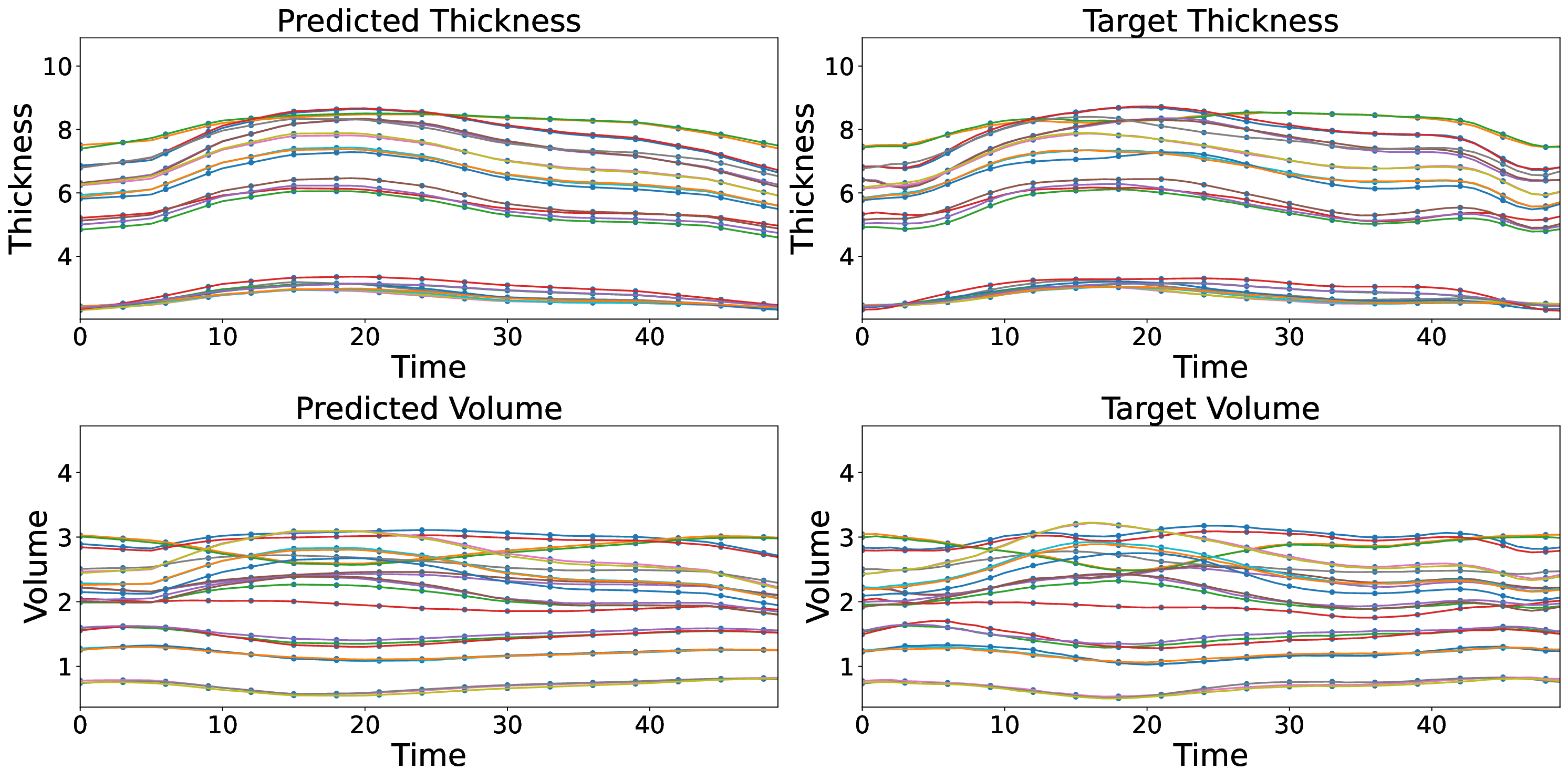}
        \subcaption{Control Trajectory}
        \label{subfig:trajectory_control_all}
    \end{minipage}
    \begin{minipage}{0.4\textwidth}
        \centering
        \includegraphics[width=\linewidth]{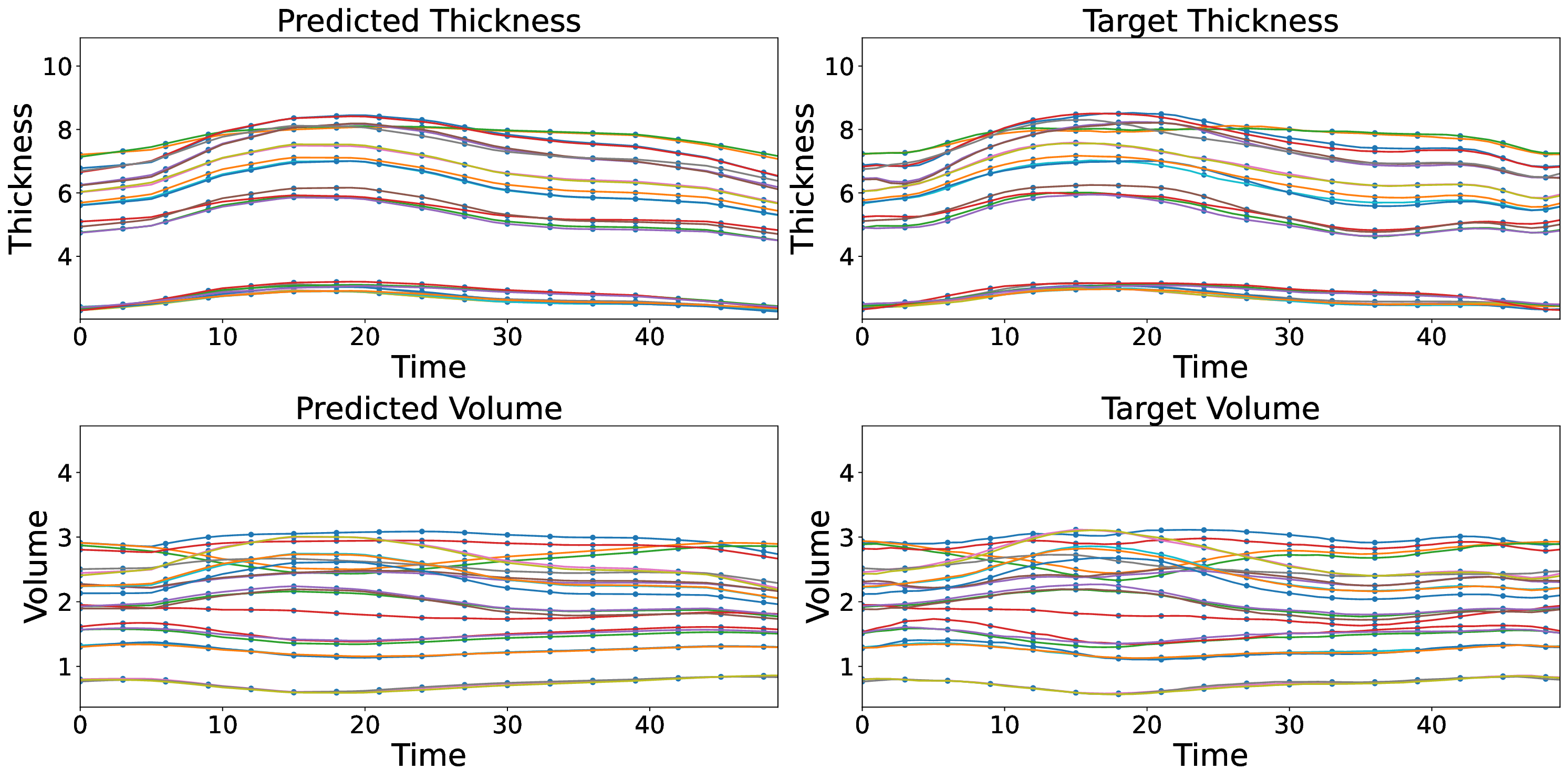}
        \subcaption{AFib Trajectory}
        \label{subfig:trajectory_afib_all}
    \end{minipage}    
    \caption{Group trajectories on UKB dataset using fully connected graph (Full). AFib: Atrial fibrillation.}
    \label{fig:group_trajectories_all_ukb}
\end{figure*}


\begin{figure*}[tbp]
    \centering
    \begin{minipage}{0.4\textwidth}
        \centering
        \includegraphics[width=\linewidth]{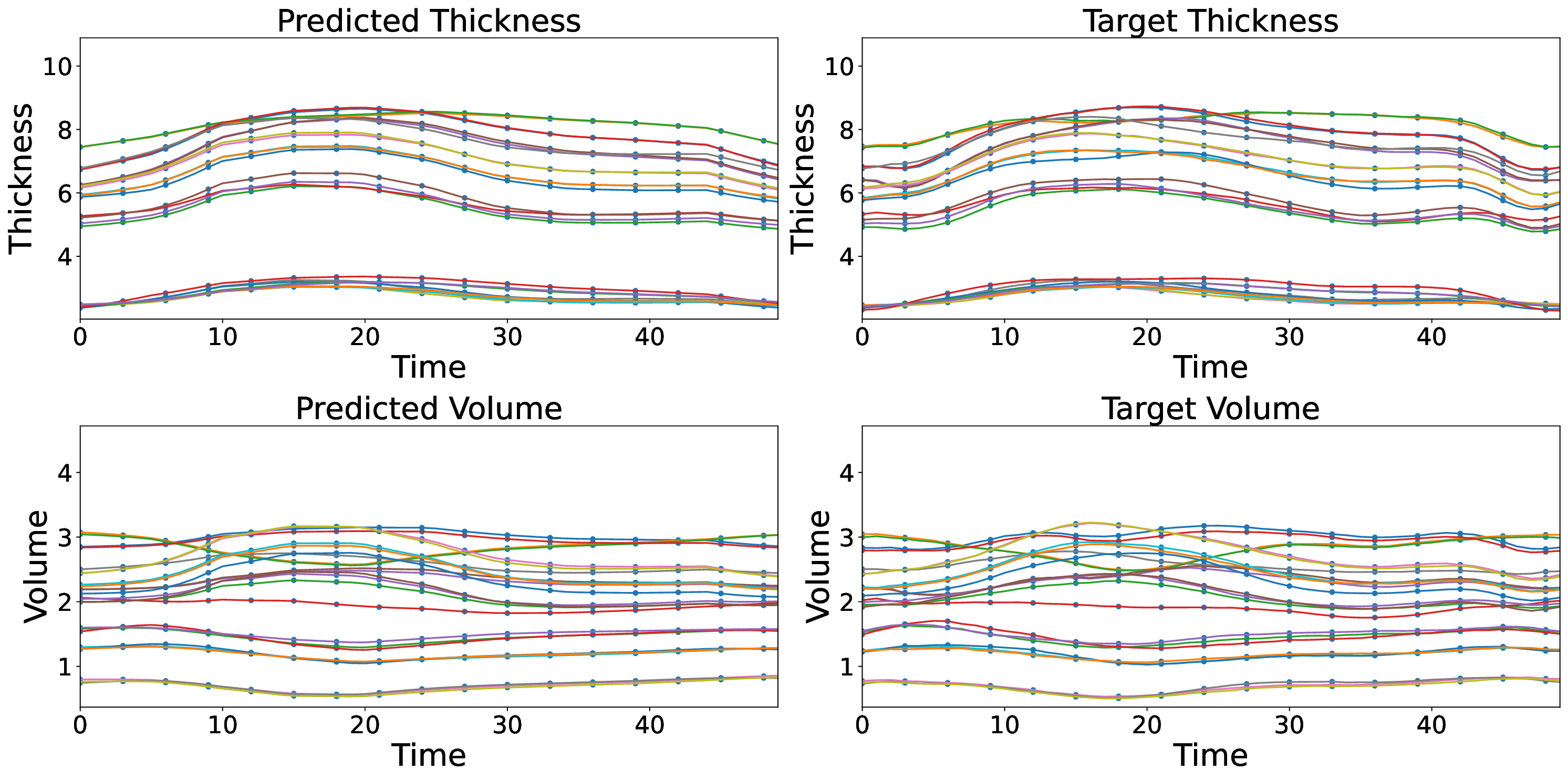}
        \subcaption{Control Trajectory}
        \label{subfig:trajectory_control_aha}
    \end{minipage}
    \begin{minipage}{0.4\textwidth}
        \centering
        \includegraphics[width=\linewidth]{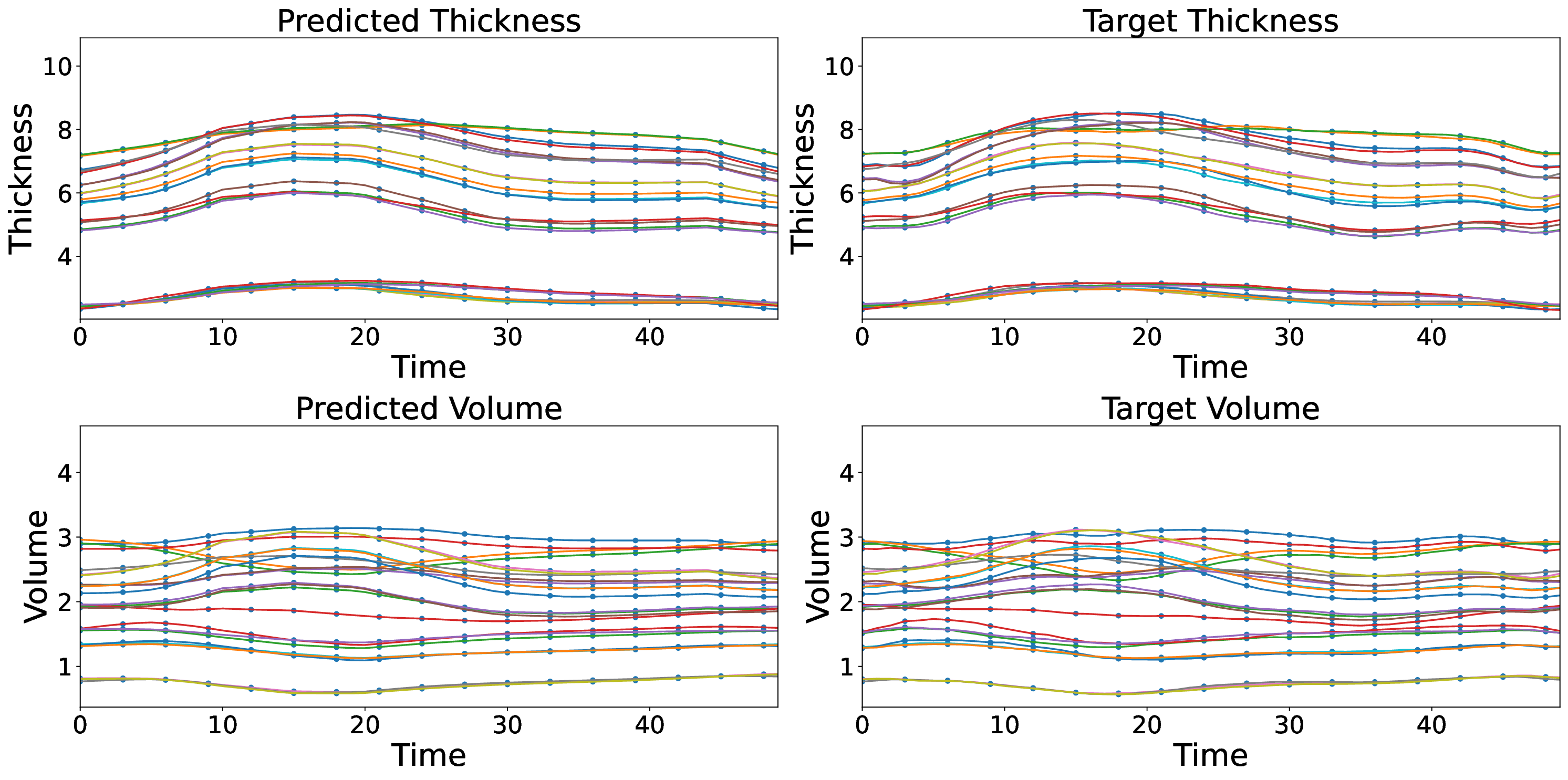}
        \subcaption{AFib Trajectory}
        \label{subfig:trajectory_afib_aha}
    \end{minipage}    
    \caption{Group trajectories on UKB dataset using anatomical connectivity (Anat). AFib: Atrial fibrillation.}
    \label{fig:group_trajectories_aha_ukb}
\end{figure*}

\end{document}